\documentclass[sn-mathphys]{sn-jnl}

\jyear{2021}%

\theoremstyle{thmstyleone}%
\theoremstyle{thmstyletwo}%

\theoremstyle{thmstylethree}%

\usepackage{multicol}
\usepackage{hyperref}
\usepackage{textgreek}
\usepackage{caption}
\usepackage{longtable}
\usepackage{subcaption}
\usepackage{diagbox}
\usepackage{rotating}

\begin{document}

\title[Explain Black-box Models for Early AD Detection using Multiple Datasets]{Machine Learning Workflow to Explain Black-box Models for Early Alzheimer's Disease Classification Evaluated for Multiple Datasets}


\author[1,2]{\fnm{Louise} \sur{Bloch}}\email{louise.bloch@fh-dortmund.de}

\author*[1,2]{\fnm{Christoph M.} \sur{Friedrich}}\email{christoph.friedrich@fh-dortmund.de}

\author[]{\fnm{} \sur{for the Alzheimer's Disease Neuroimaging Initiative}}
\equalcont{Membership of the Alzheimer's Disease Neuroimaging Initiative is listed in the Acknowledgments.}

\affil*[1]{\orgdiv{Department of Computer Science}, \orgname{University of Applied Sciences and Arts Dortmund}, \orgaddress{\street{Emil-Figge-Str. 42}, \city{Dortmund}, \postcode{44227}, 
		\country{Germany}}}

\affil[2]{\orgdiv{Institute for Medical Informatics, Biometry and Epidemiology (IMIBE)}, \orgname{University Hospital Essen}, \orgaddress{\street{Hufelandstr. 55}, \city{Essen}, \postcode{45147}, 
		\country{Germany}}}


\abstract{\textbf{Purpose:} Hard-to-interpret Black-box Machine Learning (ML) were often used for early Alzheimer's Disease (AD) detection.  
	
	\textbf{Methods:} To interpret eXtreme Gradient Boosting (XGBoost), Random Forest (RF), and Support Vector Machine (SVM) black-box models a workflow based on Shapley values was developed. All models were trained on the Alzheimer's Disease Neuroimaging Initiative (ADNI) dataset and evaluated for an independent ADNI test set, as well as the external Australian Imaging and Lifestyle flagship study of Ageing (AIBL), and Open Access Series of Imaging Studies (OASIS) datasets. Shapley values were compared to intuitively interpretable Decision Trees (DTs), and Logistic Regression (LR), as well as natural and permutation feature importances. To avoid the reduction of the explanation validity caused by correlated features, forward selection and aspect consolidation were implemented.
	
	\textbf{Results:} Some black-box models outperformed DTs and LR. The forward-selected features correspond to brain areas previously associated with AD. Shapley values identified biologically plausible associations with moderate to strong correlations with feature importances. The most important RF features to predict AD conversion were the volume of the amygdalae, and a cognitive test score. Good cognitive test performances and large brain volumes decreased the AD risk.  The models trained using cognitive test scores significantly outperformed brain volumetric models ($p<0.05$). Cognitive Normal (CN) vs. AD models were successfully transferred to external datasets. 
	
	\textbf{Conclusion:}
	In comparison to previous work, improved performances for ADNI and AIBL were achieved for CN vs. Mild Cognitive Impairment (MCI) classification using brain volumes. The Shapley values and the feature importances showed moderate to strong correlations.}




\maketitle

\section{Introduction}
\label{Sec:Introduction}
Alzheimer's Disease (AD) is a neurodegenerative disease \cite{10.1002/alz.12068} and the most frequent cause of dementia. As the number of dementia patients increases continuously, AD is a globally growing health problem \cite{WAR2018}. Currently, there is no causal therapy to cure AD \cite{10.1002/alz.12068}. To recruit and monitor subjects for therapy studies, it is important to identify patients at risk to develop AD early and to develop preclinical markers. Subjects with cognitive impairments that do not interfere with everyday activities are considered as having Mild Cognitive Impairment (MCI) due to AD~\cite{10.1016/j.jalz.2011.03.003}. The risk to develop AD is increased for subjects with MCI in comparison to cognitively normal controls (CN). However, not all subjects with MCI prospectively convert to AD. One possibility for early AD detection is to find patterns distinguishing between progressive MCI subjects (pMCI) who will develop AD and subjects with stable MCI (sMCI).

Multiple Machine Learning (ML) workflows were implemented for this differentiation. Some used models like Decision Trees (DTs) or Logistic Regression (LR), which were interpretable by design. However, black-box models like eXtreme Gradient Boosting (XGBoost) \cite{10.1145/2939672.2939785}, Random Forests (RFs) \cite{10.1023/A:1010933404324}, or Convolutional Neural Networks (CNNs) \cite{10.1038/nature14539} often outperform those models. Black-box models are designed to identify highly complex associations and are challenging to interpret. Thus, the risk of learning spurious decision functions caused by patterns occurring in the training dataset is increased for black-box models \cite{10.1038/s41467-019-08987-4}.

This research is an extended version of earlier work \cite{Bloch2021} and thus expands the previously developed ML workflow. The previously developed workflow enabled the interpretation of black-box models based on model-agnostic Shapley values. Shapley values give individual explanations for the prediction of each subject and visualize complex relationships between features and model predictions. In this research, the previous experiments are expanded by using three AD datasets and three adjusted feature sets. In addition to the previously trained tree-based models, Support Vector Machines (SVMs) \cite{SVMs} and LR models were implemented and explained. In this work, Shapley-based explanations were compared to classical feature importance methods, absolute log odd's ratios, and permutation importance.

In comparison to previous work \cite{Bloch2021}, an improvement of the classification results for ADNI and AIBL was achieved for the differentiation between Cognitive Normal (CN) controls and MCI subjects and MCI vs. AD classification for models trained without cognitive test scores and validated for AIBL. Additionally, the ADNI and AIBL results achieved for sMCI vs. pMCI classification, trained with cognitive test scores, outperformed previous work.

This article is structured as follows: In Section \ref{Sec:RelatedWork} related work is described. Section \ref{Sec:MaterialsAndMethods} introduces the datasets and methods used to implement the ML workflow and the details of the experiments. Section \ref{Sec:Results} elaborates on the experimental results. Those results are discussed including the mentioning of limitations in Section \ref{Sec:Discussion}. Finally, Section \ref{Sec:Conclusion} concludes the overall work.
\section{Related Work}\label{Sec:RelatedWork}
Interpretable ML was developed to explain black-box models \cite{Molnar2019}. As the heterogeneous etiology of AD is not completely understood yet, interpretability is important and enables the validation of the biological plausibility of ML models. Recently, some studies have used interpretable ML in AD detection.

For example, Long Short-Term Memory- (LSTM-) \cite{Hochreiter1997} based Recurrent Neural Networks (RNN) \cite{10.1038/323533a0} were trained to classify CN vs. MCI subjects in \cite{10.1371/journal.pone.0236868}. The experiments included multiple techniques to fuse sociodemographic and genetic data with Magnetic Resonance Imaging (MRI) scans. The resulting models were evaluated for two AD datasets -- the AD subset \cite{10.1159/000320988} of the Heinz Nixdorf Risk Factors Evaluation of Coronary Calcification and Lifestyle (RECALL) (HNR) \cite{10.1067/mhj.2002.123579} (61 MCI and 59 CN) and 624 subjects (397 MCI, 227 CN) of the Alzheimer's Disease Neuroimaging Initiative (ADNI) \cite{10.1212/WNL.0b013e3181cb3e25} study phase 1. To visually explain individual model decisions, Gradient-weighted Class Activation Mapping (Grad-CAM) \cite{10.1109/ICCV.2017.74} was used. A focus on biologically plausible regions was observed.

Four heatmap visualization methods -- sensitivity analysis \cite{Simonyan2014}, guided backpropagation \cite{Springenberg2014}, occlusion \cite{10.1007/978-3-319-10590-1_53}, and brain area occlusion inspired by \cite{Yang2018} -- were compared for 3D-CNNs in \cite{10.1007/978-3-030-02628-8_3}. The CNN models were trained using 969 MRI scans of 344 ADNI subjects (151 CN, 193 AD). However, it was unclear whether the described workflow ensured independent training and test sets by using multiple scans per subject \cite{10.1016/j.media.2020.101694}. Thus, the Cross-Validation (CV) accuracy of $77~\%\pm6~\%$ might be affected by data leakage. All heatmaps focused on AD-related anatomical brain areas.

An interpretable deep learning model, consisting of a Generative Adversarial Network \cite{Goodfellow2014} to extend the training dataset, a regression network to generate feature vectors from adjacent visits, and a classification model was introduced in \cite{10.1101/2019.12.12.874784}. Firstly, the regression model iteratively estimated the feature vector at the following visit. The resulting feature vector was used as input for the classification model, which predicted the final diagnosis. To classify 101 pMCI vs. 115 sMCI ADNI subjects, longitudinal volumetric MRI features were used. The model outperformed SVMs and artificial neural networks. 

A new interpretable model, based on distinct weighted rules was introduced in \cite{10.7717/peerj.6543} and evaluated for 151 subjects (97 AD and 54 CN) of the ADNI cohort. The framework is called Sparse High-order Interaction Model with Rejection option (SHIMR) and consists of two hierarchical stages. In the first stage, the interpretable model was trained using plasma features. The data of subjects with an unclear prediction in this stage were propagated to the second stage. In this stage, an SVM \cite{SVMs} was trained using invasive Cerebrospinal Fluid (CSF) markers. The evaluation included both, CV and an independent test set. The described model reached an Area Under the Receiver Operating characteristics Curve (AUROC) of 0.81 for the test set.

SHapley Additive exPlanations (SHAP) \cite{Lundberg2017} were used in \cite{10.1186/s13195-021-00879-4} to explain differences in models trained using coreset selection methods. The idea was, to determine coresets of subjects with the most informative data. RF and XGBoost models were trained on these coresets to avoid overfitting and improve ML models. The results of Data Shapley~\cite{Ghorbani2019data} coreset selection were compared to Leave-One-Out~\cite{10.2307/1268249} selection and random exclusion. All models were trained and validated for the ADNI dataset (400 sMCI, 319 pMCI) and externally validated for a subset of the AIBL dataset (16 sMCI, 12 pMCI). SHAP summary plots showed that models trained for both the entire training set and the coreset learned biologically plausible associations.

To examine the predictive influence of \textbeta-amyloid plaques, tau tangles, and neurodegeneration during the disease progression, RF feature importance was used in \cite{10.1038/s42003-020-1079-x}. The experimental data included 405 ADNI subjects (148 CN, 147 MCI, 110 AD). \textbeta-amyloid Positron Emission Tomography (PET) detected \textbeta-amyloid plaques, invasive CSF features surrogated tau tangles, and MRI and Fluorodeoxyglucose (FDG) PET scans were used to determine neurodegeneration. The experimental results showed that models trained to classify the early AD stages preferred features representing tau tangles and \textbeta-amyloid plaques. Models trained to predict later stages favored surrogates for neurodegeneration. SHAP \cite{Lundberg2017} and Gradient Tree Boosting (GTB) \cite{10.1214/aos/1013203451} reproduced those observations. The RF and the entire feature set reached accuracies of 73.17~\% (CN vs. MCI), 71.01~\% (MCI vs. AD), and 90.34~\% (CN vs. AD).

SHAP values were also used in \cite{10.3389/fdata.2021.613047} to explain population-based and individual predictions of XGBoost models and RFs. Models were trained using sociodemographic and lifestyle factors to predict the patient's risk to develop AD based on medical history. Transfer learning applied information extracted from the Survey of Health, Ageing, and Retirement in Europe (SHARE) \cite{10.1093/ije/dyt088} (80,699 CN, 4,157 AD) to the PREVENT cohort \cite{10.1136/bmjopen-2012-001893} (109 subjects with high risk to develop AD, 364 subjects with low risk). The PREVENT cohort was younger than the SHARE cohort. The models support the hypothesis that age is the most important risk factor in AD detection. Consistent with previous research \cite{10.1016/s0140-6736(20)30367-6}, amongst other factors, less education, physical inactivity, diabetes, and infrequent social contact were identified as potential risk factors.

A two stage-based classification workflow that used SHAP values to interpret RFs was developed in \cite{10.1038/s41598-021-82098-3}. In the first stage, CN vs. MCI vs. AD classification was performed. The second stage implemented the differentiation of sMCI and pMCI subjects. The models were based on multiple modalities including MRI, PET, CSF biomarkers, cognitive tests, medical history, genetics, and many more. The RFs were trained and tested using 1,048 subjects (294 CN, 254 sMCI, 232 pMCI, and 268 AD) of the ADNI dataset. For CN vs. MCI vs. AD classification, the model almost exclusively selected cognitive test scores as the most important features. The model learned bad cognitive test results increased the risk of AD and MCI. The most important features for sMCI vs. pMCI classification also were cognitive test scores followed by PET and MRI features. Bad cognitive test scores, small MRI volumes, and small PET uptakes were associated with disease progression.

	\begin{sidewaystable}
	\caption{Summary of the related work
	}
	\label{Table:RelatedWork}
		\begin{center}
				\begin{tabular}{p{0.5cm}p{2.5cm}p{4cm}p{4cm}p{3cm}p{3cm}}
					\toprule
				Ref.&Task&Subjects&Modality&ML method&Explanability method\\\midrule
				\cite{10.1371/journal.pone.0236868}&CN vs. MCI&HNR: 61 MCI, 59 CN; ADNI-1: 397 MCI, 227 CN&MRI, socio-demography, ApoE&LSTM based RNN&GradCAM\\
				\cite{10.1007/978-3-030-02628-8_3}&CN vs. AD&ADNI: 151 CN, 193 AD&MRI&CNN&sensitivity analysis, guided backpropagation, occlusion, brain area occlusion\\
				\cite{10.1101/2019.12.12.874784}&sMCI vs. pMCI&ADNI: 101 pMCI, 115 sMCI&MRI volumes&Neural network&intrinsic\\
				\cite{10.7717/peerj.6543}&CN vs. AD&ADNI: 54 CN, 97 AD&CSF, Plasma&SHIMR& intrinsic\\
				\cite{10.1186/s13195-021-00879-4}&sMCI vs. pMCI&ADNI: 400 sMCI, 319 pMCI; AIBL: 16 sMCI, 12 pMCI&MRI volumes, demography, ApoE&RF, XGBoost&SHAP\\
				\cite{10.1038/s42003-020-1079-x}&CN vs. MCI, CN vs. AD, MCI vs. AD&ADNI: 148 CN, 147 MCI, 110 AD&Amyloid-PET, MRI, FDG-PET, CSF&RF, GTB&RF-Feature importance, SHAP\\
				\cite{10.3389/fdata.2021.613047}&high vs. low risk&SHARE: 80,699 CN, 4,157 AD; PREVENT: 364 low risk, 109 high risk&sociodemography, lifestyle&RF, XGBoost&SHAP\\
				\cite{10.1038/s41598-021-82098-3}&CN vs. MCI vs. AD, sMCI vs. pMCI&ADNI: 294 CN, 254 sMCI, 232 pMCI, 268 AD&MRI, CSF, PET, cognitive tests, medical history, genetics&RF&ensemble of surrogat models, SHAP\\
					\botrule
				\end{tabular}
		\end{center}
		\end{sidewaystable}
\section{Materials and Methods}
\label{Sec:MaterialsAndMethods}
\begin{figure}
	\includegraphics[width=1\textwidth]{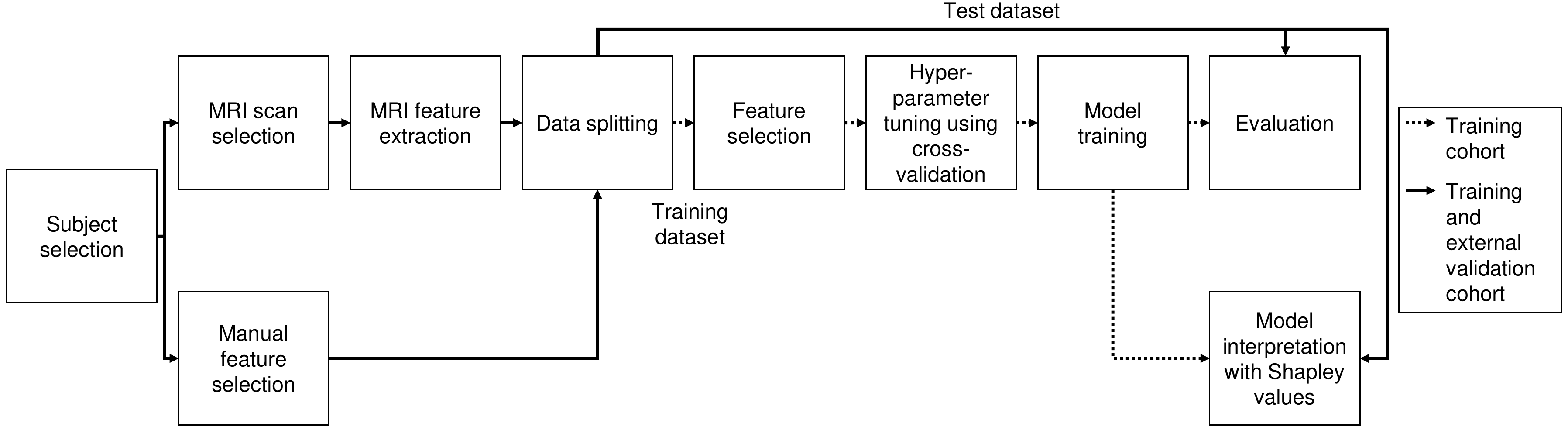}
	\caption{Implemented ML workflow. Volumetric features were extracted for one baseline (BL) MRI scan per subject. The ADNI dataset was randomly split into an 80~\% training and 20~\% test set. The most important MRI features were selected using forward feature selection, those were concatenated with sociodemographic features, number of ApolipoproteinE$\epsilon$4 (ApoE$\epsilon$4) alleles, and cognitive test scores. Bayesian Optimization implemented hyperparameter tuning. Black-box RFs, XGBoost models, LR models, as well as polynomial and radial SVMs, were trained and validated. Shapley values were calculated for black-box model interpretation. An evaluation was performed for the independent ADNI test set and for the external AIBL and OASIS datasets.}
	\label{fig:MachineLearningWorkflow} 
\end{figure}

The ML workflow, implemented using the programming language Python v3.6.9 \cite{Python3} is shown in Figure~\ref{fig:MachineLearningWorkflow}. It enables the interpretation of black-box models trained to detect early AD. In the following, the workflow and the methods used for implementation are elucidated.
\subsection{Datasets}
Data used in the preparation of this article were obtained from the ADNI \cite{10.1212/WNL.0b013e3181cb3e25}, the AIBL \cite{10.1017/S1041610209009405}, and the OASIS \cite{10.1101/2019.12.13.19014902} cohorts. 

ADNI (\url{https://adni.loni.usc.edu}, Accessed: 2022-05-01) was launched in 2003 as a public-private partnership. The primary goal of ADNI is to test whether a combination of biomarkers can measure the progression of MCI and AD. Those biomarkers include serial MRI, PET, biological markers, as well as clinical and neuropsychological assessments. The ongoing ADNI cohort recruited subjects from more than 60 sites in the United States and Canada and consists of four phases (ADNI-1, ADNI-2, ADNIGO, and ADNI-3). The subjects were assigned to three diagnostic groups. CNs have no problems with memory loss. Subjects with AD meet the criteria for probable AD defined by the National Institute of Neurological and Communicative Disorders and Stroke–Alzheimer's Disease and Related Disorders Association (NINCDS-ADRDA)~\cite{10.1212/WNL.34.7.939}. The diagnostic criteria of ADNI were explained in~\cite{10.1212/WNL.0b013e3181cb3e25}.
The dataset was downloaded on 27 Jul 2020 and initially included 2,250 subjects.
\begin{table}
	\caption{ADNI demographics at BL. The mean ($\bar{x}$) and standard deviation ($\sigma$) are given for all continuous variables
	}
	\label{Table:ADNIBaseline}
			\begin{minipage}{\linewidth}
					\begin{center}
			\begin{tabular}{p{1.65cm}|p{0.4cm}p{0.9cm}p{0.7cm}p{1cm}p{1.3cm}p{0.9cm}p{2cm}}
					\toprule
					&n&Age&Gender&Education&MMSCORE&CDR&ApoE$\epsilon$4\footnotemark[1]\\
					&&years&f~in~\%&years&$\bar{x}\pm\sigma$&$\bar{x}\pm\sigma$&0/1/2~in~\%\\
				 \midrule
					CN&512&74.2$\pm$5.8&51.8&16.3$\pm$2.7&29.1$\pm$1.1&0.0$\pm$0.0&71.3/26.2/ 2.3\\
					MCI&853&73.1$\pm$7.6&40.8&15.9$\pm$2.9&27.6$\pm$1.8&0.5$\pm$0.0&49.4/39.5/10.8\\
					sMCI&400&73.2$\pm$7.5&40.2&15.8$\pm$3.0&27.8$\pm$1.8&0.5$\pm$0.0&56.8/34.0/ 9.2\\
					pMCI&319&74.0$\pm$7.1&40.1&15.9$\pm$2.8&27.0$\pm$1.7&0.5$\pm$0.0&34.2/49.5/16.3\\
					AD&335&75.0$\pm$7.8&44.8&15.2$\pm$3.0&23.2$\pm$2.1&0.8$\pm$0.3&33.1/47.2/19.1\\\midrule
					$\Sigma_{CN, MCI,AD}$&1,700&73.8$\pm$7.2&44.9&15.9$\pm$2.9&27.2$\pm$2.7&0.4$\pm$0.3&52.8/37.0/ 9.9\\\botrule
			\end{tabular}
				\end{center}
		\footnotetext[1]{For 6 ADNI subjects (1 CN, 3 MCI, 2 AD) the number of ApoE$\epsilon$4 alleles was missing.}
		
	\end{minipage}
\end{table}

AIBL (\url{https://aibl.csiro.au/}, Accessed: 2022-05-01) is the largest AD study in Australia and was launched in 2006. AIBL aims to discover biomarkers, cognitive test results, and lifestyle factors associated with AD. As AIBL focuses on early AD stages, most of the subjects are CN. The MCI subjects of AIBL met the criteria described in \cite{10.1111/j.1365-2796.2004.01380.x}, AD diagnoses following the NINCDS-ADRDA criteria~\cite{10.1212/WNL.34.7.939} for probable AD. The diagnostic criteria of AIBL were described in~\cite{10.1017/S1041610209009405}. Approximately half of the CN subjects recruited in AIBL show memory complaints~\cite{10.1017/S1041610209009405}. 
AIBL data version 3.3.0 was downloaded on 19 Sep 2019 and originally included 858 subjects. 

\begin{table}[h]	\caption{AIBL demographics at BL. The mean ($\bar{x}$) and standard deviation ($\sigma$) are given for all continuous variables}
	\label{Table:AIBLBaseline}
			\begin{minipage}{\linewidth}
				\begin{center}
				\begin{tabular}{l|rrrrrr}
					\toprule
					&n&Age&Gender&MMSCORE&CDR&ApoE$\epsilon$4\footnotemark[1]\\
					&&years&f in \%&$\bar{x}\pm\sigma$&$\bar{x}\pm\sigma$&0/1/2~in~\%\\
					\midrule
					CN&446&72.5$\pm$6.1&57.0&28.7$\pm$1.2&0.0$\pm$0.1&69.3/26.5/ 2.7\\
					MCI&95&75.4$\pm$7.0&47.4&27.1$\pm$2.2&0.5$\pm$0.1&47.4/36.8/12.6\\
					sMCI&16&77.8$\pm$6.9&37.5&28.0$\pm$1.7&0.4$\pm$0.2&56.2/37.5/ 6.2\\
					pMCI&12&75.3$\pm$5.8&33.3&26.2$\pm$1.6&0.5$\pm$0.0&16.7/50.0/33.3\\
					AD&71&73.1$\pm$6.6&59.2&20.5$\pm$5.7&0.9$\pm$0.6&29.6/49.3/18.3\\\midrule
					$\Sigma_{CN, MCI,AD}$&612&73.0$\pm$6.6&55.7&27.5$\pm$3.5&0.2$\pm$0.4&61.3/30.7/ 6.0\\	\botrule
			\end{tabular}
		\footnotetext[1]{For 12 AIBL subjects (7 CN, 3 MCI, 2 AD) the number of ApoE$\epsilon$4 alleles was missing.}
	\end{center}
	\end{minipage}
\end{table}
The aim of the Open Access Series of Imaging Studies (OASIS) 3 (\url{https://www.oasis-brains.org/}, Accessed: 2022-05-01) \cite{10.1101/2019.12.13.19014902} dataset is, to investigate the effects of healthy ageing and AD.
The subjects of OASIS-3 were recruited from several ongoing studies in the Washington University Knight Alzheimer Disease Research Center (\url{https://knightadrc.wustl.edu/}, Accessed: 2022-05-01). The longitudinal dataset included MRI scans, fMRI scans, Amyloid- and FDG-PET scans, neuropsychological test results, and clinical data for 1,098 subjects. OASIS focuses on the preclinical stage of AD. All OASIS subjects had a Clinical Dementia Rating (CDR) less than or equal to 1. The OASIS dataset provides multiple target values. In this research, CN subjects had normal cognition and absence of MCI or AD diagnosis, MCI subjects had amnestic MCI with memory impairment and AD diagnosis follows the NINCDS-ADRDA criteria~\cite{10.1212/WNL.34.7.939} for probable AD.

\begin{table}[h]
	\caption{OASIS-3 demographics at the first visit with MRI scan and diagnosis. The mean ($\bar{x}$) and standard deviation ($\sigma$) are given for all continuous variables}
	\label{Table:OASISBaseline}
		\begin{minipage}{\linewidth}
			\begin{center}	
				\begin{tabular}{l|rrrrrr}
				\toprule
					&n&Age&Gender\footnotemark[1]&MMSCORE&CDR&ApoE$\epsilon$4\footnotemark[2]\\
					&&years&f in~\%&$\bar{x}\pm\sigma$&$\bar{x}\pm\sigma$&0/1/2~\%\\\midrule
					CN&704&68.3$\pm$9.3&58.7&29.1$\pm$1.2&0.0$\pm$0.1&65.8/29.7/4.1\\
					MCI&19&76.7$\pm$7.0&36.8&28.1$\pm$1.4&0.3$\pm$0.2&57.9/42.1/0.0\\
					AD&198&75.6$\pm$7.9&48.5&24.8$\pm$4.0&0.7$\pm$0.3&38.9/51.5/9.1\\\midrule
					$\Sigma_{CN, MCI,AD}$&921&70.1$\pm$9.5&56.0&28.1$\pm$2.8&0.2$\pm$0.3&59.8/34.6/5.1\\\botrule		
			\end{tabular}
			\end{center}
		\end{minipage}
\footnotetext[1]{For 1 OASIS subjects (1 CN) the gender was missing.}
\footnotetext[2]{For 4 OASIS subjects (3 CN, 1 AD) the number of ApoE$\epsilon$4 alleles was missing.}
\end{table}
\subsubsection{Subject Selection}\label{SEC:SubjectSelection}
For the ADNI dataset, all subjects with an MRI scan at the baseline visit were included. 521 subjects who have no MRI scan at the baseline visit were excluded, 29 subjects failed the MRI feature extraction described in Section \ref{SEC:FeatureExtraction}. The demographics of the resulting 1,700 subjects are summarized in Table \ref{Table:ADNIBaseline}.

The 853 MCI subjects were divided into two groups. The sMCI subjects had a stable MCI diagnosis at all follow-up visits and the pMCI subjects converted to a stable AD diagnosis at any visit. 38 subjects with no follow-up visits and 96 subjects who reverted to CN or MCI were excluded from this separation, resulting in 400 sMCI and 319 pMCI subjects.

For AIBL, the same exclusion criteria were applied. Therefore, 170 subjects had no MRI scan at the baseline visit, the baseline MRI scans of 76 subjects failed for the MRI feature extraction pipeline described in Section \ref{SEC:FeatureExtraction}. The demographics of the resulting 612 subjects are summarized in Table \ref{Table:AIBLBaseline}. Similar to the ADNI dataset, the 95 MCI subjects were divided into two groups. In this step, 60 subjects with no follow-up visits and 7 subjects who reverted to CN or MCI were excluded from this separation, resulting in 16 sMCI and 12 pMCI subjects.

The exclusion criteria were similarly applied for the OASIS-3 dataset, which originally included 1,098 subjects. For 983 subjects, a diagnosis of CN, MCI, or AD was assigned for at least one visit. The MRI feature extraction pipeline failed for all MRI scans of five subjects, no MRI scan was successfully matched to a diagnosis with a tolerance of 365 days for 57 subjects. In contrast to the ADNI and AIBL datasets, which exclusively included baseline visits, the first visit with an MRI scan and a diagnosis was used for OASIS. The demographics of the remaining 921 subjects are summarized in Table \ref{Table:OASISBaseline}.

The number of subjects with MCI as baseline diagnosis is 19. This number was decreased if subjects without follow-up diagnoses were excluded. Thus, no experiments were executed to separate sMCI and pMCI subjects in OASIS-3. For reproducible research, the supplementary material contains lists with the subject and MRI IDs and the diagnoses for all datasets.
\subsubsection{MRI Scan Selection}
From the ADNI dataset, T1-weighted MRI scans recorded at the baseline visit were included.
The acquisition parameters differ between scanners. During the ADNI-1 study phase, scans were recorded using a field strength of 1.5~T. In the remaining study phases, MRI scans with a field strength of 3.0~T were recorded. 

From the AIBL dataset, T1-weighted MRI scans following the protocol of the ADNI 3D T1-weighted sequences were included. All AIBL scans had a resolution of $1\times1\times1.2$~mm. 

For the OASIS-3 dataset, T1-weighted MRI scans, recorded on three scanners, were included. The field strengths of those scanners are 1.5~T and 3.0~T \cite{10.1101/2019.12.13.19014902}.
\subsection{MRI Feature Extraction}\label{SEC:FeatureExtraction}
Using FreeSurfer v6.0 \cite{10.1016/j.neuroimage.2012.01.021}, volumetric features were extracted for each MRI scan. These include the volumes of 34 cortical areas per hemisphere of the Desikan-Killiany atlas \cite{10.1016/j.neuroimage.2006.01.021}, 34 subcortical areas \cite{10.1016/S0896-6273(02)00569-X}, and the estimated Total Intracranial Volume (eTIV). As recommended for volumes in \cite{10.1007/s10548-012-0246-x}, the volumetric features were normalized by eTIV. 
This results in 103 MRI volumes, which were split into 49 features of the left hemisphere, 49 features of the right hemisphere, and five additional unpaired segmentations (3rd ventricle, 4th ventricle, brain stem, CSF, eTIV). 

After the normalization, for paired volumes, the sum (described in Equation \ref{EQN:SUM}), the difference (described in Equation \ref{EQN:DIFF}), and the ratio (described in Equation \ref{EQN:RATIO}) of both hemispheres are calculated to investigate symmetry and to decrease feature interactions. This results in 152 MRI features (49 sums, 49 differences, 49 ratios, and five unpaired features). Brain asymmetry was previously associated with AD \cite{10.3390/technologies5020016,10.1038/s41467-021-21057-y,10.3389/fnins.2020.00602,10.1016/j.dadm.2019.08.001}.
Equation \ref{EQN:DIFF} shows that differences were calculated by subtracting the right from the left volume similar to \cite{10.1038/s41467-021-21057-y}, where the cortical thickness was used instead of volumetric features.

\begin{equation}\label{EQN:SUM}
	sum_{ROI}=\frac{vol_{ROI}^{left}}{eTIV}+\frac{vol_{ROI}^{right}}{eTIV}
\end{equation}
\begin{equation}\label{EQN:DIFF}
	diff_{ROI}=\frac{vol_{ROI}^{left}}{eTIV}-\frac{vol_{ROI}^{right}}{eTIV}
\end{equation}
\begin{equation}\label{EQN:RATIO}
	ratio_{ROI}=\frac{vol_{ROI}^{left}}{vol_{ROI}^{right}}
\end{equation}
\subsection{Manual Feature Preselection}\label{SEC:ManualFeaturePreselection}
Three feature sets were investigated in the experiments. The manual feature selection aims to choose less invasive, accessible examination techniques which were able to detect early signs of AD. Feature set 1 (FS-1) includes all MRI features, and sociodemographic features including age, gender, and years of education. However, the years of education are only available for the ADNI dataset. Feature set 2 (FS-2) expands FS-1 by the number of ApoE$\epsilon$4 alleles, a genetic risk factor associated with AD, which can be obtained from blood samples or via less invasive swab tests from the inside surface of the cheek. Feature set 3 (FS-3) extended FS-2 by three cognitive tests including the score of the Mini-Mental State Examination (MMSCORE) and two logical tests to evaluate the long-term (Logical memory, delayed -- LDELTOTAL) and the short-term memory (Logical memory, immediate -- LIMMTOTAL). The CDR was strongly associated with AD diagnosis and was not included in the experiments.

\subsection{Dataset Splitting}
The ADNI dataset was split on the subject level into two distinct subsets. The training set included 80~\% of the data and the test set consisted of the remaining 20~\%. The splitting was executed within each diagnostic group to ensure similar distributions. The AIBL and OASIS datasets were used as external test sets. None of the AIBL and OASIS subjects was used in the training or model selection process.
\subsection{Feature Selection} 
Initially, 152 MRI features were extracted from the MRI scans. Those features are reduced to focus the ML models on the most important features. For this reason, feature forward selection was implemented. In comparison to feature selection methods like RF feature importance, this method avoids correlated features in the dataset \cite{10.1186/1471-2105-8-25}. Forward selection is a greedy procedure that iteratively identifies the best new feature until no improvement was reached. The training dataset was split into an 80~\% training set and 20~\% validation dataset. The training dataset was used to train the ML model used for classification with default hyperparameters on the feature set, and the validation dataset was used to calculate the validation accuracy for this feature set. The selected MRI features were expanded using the features described in Section \ref{SEC:ManualFeaturePreselection}.

\subsection{Hyperparameter Tuning}
To tune the hyperparameters of the ML models, Bayesian optimization \cite{10.1007/3-540-07165-2_55} was implemented using the Python package scikit-optimize v0.8.1 \cite{10.5281/zenodo.4014775}. Bayesian optimization maps the dependency of the hyperparameters and the model performance using a Gaussian Process. Initially, ten nearly random hyperparameter combinations were selected by a Latin Hypercube Design (LHD) \cite{10.2307/1268522}. Bayesian optimization with LHD initialization was successfully used in previous research \cite{10.1007/978-3-030-70569-5_18} to optimize the parameters for early AD detection. Each parameter was split into ten equidistant intervals and one sample was randomly chosen per interval. This results in ten samples per parameter, which were randomly matched.

A stratified $10\times10$-fold CV \cite{10.1007/978-0-387-39940-9_565} was applied to the training dataset to estimate the model accuracy for an independent test set. Stratified $10\times10$-fold CV was implemented by splitting each diagnostic group of the training dataset into ten distinct folds using the Python package scikit-learn v0.23.2 \cite{scikit-learn}. Ten iterations were performed, each with a different fold used as a validation dataset (10~\%). The training dataset included the remaining nine folds (90~\%). With shuffled data in each run, this procedure was repeated ten times. The ML model was initially evaluated for ten LHD combinations.

To predict the average CV accuracy for the initial parameter combinations, the Gaussian Process was fitted. Afterward, an optimization selected the next promising parameter combination. As an acquisition function, the Lower Confidence Bounds (equation~\ref{Eq:LCB}) was used. In this equation, $\hat{\mu}_\Theta$ is the Gaussian Process estimation of the CV accuracy and $\hat{\Sigma}_\Theta$ is the covariance at parameter combination $\Theta$. 

\begin{equation}\label{Eq:LCB}
	LCB(\Theta)=\hat{\mu}_\Theta-\hat{\Sigma}_\Theta
\end{equation}
The hyperparameter combination selected in the previous step was again evaluated using CV. Afterward, to refine the Gaussian Process and to determine the following combination, the respective tuple of hyperparameter and mean CV accuracy was added to the Gaussian Process. The procedure was repeated 25 times. The best hyperparameter combination was chosen to train the final model.
\subsection{Model Training}
During hyperparameter training and final model generation, XGBoost models, RFs, radial SVMs, polynomial SVMs, DTs, and LR models were trained. The preprocessing pipeline included centering, scaling, and median imputation. The entire preprocessing pipeline was implemented within the CV to avoid over-optimistic performance estimations \cite{10.1109/MCI.2018.2866730}. The parameters were calculated for the CV training set and reused for the test and external datasets. The preprocessing was implemented using the Python package scikit-learn v0.23.2 \cite{scikit-learn}.

Ensemble-based black-box XGBoost \cite{10.1145/2939672.2939785} models, follow the idea of gradient boosting models \cite{10.1214/aos/1013203451}. It means the combination of multiple weak classifiers results in a strong, joint classifier. By learning the gradients of the previous classifier, Gradient boosting fulfills this assumption. The final prediction consisted of the sum of weak classifier predictions. XGBoost is distributed as an open-source software library and the main advantages are scalability, parallelization, and distributed execution. The hyperparameters and intervals used during Bayesian optimization are summarized in Table~\ref{Table:Parameter}. The hyperparameter \texttt{n\_estimators} sets the number of boosting iterations, \texttt{learning\_rate} was the learning rate that preferences weak classifiers at early iterations, the minimum loss reduction required to split a node is defined by \texttt{gamma}, the hyperparameter \texttt{max\_depth} sets the maximum depth of an individual tree, and the minimum number of observations in a child node was denoted as \texttt{min\_child\_weight}, \texttt{subsample} and \textit{colsample\_bytree} set the proportion of randomly subsampled training instances and features per iteration. The Python package xgboost v1.2.0 \cite{Chen2019} implemented the XGBoost algorithm.

RF \cite{10.1023/A:1010933404324} training was implemented using the Python package scikit-learn v0.23.2 \cite{scikit-learn}. The RF algorithm is based on multiple DTs. Each DT was trained using randomly chosen features and subjects. Those subjects were selected using bootstrap sampling \cite{10.1214/ss/1177013815} on the training dataset. RF inference was computed by summarizing the individual DTs using a majority voting. The RF hyperparameters are summarized in Table \ref{Table:Parameter}. \texttt{n\_estimators} sets the number of DTs, each split used a random subset of \texttt{max\_features} features, the hyperparameter \texttt{min\_samples\_leaf} describes the minimum number of samples in a leaf node. 

Support Vector Machines (SVMs) \cite{SVMs} were implemented using the Python package scikit-learn v0.23.2 \cite{scikit-learn}. SVMs separate two classes using a decision boundary which was referred to as an $n$-dimensional hyperplane. Here, $n$ is the number of features. To increase the robustness of the hyperplane for unknown observations, SVMs select the hyperplane with the largest distance from the observations. For this reason, the distance between the hyperplane and the observations was maximized using the hinge loss function \cite{10.1162/089976604773135104}. The support vectors describe the observations closest to the hyperplane. Removing support vectors from the dataset directly influences the hyperplane. The cost parameter \texttt{C} enables SVMs to avoid overfitting, the higher \texttt{C}, the less complex an SVM is. Kernel functions help to model complex interactions. In this research, a polynomial and a radial kernel were implemented. The \texttt{degree} hyperparameter of the polynomial kernel controls the degree of the kernel, high values lead to more complex hyperplanes. The \texttt{gamma} hyperparameter constraints the influence, a single observation has on the hyperplane. If \texttt{gamma = scale}, $\frac{1}{\# features\cdot \sigma}$ was used as a value of gamma, if \texttt{gamma = auto}, a value of $\frac{1}{\# features}$ was used. The SVM hyperparameters and their ranges are summarized in Table \ref{Table:Parameter}.

\begin{table}
	\caption{Hyperparameters and intervals used to train the ML models}
	\label{Table:Parameter}
	\begin{minipage}{\linewidth}
		\begin{center}
		\begin{tabular}{llrr}
		\toprule
			Model&Hyperparameter&Minimum&Maximum\\
			\midrule
			XGBoost&\texttt{n\_estimator}&1&500\\
			&\texttt{max\_depth}&1&20\\
			&\texttt{learning\_rate}&$10^{-10}$&1\\
			&\texttt{gamma}&0&20\\
			&\texttt{min\_child\_weight}&1&30\\
			&\texttt{subsample}&0&1\\
			&\texttt{colsample\_bytree}&0&1\\
			\midrule
			RF&\texttt{n\_estimators}&250&1,250\\
			&\texttt{max\_features}&2&\# features\\
			&\texttt{min\_samples\_leaf}&1&20\\\midrule
			polynomial SVM&\texttt{C}&$10^{-4}$&$10^2$\\
			&\texttt{degree}&1&10\\
			&\texttt{gamma}&\texttt{scale}&\texttt{auto}\\\midrule
			radial SVM&\texttt{C}&$10^{-4}$&$10^2$\\
			&\texttt{gamma}&\texttt{scale}&\texttt{auto}\\\midrule
			DT&\texttt{criterion}&\texttt{gini}&\texttt{entropy}\\
			&\texttt{splitter}&\texttt{best}&\texttt{random}\\
			&\texttt{max\_depth}&1&100\\
			&\texttt{min\_samples\_split}&0&1\\
			\midrule
			LR&\texttt{C}&$10^{-4}$&$10^2$\\
			&\texttt{penalty}&\texttt{l2}&\texttt{none}\\
			
		\botrule
		\end{tabular}
	\end{center}
	\end{minipage}
\end{table}

In contrast to the black-box models, DTs \cite{10.1201/9781315139470} and LR models were selected as simple and interpretable comparison models. DTs were implemented using the Python package scikit-learn v0.23.2 \cite{scikit-learn}. A DT consists of successively learned decision rules of the form $x \leq t$ for numerical or $x\in t$ for categorical features $t$ is a threshold or a subset of values. The next decision rule was selected by the \texttt{splitter} which ranked all possible rules using a \texttt{criterion}. 
Decision rules were iteratively expanded until a maximum depth of \texttt{max\_depth} was met or a percentage \texttt{min\_samples\_split} of samples were in a split. 

LR \cite{cox1958regression} is a Generalized Linear Model (GLM) with a logistic link function. This link function allows the processing of binomial output variables. The logistic model function is given in equation \ref{Eq:logReg}. The model predicts the probability $P(Y=1\vert X=x,\Theta)$ of observation $x$ with given parameters $\Theta$ being in the positive class $Y=1$. The LR algorithm was implemented using the Python package scikit-learn v0.23.2 \cite{scikit-learn}. 
\begin{equation}\label{Eq:logReg}
	P(Y=1\vert X=x,\Theta)=\frac{1}{1+\exp(x\cdot\Theta)}
\end{equation}

\subsection{Model Interpretation with Shapley Values}
There are multiple methods to interpret ML models. An overview can be found in \cite{Molnar2019}. For example, DTs and LR models are interpretable by design. However, black-box models often outperform those interpretable models but the interpretation of black-box models is complicated. In this research, model-agnostic Shapley values were used. Shapley values are local models, which explain the predictions of individual observations and thus enable high clinical benefit and high adaption to the black-box model. 

Shapley values \cite{Shapley1953} are affiliated with coalition game theory and aim to decompose the prediction of a subject into the contributions of each feature. For this aim, Shapley values are based on the additive linear explanation model shown in equation \ref{Eq:AdditiveLinearModel}. For a subject $x$, the model prediction $f(x)$ is decomposed into the feature contributions~$\Phi_j$, a simplified representation of the feature values $x'$, and the average model prediction $\Phi_0$. A binned binary feature representation was used for tabular data, with $N$ being the number of simplified features.

\begin{equation}\label{Eq:AdditiveLinearModel}
	f(x)=\Phi_0+\sum_{j=1}^{N}\Phi_j x'_j
\end{equation}
The idea of using Shapley values to explain black-box ML models is, to fairly decompose the contribution of each feature for the subject's prediction. Due to this fairness, the sum of all Shapley values is equal to the difference between the average model prediction and the probability prediction of a subject. Equation~\ref{Eq:ShapleyValueOrig} shows Shapley values are defined as the average, weighted contribution, a simplified feature has in all subsets. For the exact calculation of a Shapley value $\Phi_i$ for a given subject and feature~$i$, it is required to determine the contribution of this feature for all subsets $S$ of the entire feature set $F$. The investigation of each subset $S$ requires the retraining and evaluation of the black-box model $f_{S}(S)$. With the help of the model performances trained with ($f_{S\cup i}(S\cup i)$) and without ($f_S(S)$) the feature at interest $i$, their differences were calculated. The weighted average difference across subsets builds the Shapley value. The weighting depends on the relative number of features $\vert S\vert$ in subset $S$. High weights were assigned to subsets with few and many features. In this way, the estimation of the main individual effects and the total effects are supported. 

\begin{equation}\label{Eq:ShapleyValueOrig}
	\Phi_i=\sum_{S\subseteq F\setminus\{i\}} \frac{\vert S\vert! (\vert F\vert-\vert S\vert-1)!}{\vert F\vert!}\bigl(f_{S\cup\{i\}}(S\cup i)-f_S(S)\bigr)
\end{equation}
However, the number of subsets increases exponentially with the number of input features, leading to high computational effort for the exact calculation of Shapley values. 
Kernel SHapley Additive exPlanations (SHAP) \cite{10.1007/s10115-013-0679-x} avoid time-consuming repeated training and evaluation by estimating Shapley values. This algorithm is based on Local Interpretable Model-agnostic Explanations (LIME)~\cite{10.18653/v1/n16-3020} and was implemented using the Python package shap v0.38.1 \cite{Lundberg2017}. A new dataset containing variants of the observation at interest is created by permuting selected features. An additive linear model (Eqn.~\ref{Eq:AdditiveLinearModel_SHAP}) with $x'$ is a simplified representation of the black-box input features and $g(x')$ is the explanation model was fitted to the generated dataset. 
\begin{equation}\label{Eq:AdditiveLinearModel_SHAP}
	g(x')=\Phi_0+\sum_{i=1}^{M}\Phi_i\cdot x'_i
\end{equation}
The weights $\Phi_i$ of the explanation model estimates the SHAP values for each subject and each feature. For tabular data, the simplified features are binned binary feature representations that represent if the original feature value or a permutation was used.

SHAP force plots \cite{Lundberg2018} explain the model prediction of individual subjects using Shapley values. Features with positive Shapley values show strong positive effects on the prediction and small negative Shapley values represent small negative effects. SHAP force plots can be found in Figure \ref{fig:SHAP_force_plots}.

SHAP summary plots \cite{Lundberg2018} summarize the explanations for the entire training dataset. Each point visualizes the feature value of a subject and the associated Shapley value. The color of a point depends on the subject's feature value. On the vertical axis, the features are ordered by the mean absolute Shapley values. The plots were limited to the top ten features. SHAP summary plots can be found in Figure \ref{fig:FeatureSet_comp_CNAD_svmPoly_rfMean_SHAP}, Figure \ref{fig:Reproducibility_Comparison_RF_CNAD_SHAP}, Figure \ref{fig:Reproducibility_Comparison_RF_CNMCI_SHAP}, and Figure \ref{fig:Comparison_ClassificationModels_sMCI_FS-3_RFMean_SHAP}.

There are some reasons, including out-of distribution sampling during Shapley value approximation and not taking into account feature correlation, why Shapley values should be used with caution for black-box model interpretability \cite{kumar2020problems}. Therefore, it is important to compare Shapley value results with other ML explanation methods, or to reduce or consolidate correlated features \cite{Pekala2021}. In this work, forward selection was implemented to reduce the number of correlated features in the dataset, Shapley values were compared to classical feature importance measurements (Section \ref{SEC:ClassificationModel}), and correlated features are consolidated to aspects.
\subsection{Evaluation}
The models were evaluated for the ADNI test set and the external AIBL and OASIS datasets. The performance was measured using accuracy (ACC) (equation \ref{EQN:ACCURACY}), balanced accuracy (BACC) (equation \ref{EQN:BALANCEDACCURACY}), F1-Score (F1) (equation \ref{EQN:F1}), and Matthews correlation coefficient (MCC) (equation \ref{EQN:MCC}). Table~\ref{TAB:CONTINGENCY_TABLE} visualizes the contingency table used for the calculation of those scores. Providing multiple scores for evaluation increased the comparability to other research. In comparison to accuracy, which focuses on correctly classified cases, the F1-Score focuses on incorrectly classified cases. The macro averaging F1-Score was calculated to address both, the diseased and the healthy subject classification. Balanced accuracy is based on both, sensitivity and specificity and thus is suitable to evaluate imbalanced class problems. The MCC returns a value between 0 and 1 and is also suitable to handle imbalanced datasets.

\begin{table}[h!]
	\caption{Contingency table for the classification between patients and controls}\label{TAB:CONTINGENCY_TABLE}
		\begin{minipage}{\linewidth}
\begin{center}
	
	\begin{tabular}{c|cc}
	\toprule
		\diagbox{Prediction}{Diagnosis}&patient&control\\\midrule
		patient&True positive (TP)& False positive (FP)\\
		control&False negative (FN)&True negative (TN)\\\botrule
	\end{tabular}
		\end{center}
\end{minipage}
\end{table}
\begin{equation}\label{EQN:ACCURACY}
	ACC=\frac{TP+TN}{TP+TN+FP+FN}
\end{equation}
\begin{equation}\label{EQN:BALANCEDACCURACY}
	BACC=\frac{\frac{TP}{TP+FN}+\frac{TN}{TN+FP}}{2}
\end{equation}
\begin{equation}\label{EQN:F1}
	F1=\frac{TP}{TP+\frac{1}{2}(FP+FN)}
\end{equation}
\begin{equation}\label{EQN:MCC}
	MCC=\frac{TP\cdot TN -FP \cdot FN}{\sqrt{(TP+FP)\cdot(TP+FN)\cdot (TN+FP)\cdot (TN+FN)}}
\end{equation}
Additionally, the Area under the Receiver Operating Curve (AUROC), which models the relationship between the True Positive Rate (TPR -- equation \ref{EQN:TPR}) and the False Positive Rate (FPR -- equation \ref{EQN:FPR}) for different classification thresholds was computed for all models. AUROC is suitable to investigate classification tasks with imbalanced datasets.
\begin{equation}\label{EQN:TPR}
	TPR=\frac{TP}{TP+FN}
\end{equation}
\begin{equation}\label{EQN:FPR}
	FPR=\frac{FP}{TN+FP}
\end{equation}
\section{Results} \label{Sec:Results}
In the following, the experimental results are presented. The MRI features selected using forward selection and the performances achieved for CN vs. AD, CN vs. MCI, MCI vs. AD, and sMCI vs. pMCI classification were given. SHAP summary plots compared the models trained using different feature sets, validation datasets, and classification models. The results of SHAP summary plots are compared to natural RF- and XGBoost-based feature importance scores and permutation importance scores. The influence of feature interactions for Shapley values is investigated and SHAP force plots explain individual model predictions of interesting subjects.
\subsection{Feature Selection}
The MRI features selected during forward selection for CN vs. AD classification and different ML methods used as base classifiers are summarized in Table \ref{Table:ForwardSelectionCNAD}.
In this research, feature forward selection was used to reduce the number of MRI features and the influence of correlated features.
\begin{table}
\caption{Features selected by forward selection using different ML methods as base classifiers for CN vs. AD classification. Feature selection was exclusively used to reduce the number of MRI features.}\label{Table:ForwardSelectionCNAD}
\begin{minipage}{\linewidth}
	\begin{center}
			\begin{tabular}{ccc}
				\toprule
				LR&DT&RF\\\midrule
				sum\_entorhinal&sum\_Amygdala&sum\_Amygdala\\
				sum\_Amygdala&sum\_entorhinal&diff\_parstriangularis\\
				ratio\_lingual&sum\_Hippocampus&diff\_superiorparietal\\
				sum\_middletemporal&ratio\_supramarginal&sum\_lateralorbitofrontal\\
				diff\_Lateral.Ventricle&&sum\_medialorbitofrontal\\
				ratio\_entorhinal&&\\
				\midrule
				XGBoost&SVM poly&SVM radial\\\midrule
			sum\_Amygdala&sum\_entorhinal&sum\_Amygdala\\
			sum\_middletemporal&sum\_inferiorparietal&sum\_entorhinal\\
			sum\_entorhinal&diff\_Cortex&diff\_Cortex\\
			diff\_lateralorbitofrontal&sum\_Amygdala&sum\_VentralDC\\
			&ratio\_paracentral&\\
			\botrule
			\end{tabular}
	\end{center}
	
\end{minipage}
\end{table}

 For the CN vs. AD detection task, the RF, and the polynomial SVM chose five features, the XGBoost, the DT, and the radial SVM chose four features and the LR chose six features. Overall, the six methods chose 16 different features. The most important feature for the RF, the XGBoost, the DT and the radial SVM was the sum of the left and right amygdalae. For the polynomial SVM and the LR, the most important feature was the sum of the entorhinal cortices. Both features were previously associated with AD detection \cite{10.1038/nrneurol.2009.215,10.1002/hbm.20934,10.1016/j.neurobiolaging.2003.12.007,10.1016/j.pscychresns.2011.06.014}. Previous research also shows, that most of the selected features are associated with atrophy patterns in AD \cite{10.1136/gpsych-2018-100005}. All methods also selected at least one difference or ratio of the left and right cortical or subcortical areas. Those features describe the asymmetry of both hemispheres. Brain asymmetry measurements were associated with AD \cite{10.3390/technologies5020016,10.1038/s41467-021-21057-y,10.3389/fnins.2020.00602,10.1016/j.dadm.2019.08.001} and were also successfully applied for ML models in this field \cite{10.3390/s21030778}.

The rankings of the forward selection for CN vs. MCI detection and different base classifiers are given in Table \ref{Table:ForwardSelectionCNMCI}. 

\begin{table}
	\caption{Features selected by forward selection using different ML methods as base classifiers for CN vs. MCI classification. Feature selection was exclusively used to reduce the number of MRI features.}\label{Table:ForwardSelectionCNMCI}
	\begin{minipage}{\linewidth}
		\begin{center}
				\begin{tabular}{ccc}
						\toprule LR&DT&RF\\\midrule
						sum\_middletemporal&sum\_insula&sum\_insula\\
						ratio\_isthmuscingulate&diff\_insula&diff\_isthmuscingulate\\
						diff\_paracentral&sum\_fusiform&sum\_inferiorparietal\\
						diff\_Cerebellum.White.Matter&&sum\_Cerebellum.White.Matter\\
						\midrule
						XGBoost&SVM poly& SVM radial\\\midrule
						ratio\_inferiorparietal&sum\_lingual&sum\_temporalpole\\
						sum\_CerebralWhiteMatter&sum\_Hippocampus&sum\_inferiortemporal\\
						ratio\_VentralDC&ratio\_rostralmiddlefrontal&sum\_caudalanteriorcingulate\\
						diff\_caudalanteriorcingulate&CSF&sum\_Lateral.Ventricle\\
						&sum\_caudalanteriorcingulate&diff\_precentral\\
						&diff\_isthmuscingulate&diff\_Amygdala\\
						&diff\_Cerebellum.White.Matter&\\
						&ratio\_isthmuscingulate&\\
					\botrule
				\end{tabular}
		\end{center}
		
	\end{minipage}
\end{table}

For CN vs. MCI detection, the RF, XGBoost, and LR base classifiers chose four features, the DT chose three features, the polynomial SVM chose eight features, and the radial SVM chose six features. Overall, the six ML methods chose 25 different features. Thus, in comparison to the CN vs. AD classification, the ML models show less agreement about the selected features. Consequently, the feature which was selected first in the forward selection process differed in five out of six methods. For the RF and the DT, the sum of the insular cortices was selected, the XGBoost classifier chose the ratio of the inferior parietal lobule, the polynomial SVM selected the sum of the lingual gyri, the SVM with the radial kernel chose the sum of the temporal pole volumes and the LR selected the sum of the left and right middle temporal gyri. Those features were previously associated with AD progression \cite{10.1136/gpsych-2018-100005,Foundas1997-ik,10.1016/j.neurobiolaging.2010.04.026,10.1371/journal.pcbi.1001006,10.1038/nrneurol.2009.215,10.1002/hbm.20934,10.1016/j.neurobiolaging.2003.12.007,10.1016/j.pscychresns.2011.06.014}. Similar to the CN vs. AD classification, all models selected at least one feature describing the asymmetry of the cortical and subcortical brain regions.

The forward feature selection results of the six ML models for MCI vs. AD classification are summarized in Table \ref{Table:ForwardSelectionMCIAD}. Four of the six models, namely RF, XGBoost, SVM poly, and LR selected five features. The DT chose six different features and the radial SVM selected two MRI features. Overall, the six methods selected 22 unique features. 

The most important features were the sum of the left and right hippocampi for the RF and DT model, the difference of the lateral ventricles for the XGBoost model, the sum of the inferior temporal gyri for both SVMs, and the sum of the entorhinal cortex volumes for the LR. Those features were previously associated with AD detection \cite{10.1038/nrneurol.2009.215,10.1002/hbm.20934,10.1016/j.neurobiolaging.2003.12.007,10.1016/j.pscychresns.2011.06.014,10.3233/jad-2011-101782,10.1136/jnnp.72.4.491}.

\begin{table}
	\caption{Features selected by forward selection using different ML methods as base classifiers for MCI vs. AD classification. Feature selection was exclusively used to reduce the number of MRI features.}\label{Table:ForwardSelectionMCIAD}
	\begin{minipage}{\linewidth}
		\begin{center}
				\begin{tabular}{ccc}
					\toprule LR&DT&RF\\\midrule
					sum\_entorhinal&sum\_Hippocampus&sum\_Hippocampus\\
					sum\_precuneus&sum\_cuneus&sum\_Amygdala\\
					sum\_VentralDC&sum\_posteriorcingulate&diff\_entorhinal\\
					diff\_frontalpole&ratio\_Putamen&sum\_isthmuscingulate\\
					diff\_rostralanteriorcingulate&sum\_Cortex&ratio\_lateralorbitofrontal\\
					&ratio\_parstriangularis&\\
										\midrule
					XGBoost&SVM poly & SVM radial\\\midrule
					diff\_Lateral.Ventricle&sum\_inferiortemporal&sum\_inferiortemporal\\
					diff\_Cortex&Brain.Stem&ratio\_frontalpole\\
					sum\_Cortex&sum\_entorhinal&\\
					sum\_pericalcarine&sum\_precuneus&\\
					sum\_precentral&ratio\_precuneus&\\
					\botrule
				\end{tabular}
		\end{center}
		
	\end{minipage}
\end{table}

The results of the forward selection for the sMCI vs. pMCI classification task are summarized in Table \ref{Table:ForwardSelectionsMCIpMCI}. Five features were selected by the RF model, the XGBoost model chose six features, the DT selected only one feature, both SVMs chose four features, and the LR selected three features. Overall, the six methods picked 19 unique features. Three methods, namely the RF, the DT, and the SVM with the radial kernel selected the sum of the left and right amygdalae as the most important feature. The forward selection with the XGBoost base model first picked the sum of the hippocampi. The polynomial SVM selected the sum of the left and right precuneus and the LR chose the sum of the inferior temporal gyri. Those features were previously associated with AD detection \cite{10.1038/nrneurol.2009.215,10.1002/hbm.20934,10.1016/j.neurobiolaging.2003.12.007,10.1016/j.pscychresns.2011.06.014,10.1136/jnnp.72.4.491,10.3389/fnagi.2018.00304}.

\begin{table}
	\caption{Features selected by forward selection using different ML methods as base classifiers for sMCI vs. pMCI classification. Feature selection was exclusively used to reduce the number of MRI features.}\label{Table:ForwardSelectionsMCIpMCI}
	\begin{minipage}{\linewidth}
		\begin{center}
				\begin{tabular}{ccc}
					\toprule LR&DT&RF\\\midrule
					sum\_inferiortemporal&sum\_Amygdala&sum\_Amygdala\\
					diff\_middletemporal&&sum\_inferiorparietal\\
					sum\_precentral&&sum\_entorhinal\\
					&&sum\_lateraloccipital\\
					&&diff\_superiorparietal\\
					\midrule
					XGBoost&SVM poly& SVM radial\\\midrule
					sum\_Hippocampus&sum\_precuneus&sum\_Amygdala\\
					diff\_lateralorbitofrontal&sum\_inferiortemporal&diff\_Inf.Lat.Vent\\
					Brain.Stem&sum\_rostralanteriorcingulate&sum\_precuneus\\
					sum\_caudalmiddlefrontal&ratio\_rostralanteriorcingulate&diff\_middletemporal\\
					diff\_precentral&&\\
					sum\_postcentral&&\\
					\botrule
				\end{tabular}
		\end{center}
		
	\end{minipage}
\end{table}

\subsection{Classification Tasks}
In the following the classification performances achieved for the four classification tasks are elaborated. The hyperparameters which reached the best accuracies during CV and which were thus used during training of the final models are summarized in Table \ref{Table:ResultsHyperparameters}.   
\begin{table}[h]\caption{Hyperparameters tuned for CN vs. AD, CN vs. MCI, MCI vs. AD, and sMCI vs. pMCI classification. Hyperparameters: LR: {C; penalty}, DT: {criterion; max\_depth; min\_samples\_split; splitter}, RF: {max\_features; min\_samples\_leaf; n\_estimators}, XGBoost: {colsample\_bytree; gamma; learning\_rate; max\_depth; min\_child\_weight; n\_estimators; subsample}, SVM poly: {C; degree; gamma}, SVM radial: {C; gamma}.}
	\label{Table:ResultsHyperparameters}
	\begin{center}\resizebox{\textwidth}{!}{
			\begin{tabular}{ll|llll}
				\toprule
				&Feature&\multicolumn{4}{c}{Hyperparameters}\\\
				&&CN vs. AD&CN vs. MCI&MCI vs. AD&sMCI vs. pMCI\\\midrule
				\multicolumn{6}{c}{FS-1}\\
				\midrule
				LR&yes&\{ 76.096;  \texttt{l2}\}&\{ 0.073;  \texttt{l2}\}&\{ 10.047;  \texttt{l2}\}&\{ 0.035;  \texttt{l2}\}\\
				LR&no&\{ 0.0297;  \texttt{l2}\}&\{ 0.034;  \texttt{l2}\}&\{ 0.029;  \texttt{l2}\}&\{ 0.039;  \texttt{l2}\}\\
				DT&yes&\{  100;  0.143;  \texttt{best}\}&\{  100;  0.375;  \texttt{best}\}&\{  49;  0.994;  \texttt{best}\}&\{  60;  0.236;  \texttt{random}\}\\
				DT&no&\{  49;  0.994;  \texttt{best}\}&\{  31;  0.476;  \texttt{best}\}&\{  94;  0.823;  \texttt{random}\}&\{  100;  0.461;  \texttt{best}\}\\
				RF&yes&\{ 5;  4;  1250\}&\{ 4;  8;  955\}&\{ 5;  1;  1250\}&\{ 2;  8;  251\}\\
				RF&no&\{ 77;  4;  1250\}&\{ 95;  1;  1250\}&\{ 28;  1;  250\}&\{ 71;  8;  1222\}\\
				XGBoost&yes&\{ 0.814;  3.551;  0.025;  8;  1.0;  459;  0.765\}&\{ 0.899;  0.660;  0.000;  13;  11.710;  488;  1.0\}&\{ 1.0;  20.0;  1.0;  20;  1.0;  500;  1.0\}&\{ 1.0;  0.0;  0.048;  14;  30.0;  334;  0.672\}\\
				XGBoost&no&\{ 0.924;  3.795;  0.202;  12;  10.938;  299;  1.0\}&\{ 0.671;  15.195;  0.000;  14;  7.070;  136;  0.967\}&\{ 0.244;  4.526;  0.010;  13;  10.171;  500;  0.508\}&\{ 0.934;  10.905;  0.003;  14;  6.850;  366;  0.485\}\\
				SVM poly&yes&\{ 962.766;  1;  \texttt{scale}\}&\{ 23.770;  1;  \texttt{auto}\}&\{ 8.965;  1;  \texttt{auto}\}&\{ 972.148;  1;  \texttt{scale}\}\\
				SVM poly&no&\{ 3.253;  1;  \texttt{auto}\}&\{ 1.481;  1;  \texttt{auto}\}&\{ 13.996;  3;  \texttt{auto}\}&\{ 13.996;  3;  \texttt{auto}\}\\
				SVM radial&yes&\{ 0.717;  \texttt{scale}\}&\{ 0.331;  \texttt{auto}\}&\{ 1.483;  \texttt{scale}\}&\{ 1.064;  \texttt{auto}\}\\
				SVM radial&no&\{ 1.772;  \texttt{scale}\}&\{ 1.685;  \texttt{scale}\}&\{ 1.144;  \texttt{scale}\}&\{ 0.647;  \texttt{auto}\}\\
				\midrule
				\multicolumn{6}{c}{FS-2}\\
				\midrule
				LR&yes&\{ 0.095;  \texttt{l2}\}&\{ 24.121;  \texttt{none}\}&\{ 0.083;  \texttt{l2}\}&\{ 0.055;  \texttt{l2}\}\\
				LR&no&\{ 0.013;  \texttt{l2}\}&\{ 0.082;  \texttt{l2}\}&\{ 0.020;  \texttt{l2}\}&\{ 0.017;  \texttt{l2}\}\\
				DT&yes&\{  75;  0.354;  \texttt{best}\}&\{  47;  0.098;  \texttt{best}\}&\{  49;  0.994;  \texttt{best}\}&\{  23;  0.432;  \texttt{best}\}\\
				DT&no&\{  49;  0.994;  \texttt{best}\}&\{  100;  0.487;  \texttt{best}\}&\{  100;  0.366;  \texttt{best}\}&\{  12;  0.325;  \texttt{random}\}\\
				RF&yes&\{ 2;  6;  1250\}&\{ 3;  11;  250\}&\{ 2;  1;  270\}&\{ 2;  6;  1248\}\\
				RF&no&\{ 53;  3;  1250\}&\{ 152;  1;  1250\}&\{ 56;  1;  1250\}&\{ 81;  1;  1227\}\\
				XGBoost&yes&\{ 0.885;  5.554;  0.012;  7;  3.119;  331;  0.296\}&\{ 0.995;  6.791; 0.000;  9;  15.455;  477;  0.875\}&\{ 1.0;  20.0;  1.0;  20;  1.0;  500;  1.0\}&\{ 1.0;  5.055;  0.000;  18;  7.984;  413;  0.592\}\\
				XGBoost&no&\{ 0.446;  1.499;  0.086;  10;  9.243;  361;  0.720\}&\{ 0.897;  8.254;  0.120;  14;  4.872;  112;  0.936\}&\{ 0.903;  11.976;  0.004;  5;  7.337;  376;  0.485\}&\{ 0.151;  9.387;  0.010;  8;  19.337;  305;  0.794\}\\
				SVM poly&yes&\{ 188.250;  1;  \texttt{auto}\}&\{ 1000.0;  1;  \texttt{auto}\}&\{ 1000.0;  1;  \texttt{scale}\}&\{ 13.721;  3;  \texttt{scale}\}\\
				SVM poly&no&\{ 13.996;  3;  \texttt{auto}\}&\{ 7.171;  1;  \texttt{auto}\}&\{ 184.588;  3;  \texttt{auto}\}&\{ 13.996;  3;  \texttt{auto}\}\\
				SVM radial&yes&\{ 2.526;  \texttt{scale}\}&\{ 0.727;  \texttt{auto}\}&\{ 1.343;  \texttt{auto}\}&\{ 0.375;  \texttt{auto}\}\\
				SVM radial&no&\{ 1.315;  \texttt{auto}\}&\{ 1.367;  \texttt{auto}\}&\{ 1.165;  \texttt{scale}\}&\{ 1.372;  \texttt{auto}\}\\
				\midrule
				\multicolumn{6}{c}{FS-3}\\
				\midrule
				LR&yes&\{ 26.861;  \texttt{l2}\}&\{ 0.586;  \texttt{l2}\}&\{ 0.375;  \texttt{l2}\}&\{ 0.021;  \texttt{l2}\}\\
				LR&no&\{ 4.893;  \texttt{l2}\}&\{ 0.609;  \texttt{l2}\}&\{ 0.0209;  \texttt{l2}\}&\{ 0.027;  \texttt{l2}\}\\
				DT&yes&\{  100;  0.010;  \texttt{best}\}&\{  69;  0.079;  \texttt{best}\}&\{  87;  0.146;  \texttt{best}\}&\{  100;  0.304;  \texttt{random}\}\\
				DT&no&\{  100;  0.010;  \texttt{best}\}&\{  69;  0.079;  \texttt{best}\}&\{  69;  0.079;  \texttt{best}\}&\{  6;  0.293;  \texttt{random}\}\\
				RF&yes&\{ 5;  1;  1236\}&\{ 4;  5;  1250\}&\{ 5;  5;  1126\}&\{ 2;  1;  1024\}\\
				RF&no&\{ 36;  1;  250\}&\{ 152;  20;  250\}&\{ 41;  1;  1250\}&\{ 43;  1;  1250\}\\
				XGBoost&yes&\{ 0.296;  19.279;  0.000;  14;  16.725;  445;  0.501\}&\{ 1.0;  20.0;  1.0;  20;  1.0;  500;  0.571\}&\{ 0.702;  3.777;  0.0013;  11;  5.021;  385;  0.230\}&\{ 0.702;  3.777;  0.001;  11;  5.021;  385;  0.229\}\\
				XGBoost&no&\{ 0.296;  19.279;  0.000;  14;  16.725;  445;  0.501\}&\{ 1.0;  2.266;  0.000;  20;  1.0;  500;  0.685\}&\{ 0.527;  13.650;  0.000;  19;  16.963;  254;  0.698\}&\{ 0.151;  9.387;  0.010;  8;  19.337;  305;  0.794\}\\
				SVM poly&yes&\{ 2.729;  1;  \texttt{auto}\}&\{ 12.554;  1;  \texttt{auto}\}&\{ 2.360;  1;  \texttt{auto}\}&\{ 1.013;  3;  \texttt{scale}\}\\
				SVM poly&no&\{ 1000.0;  1;  \texttt{auto}\}&\{ 62.015;  1;  \texttt{scale}\}&\{ 58.631;  3;  \texttt{auto}\}&\{ 13.996;  3;  \texttt{auto}\}\\
				SVM radial&yes&\{ 3.342;  \texttt{auto}\}&\{ 0.667;  \texttt{scale}\}&\{ 0.464;  \texttt{auto}\}&\{ 0.270;  \texttt{scale}\}\\
				SVM radial&no&\{ 10.047;  \texttt{auto}\}&\{ 10.047;  \texttt{auto}\}&\{ 1.341;  \texttt{auto}\}&\{ 1.611;  \texttt{auto}\}\\
				\botrule
		\end{tabular}}
	\end{center}
\end{table}
\subsubsection{CN vs. AD}
The results achieved for CN vs. AD classification are summarized in Table \ref{Table:ResultsCNAD_FS1}. The no information rates were 60.36~\% for the independent ADNI test set, 86.27~\% for AIBL, and 78.05~\% for OASIS. CN was the most frequent class in all datasets.

The best accuracy during CV of 99.68~\%~$\pm$~0.74 was achieved for the DT trained with feature selection and FS-3. This model also reached a perfect classification for the ADNI test set. All models trained for the CN vs. AD task reached accuracies higher than the no information rate for the ADNI dataset. The best AIBL accuracy of 95.94~\% was achieved for the XGBoost model trained for FS-3 and with feature selection. This model also reached the best F1-Score (91.48~\%) and the best MCC (0.830) for the AIBL dataset. The best balanced accuracies of 95.45~\% for the AIBL dataset were reached for both SVMs trained with feature selection and FS-3. The LR model trained with feature selection for FS-3 reached the best AIBL AUROC of 99.55~\%. Overall, two models achieved AIBL accuracies smaller than the no information rate of 86.27~\%. Those models were the DTs trained with feature selection for FS-1 and FS-2. 

The best OASIS accuracy of 90.58~\% was achieved for the polynomial SVM which was trained with feature selection and FS-3. For OASIS, four models achieved accuracies worse than the no information rate of 78.05~\%. Three of those models were trained on FS-1 and with feature selection, namely, the RF, the DT, and the SVM. The last model reaching an accuracy worse than the no information rate was the DT trained with FS-2 and feature selection. 

\begin{table}[h]	\caption{CV and test results for CN vs. AD classification. 
		All models were trained on ADNI and validated for an independent ADNI test set, and external AIBL and OASIS datasets. The best results in each section are highlighted in bold. No information rates: ADNI test set: 60.36~\%, AIBL dataset 86.27~\%, OASIS dataset: 78.05~\%
	}
	\label{Table:ResultsCNAD_FS1}
	\begin{center}
		\resizebox{\textwidth}{!}{
			\begin{tabular}{ll||r|rrrrr|rrrrr|rrrrr}
				\toprule
				&Feature&CV&\multicolumn{5}{c|}{ADNI}&\multicolumn{5}{c|}{AIBL}&\multicolumn{5}{c}{OASIS}\\
				Model&selection&ACC ($\bar{x}\pm\sigma$)
				&ACC&BACC&AUROC&F1&MCC&ACC&BACC&AUROC&F1&MCC&ACC&BACC&AUROC&F1&MCC\\
				&&(in \%)&(in \%)&(in \%)&(in \%)&(in \%)&&(in \%)&(in \%)&(in \%)&(in \%)&&(in \%)&(in \%)&(in \%)&(in \%)&\\\midrule
				\midrule
				\multicolumn{18}{c}{FS-1}\\
					\midrule
					LR&yes&$87.65\pm3.72$&85.80&84.91&92.54&85.08&0.702&90.72&83.96&91.09&81.69&0.637&78.94&75.80&82.16&72.59&0.466\\
					LR&no&$89.25\pm3.91$&89.35&87.85&94.85&88.61&0.777&88.97&\textbf{85.91}&\textbf{92.78}&80.22&0.621&81.60&79.50&85.22&75.99&0.534\\
					DT&yes&$84.56\pm4.32$&81.07&79.70&89.52&80.00&0.601&82.21&74.88&86.10&69.59&0.414&61.75&63.70&76.21&57.31&0.228\\
					DT&no&$82.21\pm4.35$&84.02&83.44&83.44&83.35&0.667&87.23&74.84&74.84&73.98&0.480&79.93&76.98&76.98&73.78&0.489\\
				RF&yes&$83.08\pm4.40$&86.39&85.40&89.01&85.67&0.714&87.81&76.95&86.46&75.57&0.513&68.96&67.05&75.52&62.68&0.292\\
				RF&no&$88.16\pm3.77$&\textbf{91.72}&\textbf{90.58}&\textbf{96.22}&\textbf{91.20}&\textbf{0.827}&91.30&83.70&91.86&82.36&0.648&83.37&80.45&85.64&77.73&0.563\\
				XGBoost&yes&$86.93\pm3.71$&88.76&87.61&93.93&88.09&0.764&89.17&80.70&91.32&78.64&0.576&78.05&76.50&81.28&72.26&0.469\\
				XGBoost&no&$86.99\pm3.90$&90.53&89.60&94.09&90.00&0.801&90.91&85.85&92.76&82.53&0.657&\textbf{84.15}&80.77&87.09&\textbf{78.47}&\textbf{0.576}\\
				SVM poly&yes&$87.79\pm3.85$&82.84&80.15&91.94&81.19&0.639&90.72&78.04&90.95&79.42&0.590&80.71&75.12&82.33&73.52&0.474\\
				SVM poly&no&$\mathbf{89.63\pm3.58}$&\textbf{91.72}&90.320&94.81&91.14&0.828&86.46&85.05&92.40&77.25&0.577&80.38&78.35&84.06&74.64&0.510\\
				SVM radial&yes&$88.67\pm3.60$&88.76&87.36&93.28&88.02&0.764&\textbf{91.49}&84.41&91.94&\textbf{82.84}&\textbf{0.658}&76.72&73.29&80.28&70.03&0.418\\
				SVM radial&no&$88.88\pm3.59$&89.94&88.34&95.30&89.21&0.790&89.17&85.43&91.53&80.29&0.620&82.04&\textbf{80.87}&\textbf{86.55}&76.82&0.555\\
					\midrule
				\multicolumn{18}{c}{FS-2}\\
					\midrule
					LR&yes&$88.11\pm3.39$&87.57&86.12&95.02&86.76&0.738&92.26&87.23&92.49&84.74&0.698&82.04&77.60&84.54&75.58&0.517\\
					LR&no&$\mathbf{89.93\pm3.84}$&89.35&87.08&\textbf{96.52}&88.37&0.782&90.33&86.70&\textbf{93.44}&82.06&0.652&\textbf{84.15}&81.13&86.11&\textbf{78.61}&0.579\\
					DT&yes&$84.44\pm4.47$&81.07&79.70&88.88&80.00&0.601&82.21&74.88&85.68&69.59&0.414&61.75&63.70&75.36&57.31&0.228\\
					DT&no&$82.21\pm4.35$&84.02&83.44&83.44&83.35&0.667&87.23&74.84&74.84&73.98&0.480&79.93&76.98&76.98&73.78&0.489\\
				RF&yes&$84.84\pm3.91$&86.98&85.37&92.99&86.08&0.726&89.56&77.96&89.90&77.96&0.559&81.37&76.45&80.02&74.59&0.497\\
				RF&no&$88.55\pm3.68$&\textbf{90.53}&89.08&\textbf{96.52}&89.88&\textbf{0.802}&91.68&85.11&92.21&83.33&0.668&83.15&80.31&86.08&77.50&0.559\\
				XGBoost&yes&$87.42\pm3.86$&89.94&88.34&94.67&89.21&0.790&90.52&81.48&92.59&80.57&0.612&82.37&78.54&85.47&76.21&0.531\\
				XGBoost&no&$88.32\pm3.48$&\textbf{90.53}&\textbf{89.34}&95.08&\textbf{89.94}&0.801&90.72&\textbf{89.88}&93.08&83.42&0.687&83.70&\textbf{81.76}&\textbf{87.43}&78.46&\textbf{0.581}\\
				SVM poly&yes&$87.15\pm3.54$&85.80&84.14&94.09&84.82&0.701&\textbf{92.65}&86.86&91.98&\textbf{85.18}&\textbf{0.705}&82.04&77.60&83.69&75.58&0.517\\
				SVM poly&no&$81.55\pm4.10$&81.07&76.38&92.51&77.78&0.624&91.88&72.20&91.92&77.93&0.608&83.92&70.83&85.56&73.30&0.483\\
				SVM radial&yes&$89.26\pm3.38$&82.25&80.17&91.35&80.89&0.624&91.30&84.89&91.94&82.74&0.657&78.38&74.35&81.63&71.59&0.444\\
				SVM radial&no&$89.78\pm3.48$&88.76&86.85&96.22&87.86&0.766&90.14&85.99&92.59&81.61&0.642&82.04&80.33&87.16&76.63&0.548\\
					\midrule
				\multicolumn{18}{c}{FS-3}\\
					\midrule
					LR&yes&$99.45\pm0.90$&99.41&99.25&\textbf{100.00}&99.38&0.988&92.65&94.56&\textbf{99.55}&86.99&0.762&89.58&83.34&90.18&84.33&0.688\\
					LR&no&$98.50\pm1.41$&97.63&97.27&99.94&97.51&0.951&94.58&95.08&99.37&89.85&0.808&89.58&84.79&90.63&84.79&0.696\\
					DT&yes&$\mathbf{99.68\pm0.74}$&\textbf{100.00}&\textbf{100.00}&\textbf{100.00}&\textbf{100.00}&\textbf{1.000}&94.39&92.60&92.60&89.11&0.788&88.58&77.80&77.80&81.12&0.641\\
					DT&no&$98.92\pm1.18$&98.82&98.76&98.76&98.76&0.975&94.39&92.60&92.60&89.11&0.788&88.58&77.62&77.62&81.03&0.641\\
				RF&yes&$99.56\pm0.85$&99.41&99.25&\textbf{100.00}&99.38&0.988&95.16&93.05&98.52&90.41&0.811&88.69&78.60&88.57&81.60&0.646\\
				RF&no&$99.34\pm0.99$&98.82&98.51&\textbf{100.00}&98.76&0.975&95.16&93.05&98.35&90.41&0.811&89.91&81.19&91.60&83.91&0.688\\
				XGBoost&yes&$99.06\pm1.24$&98.82&98.51&\textbf{100.00}&98.76&0.975&\textbf{95.94}&91.72&98.41&\textbf{91.48}&\textbf{0.830}&90.35&81.84&92.50&84.61&0.702\\
				XGBoost&no&$99.01\pm1.22$&98.22&97.76&\textbf{100.00}&98.13&0.963&95.36&91.98&98.30&90.54&0.812&90.02&81.81&\textbf{92.80}&84.25&0.693\\
				SVM poly&yes&$99.65\pm0.73$&99.41&99.25&\textbf{100.00}&99.38&0.988&94.20&\textbf{95.45}&99.52&89.34&0.801&\textbf{90.58}&84.34&91.23&\textbf{85.69}&\textbf{0.716}\\
				SVM poly&no&$98.51\pm1.53$&97.63&97.27&99.85&97.51&0.951&93.23&93.71&99.27&87.62&0.768&87.80&84.38&89.71&82.93&0.661\\
				SVM radial&yes&$99.54\pm0.80$&98.82&98.51&99.96&98.76&0.975&94.20&\textbf{95.45}&99.04&89.34&0.801&86.14&81.50&88.01&80.440&0.610\\
				SVM radial&no&$97.83\pm1.55$&98.82&98.51&99.87&98.76&0.975&94.97&94.72&99.00&90.390&0.815&87.25&\textbf{85.12}&92.33&82.64&0.659\\
				
				\botrule
		\end{tabular}}
	\end{center}
\end{table}

\subsubsection{CN vs. MCI}

The results achieved for CN vs. MCI classification are summarized in Table \ref{Table:Results_CNMCI_FS-1}. The no information rate for this task was 62.64~\% for the ADNI test set, 82.44~\% for AIBL, and 97.39~\% for OASIS. MCI was the most frequent class in the ADNI dataset, whereas, for AIBL and OASIS, CN was.

The results achieved for CN vs. MCI classification were worse than those for the CN vs. AD task. 
The best accuracy during CV of $90.21~\%\pm2.72$ was achieved for the XGBoost model trained for FS-3 and without feature selection. 
The best accuracy for the ADNI test set was 91.58~\% reached for two models. Both models, the radial SVM and the XGBoost model were trained for FS-3 and with feature selection. The latter model also reached the best ADNI balanced accuracy and ADNI F1-Score. Overall, none of the models reached an ADNI accuracy worse than the no information rate of 62.51~\%. 

The results achieved for AIBL and OASIS were worse than the ADNI results. The best AIBL accuracy was 68.95~\% achieved for two DTs trained with forward feature selection for FS-1 and FS-2. These models also reached the best AIBL balanced accuracies, AIBL F1-Scores, and AIBL MCCs. 

The best OASIS accuracy of 55.05~\% was reached for the DT trained without feature selection for FS-3. For the CN vs. MCI classification, all models achieved accuracies worse than the no information rates for OASIS and AIBL.

\begin{table}[h]\caption{CV and test results for the CN vs. MCI classification. 
		All models were trained on ADNI and validated for an independent ADNI test set, and external AIBL and OASIS datasets. The best results in each section are highlighted in bold. No information rates: ADNI test set: 62.63~\%, AIBL dataset: 82.44~\%, OASIS dataset: 97.39~\%
	}
	\label{Table:Results_CNMCI_FS-1}
	\begin{center}
		\resizebox{\textwidth}{!}{
			\begin{tabular}{ll||r|rrrrr|rrrrr|rrrrr}
				\toprule
				&Feature&CV&\multicolumn{5}{c|}{ADNI}&\multicolumn{5}{c|}{AIBL}&\multicolumn{5}{c}{OASIS}\\
				Model&selection&ACC ($\bar{x}\pm\sigma$)&
				ACC&BACC&AUROC&F1&MCC&ACC&BACC&AUROC&F1&MCC&ACC&BACC&AUROC&F1&MCC\\
				&&(in \%)&(in \%)&(in \%)&(in \%)&(in \%)&&(in \%)&(in \%)&(in \%)&(in \%)&&(in \%)&(in \%)&(in \%)&(in \%)&\\\midrule
		\midrule
		\multicolumn{18}{c}{FS-1}\\
		\midrule
		LR&yes&$65.20\pm3.46$&65.93&59.16&65.14&58.74&0.218&42.88&57.07&62.40&41.54&0.115&17.98&52.76&61.51&16.51&0.024\\
		LR&no&$67.46\pm3.74$&69.96&65.74&76.74&66.21&0.335&38.63&56.98&67.21&38.17&0.121&24.76&58.80&\textbf{62.69}&21.69&0.068\\
		
		DT&yes&$65.87\pm2.96$&66.67&60.14&71.54&59.92&0.238&43.99&55.68&56.77&42.09&0.090&23.51&58.16&45.56&20.79&0.064\\
		DT&no&$69.18\pm4.39$&73.63&\textbf{73.80}&75.03&\textbf{72.75}&\textbf{0.463}&\textbf{68.95}&\textbf{66.67}&64.78&\textbf{60.25}&\textbf{0.265}&\textbf{46.47}&57.15&48.21&\textbf{34.42}&0.046\\
		RF&yes&$66.80\pm4.36$&63.74&58.79&68.01&58.86&0.189&53.23&57.55&60.89&48.37&0.115&31.81&52.18&44.74&26.04&0.015\\
		RF&no&$\mathbf{71.08\pm3.99}$&73.63&70.84&79.77&71.22&0.427&51.20&62.95&69.77&48.54&0.201&32.64&\textbf{62.85}&57.22&27.06&\textbf{0.089}\\
		XGBoost&yes&$63.64\pm3.32$&64.47&56.60&66.72&55.23&0.169&33.09&52.38&53.08&32.97&0.044&19.50&48.42&43.22&17.56&-0.013\\
		XGBoost&no&$68.97\pm4.45$&\textbf{73.99}&71.13&\textbf{81.11}&71.55&0.434&58.04&66.27&\textbf{72.64}&53.79&0.248&33.20&60.57&61.29&27.29&0.073\\
		SVM poly&yes&$67.42\pm4.17$&67.40&62.70&75.35&63.01&0.273&41.22&56.89&63.35&40.25&0.114&29.74&56.24&57.97&24.93&0.044\\
		SVM poly&no&$66.25\pm3.53$&65.57&60.25&73.40&60.35&0.225&37.71&56.42&66.12&37.33&0.112&22.96&55.32&61.80&20.30&0.042\\
		SVM radial&yes&$67.01\pm3.08$&64.84&56.50&70.28&54.73&0.174&41.77&59.72&65.23&41.05&0.164&14.11&55.89&51.92&13.43&0.059\\
		SVM radial&no&$67.02\pm4.25$&67.40&62.90&74.17&63.22&0.275&36.41&57.29&64.64&36.26&0.133&21.44&57.10&62.68&19.25&0.058\\
		\midrule
		\multicolumn{18}{c}{FS-2}\\
		\midrule
		LR&yes&$65.40\pm3.59$&68.86&63.08&70.58&63.34&0.297&49.54&59.87&68.03&46.74&0.153&23.37&55.53&64.30&20.61&0.043\\
		LR&no&$67.68\pm4.39$&69.60&65.45&78.20&65.89&0.327&40.67&57.80&68.53&39.93&0.131&25.59&56.67&60.96&22.19&0.050\\
		DT&yes&$64.87\pm4.31$&71.79&68.58&72.67&69.00&0.384&58.60&61.63&65.10&52.70&0.177&35.96&56.87&49.86&28.75&0.046\\
		DT&no&$69.14\pm4.43$&73.63&\textbf{73.80}&75.03&\textbf{72.75}&\textbf{0.463}&\textbf{68.95}&\textbf{66.67}&64.78&\textbf{60.25}&\textbf{0.265}&\textbf{46.47}&57.15&48.21&\textbf{34.42}&0.046\\
		RF&yes&$68.25\pm4.18$&66.30&61.63&73.66&61.88&0.249&58.78&60.09&65.46&52.25&0.154&36.51&59.72&50.76&29.23&0.065\\
		RF&no&$\mathbf{71.09\pm4.25}$&\textbf{73.99}&71.92&77.90&72.05&0.441&52.13&64.34&69.49&49.47&0.222&34.44&\textbf{61.21}&56.94&28.08&\textbf{0.077}\\
		XGBoost&yes&$64.95\pm4.07$&69.60&62.48&72.93&62.32&0.310&41.22&58.55&61.65&40.46&0.144&23.93&55.82&46.43&21.01&0.045\\
		XGBoost&no&$69.60\pm3.82$&72.16&68.48&\textbf{81.40}&69.01&0.387&46.40&61.69&\textbf{70.63}&44.90&0.188&26.56&54.61&51.47&22.77&0.034\\
		SVM poly&yes&$67.01\pm4.03$&72.53&69.56&77.21&69.95&0.402&43.62&57.52&65.77&42.16&0.121&32.09&52.33&59.15&26.21&0.016\\
		SVM poly&no&$67.22\pm4.05$&69.23&65.35&77.06&65.75&0.322&39.19&57.32&66.21&38.66&0.126&24.07&53.33&57.38&21.02&0.026\\
		SVM radial&yes&$68.55\pm3.64$&71.79&66.21&76.08&66.79&0.368&50.09&62.28&69.82&47.64&0.191&20.19&51.34&52.36&18.16&0.011\\
		SVM radial&no&$67.77\pm4.40$&67.40&63.10&75.17&63.42&0.277&35.67&56.02&66.78&35.51&0.110&20.33&53.97&\textbf{64.37}&18.34&0.033\\
		\midrule
		\multicolumn{18}{c}{FS-3}\\
		\midrule
		LR&yes&$88.48\pm3.08$&91.21&90.21&97.50&90.53&0.811&\textbf{66.36}&\textbf{76.70}&\textbf{88.47}&\textbf{62.01}&\textbf{0.406}&50.62&54.16&59.69&36.17&0.027\\
		LR&no&$87.89\pm3.22$&90.84&89.13&96.19&89.98&0.803&58.23&72.18&87.64&55.36&0.341&49.24&53.45&62.04&35.47&0.022\\
		DT&yes&$88.90\pm3.28$&89.01&86.88&95.83&87.89&0.763&59.15&73.15&80.46&56.22&0.355&46.47&49.47&59.30&33.83&-0.003\\
		DT&no&$88.92\pm3.22$&87.91&86.59&95.32&86.95&0.740&58.60&71.99&80.30&55.58&0.337&\textbf{55.05}&53.87&62.10&\textbf{38.14}&0.025\\
		RF&yes&$88.94\pm3.03$&89.74&87.46&97.51&88.64&0.780&61.00&74.27&87.26&57.75&0.371&46.33&51.96&61.03&33.96&0.013\\
		RF&no&$90.04\pm2.74$&88.28&86.09&96.18&87.08&0.747&58.04&72.07&85.10&55.21&0.339&50.35&51.46&60.55&35.82&0.009\\
		XGBoost&yes&$88.48\pm2.95$&\textbf{91.58}&\textbf{90.31}&96.65&\textbf{90.87}&0.819&57.12&71.92&82.96&54.53&0.339&44.26&50.89&58.47&32.85&0.006\\
		XGBoost&no&$\mathbf{90.21\pm2.72}$&89.74&88.05&95.38&88.81&0.779&57.86&71.95&81.46&55.06&0.338&51.04&51.81&59.88&36.16&0.012\\
		SVM poly&yes&$88.77\pm3.28$&91.21&89.42&\textbf{97.59}&90.36&0.811&62.85&74.57&88.18&59.09&0.374&50.07&51.32&59.65&35.68&0.008\\
		SVM poly&no&$87.33\pm3.27$&89.74&88.05&95.59&88.81&0.779&59.33&72.44&86.97&56.19&0.344&51.31&51.95&60.51&36.30&0.013\\
		SVM radial&yes&$87.83\pm3.13$&\textbf{91.58}&89.32&96.58&90.65&\textbf{0.822}&61.74&74.72&86.05&58.36&0.377&43.85&\textbf{58.36}&60.20&33.17&\textbf{0.054}\\
		SVM radial&no&$85.87\pm2.93$&87.55&85.71&94.51&86.41&0.731&52.31&69.83&86.00&50.66&0.315&40.53&56.66&\textbf{65.05}&31.30&0.044\\
		
			\botrule
		\end{tabular}}
	\end{center}
\end{table}

\subsubsection{MCI vs. AD}
The MCI vs. AD classification results are summarized in Table \ref{Table:Results_MCIAD_FS1}. The no information rate was 71.85~\% for the ADNI test set, 57.23~\% for AIBL, and 91.24~\% for OASIS. The most frequent class was MCI for ADNI and AIBL as well as AD for OASIS.

The best CV accuracy of 89.39~\%~$\pm$~2.99 was achieved for the RF trained without feature selection and FS-3. For the independent ADNI test set, the best accuracy was 88.66~\%, reached by the RF and LR models trained with feature selection and FS-3. The first of those models also reached the best ADNI AUROC of 95.50~\%, whereas the second model achieved the best ADNI F1-Score (85.50~\%), and ADNI MCC (0.712). None of the models reached an ADNI accuracy worse than the no information rate. However, the DT trained without feature selection for FS-1 as well as the XGBoost and the DT both trained for FS-2 and with feature selection exactly achieved the no information rate of 71.85~\% for the independent ADNI test set. The two DT models mentioned predicted the MCI class for all subjects and thus represented random classifiers. 

The best AIBL accuracy was 84.94~\% reached by the RF model trained without feature selection and FS-3. This model also reached the best AIBL balanced accuracy (83.64~\%), AIBL F1-Score (84.24~\%), and AIBL MCC (0.693). Except for the previously mentioned random classifiers, all models outperformed the no information rate for the AIBL dataset which was 57.23~\%. The performances for OASIS were worse than those achieved for ADNI and AIBL. The best OASIS accuracy of 57.14~\% was achieved for the radial SVM trained without feature selection and FS-3. The random classifiers achieved a worse accuracy of 8.76~\% for OASIS. These accuracies correspond to the ratio of MCI subjects in the dataset. MCI was the most frequent class for the ADNI dataset and the rarest class for OASIS. All models achieved OASIS accuracies worse than the no information rate of 91.24~\%.

\begin{table}[h]\caption{CV and test results for MCI vs. AD classification.  
		All models were trained on ADNI and validated for an independent ADNI test set, and external AIBL and OASIS datasets. The best results in each section are highlighted in bold. No information rates: ADNI test set: 71.85~\%, AIBL dataset: 57.23~\%, OASIS dataset: 91.24~\%
	}
	\label{Table:Results_MCIAD_FS1}
	\begin{center}
		\resizebox{\textwidth}{!}{
			\begin{tabular}{ll||r|rrrrr|rrrrr|rrrrr}
			\toprule
				&Feature&CV&\multicolumn{5}{c|}{ADNI}&\multicolumn{5}{c|}{AIBL}&\multicolumn{5}{c}{OASIS}\\
				Model&selection&ACC ($\bar{x}\pm\sigma$)&
				ACC&BACC&AUROC&F1&MCC&ACC&BACC&AUROC&F1&MCC&ACC&BACC&AUROC&F1&MCC\\
				&&(in \%)&(in \%)&(in \%)&(in \%)&(in \%)&&(in \%)&(in \%)&(in \%)&(in \%)&&(in \%)&(in \%)&(in \%)&(in \%)&\\\midrule
				\midrule\multicolumn{18}{c}{FS-1}\\\midrule
				LR&yes&$74.65\pm3.24$&73.95&60.54&74.51&61.32&0.268&63.25&57.93&73.14&53.83&0.241&29.95&61.62&65.18&28.85&\textbf{0.161}\\
				LR&no&$75.60\pm3.11$&78.15&\textbf{69.82}&81.99&71.03&0.428&\textbf{65.66}&\textbf{62.53}&73.70&\textbf{62.08}&0.281&46.08&58.56&67.38&39.41&0.098\\
				DT&yes&$71.79\pm0.42$&71.85&50.00&69.09&41.81&0.000&57.23&50.00&64.49&36.40&0.000&8.76&50.00&59.50&8.05&0.000\\
				DT&no&$73.29\pm3.44$&74.37&66.28&66.28&66.91&0.341&64.46&60.76&60.76&59.71&0.254&\textbf{54.84}&60.98&60.98&\textbf{44.68}&0.124\\
				RF&yes&$74.05\pm3.62$&74.79&63.39&80.36&64.53&0.313&62.65&57.58&71.35&54.00&0.217&32.72&55.99&64.51&30.47&0.077\\
				RF&no&$\mathbf{75.79\pm3.53}$&77.31&65.60&\textbf{82.27}&67.23&0.379&65.06&60.58&73.83&58.50&0.276&37.79&61.15&62.19&34.59&0.136\\
				XGBoost&yes&$73.62\pm3.79$&73.11&61.77&74.57&62.64&0.270&62.05&56.52&63.97&51.64&0.208&34.56&\textbf{61.76}&64.85&32.38&0.150\\
				XGBoost&no&$75.55\pm3.39$&\textbf{78.99}&69.49&81.51&\textbf{71.16}&\textbf{0.440}&65.06&60.22&\textbf{74.90}&57.50&\textbf{0.283}&40.09&57.66&68.63&35.65&0.091\\
				SVM poly&yes&$73.95\pm2.34$&76.05&57.92&79.96&56.75&0.325&62.05&55.63&74.66&47.68&0.260&21.66&54.69&69.11&21.46&0.077\\
				SVM poly&no&$73.35\pm2.83$&72.27&54.83&74.17&53.17&0.160&61.45&55.64&73.65&49.78&0.195&29.49&56.61&\textbf{71.00}&28.12&0.089\\
				SVM radial&yes&$73.03\pm2.51$&76.89&60.77&73.10&61.32&0.349&62.05&56.35&61.79&50.94&0.213&28.57&51.34&62.01&26.98&0.018\\
				SVM radial&no&$75.72\pm3.13$&76.89&65.31&76.98&66.85&0.368&64.46&60.05&70.17&58.02&0.258&41.94&53.91&64.59&36.21&0.045\\
				\midrule\multicolumn{18}{c}{FS-2}\\\midrule
				LR&yes&$75.01\pm3.40$&75.63&62.62&74.80&63.81&0.320&63.25&58.11&73.42&54.44&0.237&26.73&59.85&67.49&26.10&0.145\\
				LR&no&$\mathbf{75.93\pm3.13}$&\textbf{78.99}&\textbf{70.86}&81.65&\textbf{72.14}&\textbf{0.451}&63.86&\textbf{60.59}&74.16&\textbf{59.92}&0.239&43.32&59.42&68.45&\textbf{37.92}&0.110\\
				DT&yes&$71.79\pm0.42$&71.85&50.00&69.09&41.81&0.000&57.23&50.00&64.49&36.40&0.000&8.76&50.00&59.50&8.05&0.000\\
				DT&no&$72.61\pm4.05$&74.37&62.65&77.75&63.71&0.298&60.24&55.66&67.58&52.78&0.147&35.02&\textbf{62.01}&70.39&32.74&\textbf{0.153}\\
				RF&yes&$74.45\pm3.68$&78.15&66.64&79.32&68.44&0.404&60.24&54.77&71.88&49.70&0.148&31.80&57.87&66.17&29.98&0.103\\
				RF&no&$75.59\pm3.40$&78.15&67.09&\textbf{83.04}&68.86&0.407&63.25&58.82&72.13&56.60&0.227&43.32&59.42&65.91&\textbf{37.92}&0.110\\
				XGBoost&yes&$73.23\pm3.65$&71.85&59.08&73.82&59.61&0.218&60.24&54.41&69.84&48.21&0.152&30.41&59.49&68.20&29.06&0.128\\
				XGBoost&no&$74.79\pm3.26$&74.37&60.38&82.71&61.10&0.274&63.25&57.75&74.97&53.18&0.247&33.64&61.26&\textbf{74.80}&31.65&0.145\\
				SVM poly&yes&$75.17\pm3.01$&78.15&63.92&80.00&65.53&0.395&63.25&57.58&\textbf{75.49}&52.49&0.254&24.88&54.08&70.47&24.24&0.060\\
				SVM poly&no&$73.59\pm2.94$&72.69&55.12&73.52&53.45&0.174&60.24&54.41&74.37&48.21&0.152&29.49&56.61&73.18&28.12&0.089\\
				SVM radial&yes&$72.68\pm2.74$&76.47&62.29&73.61&63.46&0.338&\textbf{65.06}&59.69&66.99&55.79&\textbf{0.301}&29.95&52.10&62.55&28.08&0.027\\
				SVM radial&no&$75.54\pm3.44$&76.47&65.47&77.38&66.90&0.362&62.05&57.59&70.66&55.18&0.195&\textbf{43.78}&54.92&66.06&37.46&0.057\\
				\midrule\multicolumn{18}{c}{FS-3}\\\midrule
				LR&yes&$88.26\pm2.96$&\textbf{88.66}&84.39&94.40&\textbf{85.50}&\textbf{0.712}&82.53&81.89&90.63&82.06&0.642&49.31&72.22&81.05&43.61&0.256\\
				LR&no&$87.38\pm3.06$&85.71&82.34&93.63&82.34&0.647&82.53&81.53&89.58&81.90&0.641&52.53&\textbf{73.99}&81.31&45.90&\textbf{0.273}\\
				DT&yes&$87.40\pm2.98$&87.82&\textbf{86.07}&94.89&85.27&0.706&83.13&82.42&85.51&82.64&0.654&50.69&72.98&76.42&44.59&0.263\\
				DT&no&$86.29\pm3.61$&81.93&79.26&89.27&78.33&0.569&81.33&80.66&86.03&80.82&0.617&51.15&70.85&75.12&44.53&0.238\\
				RF&yes&$88.81\pm2.94$&\textbf{88.66}&83.94&\textbf{95.50}&85.35&0.711&83.13&82.24&88.07&82.56&0.654&49.77&72.47&80.30&43.94&0.258\\
				RF&no&$\mathbf{89.39\pm2.99}$&85.71&80.98&94.71&81.83&0.638&\textbf{84.94}&\textbf{83.64}&87.49&\textbf{84.24}&\textbf{0.693}&50.23&70.35&78.99&43.88&0.233\\
				XGBoost&yes&$88.21\pm3.24$&86.97&81.86&95.41&83.18&0.667&81.93&80.65&88.22&81.14&0.629&47.00&70.96&81.77&41.96&0.244\\
				XGBoost&no&$88.51\pm2.94$&86.13&81.27&95.11&82.28&0.648&81.33&80.13&87.86&80.56&0.616&48.85&71.97&82.26&43.28&0.253\\
				SVM poly&yes&$88.62\pm2.75$&86.97&82.77&94.65&83.52&0.671&81.33&80.66&\textbf{91.06}&80.82&0.617&51.61&73.48&81.39&45.25&0.268\\
				SVM poly&no&$79.79\pm2.87$&78.99&65.86&87.82&67.88&0.424&73.49&70.26&86.88&70.36&0.465&38.25&66.16&\textbf{83.21}&35.47&0.200\\
				SVM radial&yes&$88.32\pm3.17$&88.24&83.19&94.06&84.73&0.699&80.12&78.54&87.15&79.08&0.592&48.39&71.72&70.81&42.95&0.251\\
				SVM radial&no&$86.02\pm3.14$&83.19&78.32&90.40&78.84&0.577&80.12&78.90&88.35&79.30&0.591&\textbf{57.14}&69.38&77.80&\textbf{47.75}&0.219\\
				\bottomrule
				
		\end{tabular}}
	\end{center}
\end{table}

\subsubsection{sMCI vs. pMCI}
The results reached for sMCI vs. pMCI classification with no information rates of 55.56~\% for the ADNI test set and 57.14~\% for AIBL are summarized in Table \ref{Table:ResultssMCIpMCI_FS1}. As previously mentioned, due to the lack of available data, OASIS was not used for this comparison.

The best CV accuracy of $70.75~\%~\pm~5.94$ was achieved for the RF trained with feature selection and FS-3. For the independent ADNI test set, the radial SVM which was trained with forward feature selection for FS-3 reached the best accuracy (75.00~\%), balanced accuracy (74.38~\%), F1-Score (74.51~\%), and MCC (0.491). The best AUROC of 80.14~\% was reached for the XGBoost model trained with forward feature selection for FS-3.  None of the models achieved worse results than the ADNI no information rate of 55.56~\%. The best AIBL accuracy was 82.14~\% reached for the radial SVM which was trained for FS-3, and without feature selection. Five models achieved AIBL accuracies worse than the no information rate. Those models were all DTs trained with FS-1 and FS-3, and the LR trained with feature selection and FS-3.

\begin{table}[h]	\caption{CV and test results for sMCI vs. pMCI classification. 
		All models were trained on ADNI and validated for an independent ADNI test set and external AIBL dataset. The best results in each section are highlighted in bold. No information rates: ADNI test set: 55.56~\%, AIBL dataset: 57.14~\%
	}
	\label{Table:ResultssMCIpMCI_FS1}
	\begin{center}
		\resizebox{\textwidth}{!}{
			\begin{tabular}{ll||r|rrrrr|rrrrr}
			\toprule
				&&CV&\multicolumn{5}{c|}{ADNI}&\multicolumn{5}{c}{AIBL}\\
				Model&Feature selection&ACC ($\bar{x}\pm\sigma$)&
				ACC&BACC&AUROC&F1&MCC&ACC&BACC&AUROC&F1&MCC\\
				&&(in \%)&(in \%)&(in \%)&(in \%)&(in \%)&&(in \%)&(in \%)&(in \%)&(in \%)&\\
				\midrule\multicolumn{13}{c}{FS-1}\\\midrule
				LR&yes&$62.21\pm5.47$&65.28&63.59&67.68&63.47&0.287&64.29&59.38&54.17&56.25&0.265\\
				LR&no&$67.69\pm5.81$&66.67&64.84&74.22&64.70&0.316&67.86&64.58&59.38&64.15&0.333\\
				DT&yes&$64.73\pm5.33$&68.06&66.56&69.79&66.61&0.346&53.57&51.04&47.92&50.48&0.022\\
				DT&no&$64.92\pm5.84$&68.06&67.19&70.12&67.29&0.348&53.57&51.04&45.83&50.48&0.022\\
				RF&yes&$68.30\pm5.90$&68.75&67.66&71.13&67.78&0.361&67.86&63.54&59.38&61.99&0.350\\
				RF&no&$\mathbf{68.64\pm6.21}$&70.14&69.38&\textbf{76.41}&69.49&0.392&67.86&63.54&53.65&61.99&0.350\\
				XGBoost&yes&$65.50\pm5.78$&68.75&67.66&73.45&67.78&0.361&57.14&57.29&58.85&56.92&0.144\\
				XGBoost&no&$68.31\pm5.91$&\textbf{70.83}&\textbf{69.84}&76.07&\textbf{70.00}&\textbf{0.405}&57.14&54.17&54.17&53.33&0.091\\
				SVM poly&yes&$64.80\pm5.57$&59.03&57.19&65.51&56.76&0.152&57.14&51.04&51.04&42.86&0.040\\
				SVM poly&no&$64.33\pm5.01$&61.81&58.59&70.00&56.33&0.213&64.29&60.42&63.54&59.06&0.251\\
				SVM radial&yes&$66.30\pm5.85$&65.28&63.59&70.21&63.47&0.287&67.86&63.54&54.17&61.99&0.350\\
				SVM radial&no&$67.58\pm5.35$&68.75&67.34&75.43&67.43&0.360&\textbf{78.57}&\textbf{75.00}&\textbf{66.67}&\textbf{75.44}&\textbf{0.603}\\
				\midrule\multicolumn{13}{c}{FS-2}\\\midrule
				LR&yes&$65.60\pm5.68$&62.50&61.09&69.63&61.03&0.230&60.71&57.29&71.88&56.19&0.167\\
				LR&no&$68.24\pm5.48$&69.44&67.81&76.05&67.86&0.376&71.43&67.71&68.23&67.25&0.427\\
				DT&yes&$67.29\pm5.43$&65.97&63.59&72.14&62.97&0.304&67.86&65.62&55.47&65.71&0.331\\
				DT&no&$65.93\pm5.69$&61.81&61.56&68.02&61.49&0.230&64.29&65.62&66.41&64.29&0.312\\
				RF&yes&$\mathbf{70.14\pm6.24}$&67.36&66.25&70.66&66.35&0.332&71.43&69.79&66.67&70.05&0.409\\
				RF&no&$69.01\pm5.53$&\textbf{70.14}&\textbf{69.06}&\textbf{77.54}&\textbf{69.21}&\textbf{0.390}&67.86&64.58&57.81&64.15&0.333\\
				XGBoost&yes&$66.70\pm5.77$&68.75&\textbf{69.06}&74.15&68.68&0.379&60.71&60.42&69.79&60.26&0.207\\
				XGBoost&no&$68.77\pm5.60$&\textbf{70.14}&\textbf{69.06}&76.31&\textbf{69.21}&\textbf{0.390}&64.29&60.42&59.38&59.06&0.251\\
				SVM poly&yes&$60.59\pm5.53$&57.64&56.25&58.09&56.10&0.129&64.29&61.46&71.35&61.11&0.251\\
				SVM poly&no&$64.83\pm4.78$&66.67&63.91&71.02&62.88&0.325&67.86&65.62&66.15&65.71&0.331\\
				SVM radial&yes&$68.89\pm5.47$&65.28&63.91&70.37&63.91&0.288&71.43&68.75&66.15&68.89&0.411\\
				SVM radial&no&$68.12\pm5.85$&69.44&68.12&76.23&68.24&0.375&\textbf{75.00}&\textbf{72.92}&\textbf{72.40}&\textbf{73.33}&\textbf{0.486}\\
				\midrule\multicolumn{13}{c}{FS-3}\\\midrule
				LR&yes&$69.79\pm5.97$&74.31&73.44&79.63&73.63&0.476&60.71&58.33&63.54&58.10&0.177\\
				LR&no&$69.55\pm5.40$&71.53&70.31&79.63&70.49&0.419&71.43&67.71&64.58&67.25&0.427\\
				DT&yes&$68.70\pm5.85$&74.31&73.28&79.72&73.51&0.476&53.57&53.12&44.53&53.03&0.062\\
				DT&no&$66.83\pm5.47$&68.06&66.41&75.28&66.40&0.346&53.57&51.04&43.75&50.48&0.022\\
				RF&yes&$\mathbf{70.75\pm5.94}$&71.53&71.09&77.02&71.13&0.423&60.71&59.38&63.02&59.42&0.190\\
				RF&no&$70.64\pm5.76$&70.83&70.00&79.60&70.14&0.405&67.86&64.58&56.25&64.15&0.333\\
				XGBoost&yes&$69.11\pm5.78$&73.61&72.97&\textbf{80.14}&73.09&0.463&53.57&53.12&54.69&53.03&0.062\\
				XGBoost&no&$69.36\pm5.58$&73.61&72.50&78.48&72.72&0.462&64.29&61.46&52.60&61.11&0.251\\
				SVM poly&yes&$66.94\pm5.32$&64.58&62.50&72.83&62.08&0.271&60.71&58.33&58.33&58.10&0.177\\
				SVM poly&no&$66.35\pm4.77$&66.67&64.06&73.75&63.23&0.323&64.29&62.50&66.67&62.57&0.258\\
				SVM radial&yes&$70.09\pm5.46$&\textbf{75.00}&\textbf{74.38}&79.11&\textbf{74.51}&\textbf{0.491}&64.29&62.50&61.98&62.57&0.258\\
				SVM radial&no&$70.68\pm5.19$&68.06&66.72&79.00&66.80&0.346&\textbf{82.14}&\textbf{80.21}&\textbf{73.96}&\textbf{80.95}&\textbf{0.640}\\
				\botrule
		\end{tabular}}
	\end{center}
\end{table}

\subsection{Feature Sets}
\begin{figure}
	\centering
	\begin{subfigure}[b]{0.49\textwidth}
		\centering
		\includegraphics[width=\textwidth]{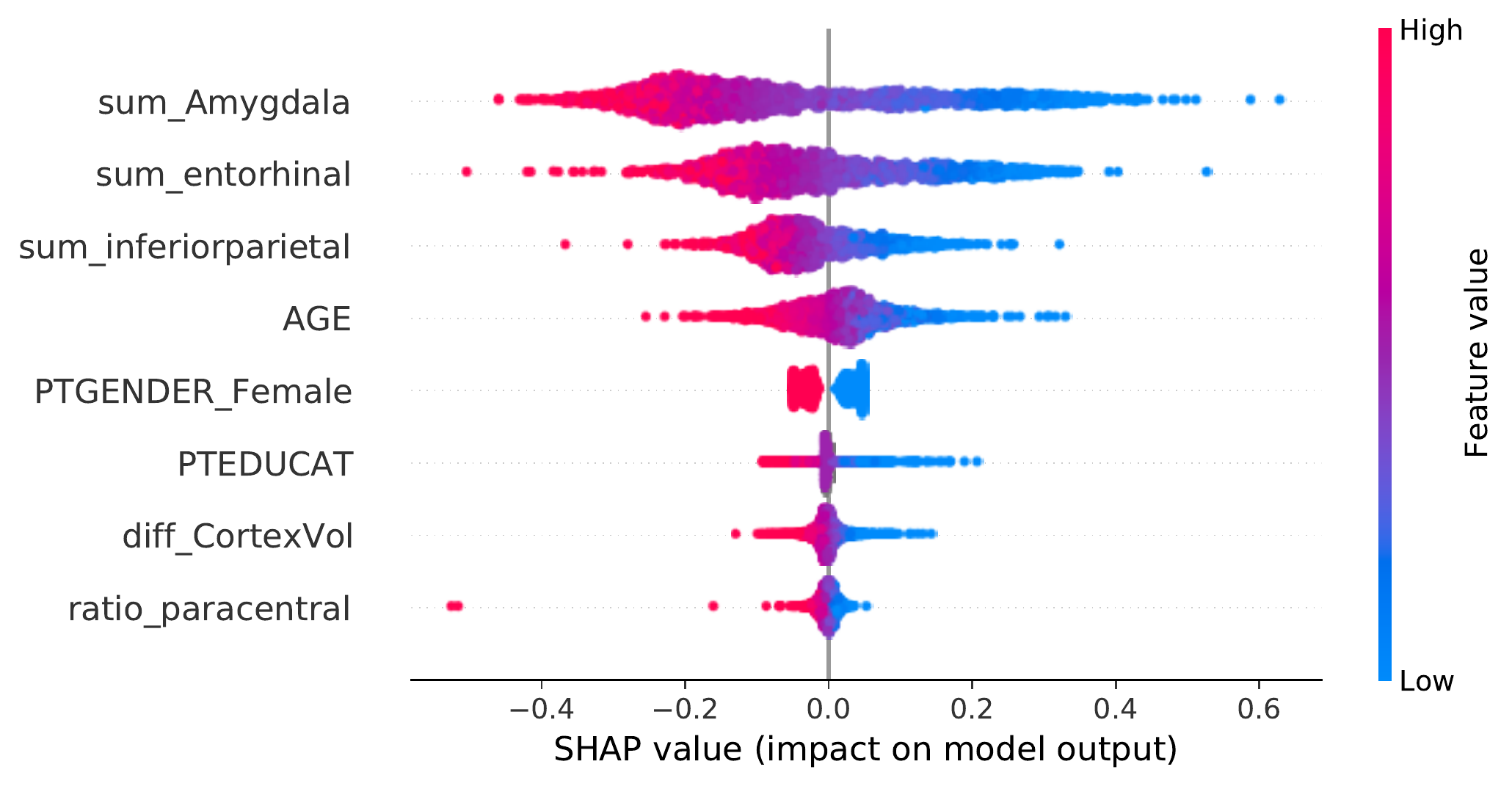}
		\caption{FS-1}
		\label{fig:CNAD_FS-1_svmPoly_rfMean_SHAP}
	\end{subfigure}
	\hfill
	\begin{subfigure}[b]{0.49\textwidth}
		\centering
		\includegraphics[width=\textwidth]{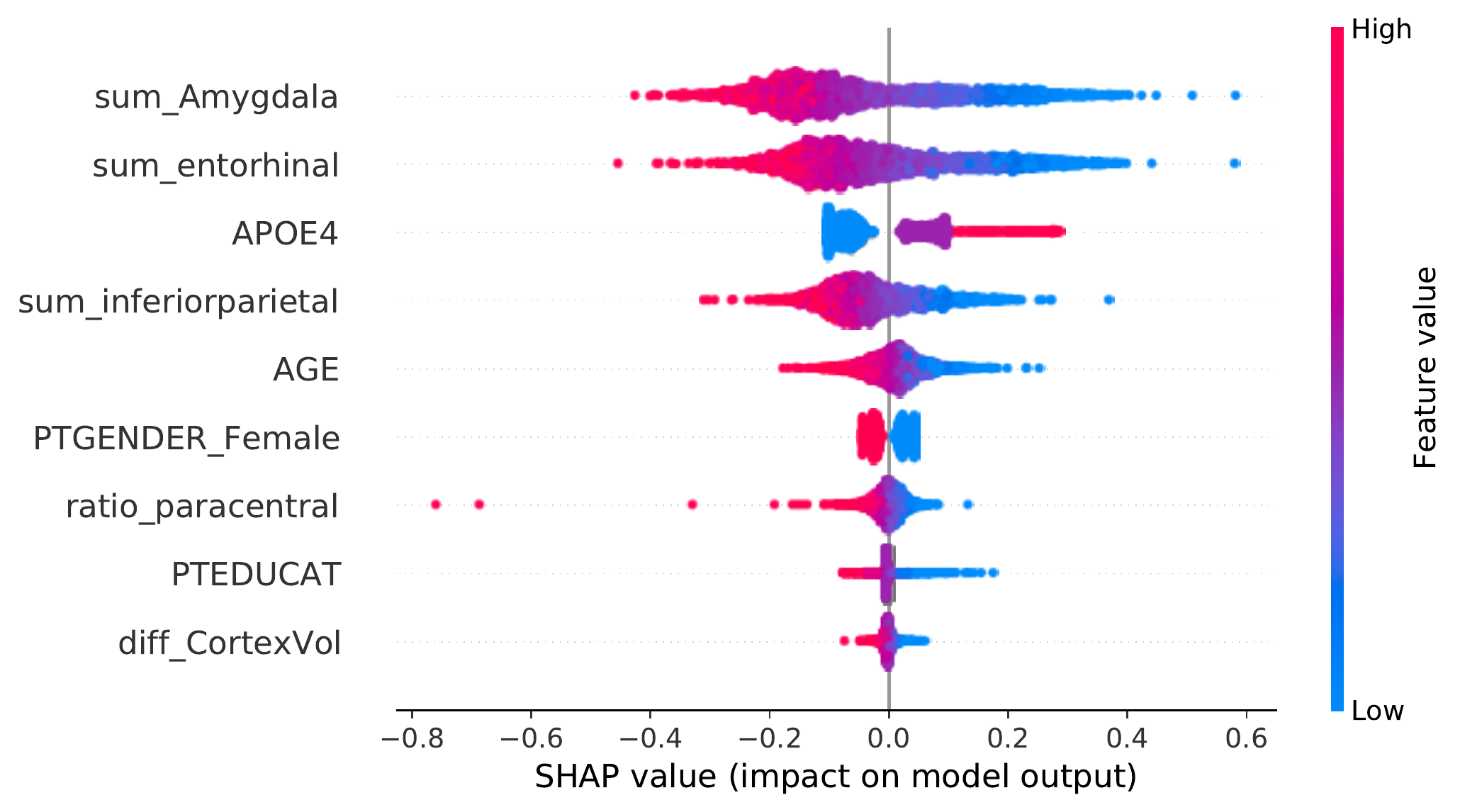}
		\caption{FS-2}
		\label{fig:CNAD_FS-2_svmPoly_rfMean_SHAP}
	\end{subfigure}\\\vspace{4pt} 
	\begin{subfigure}[b]{0.49\textwidth}
		\centering
		\includegraphics[width=\textwidth]{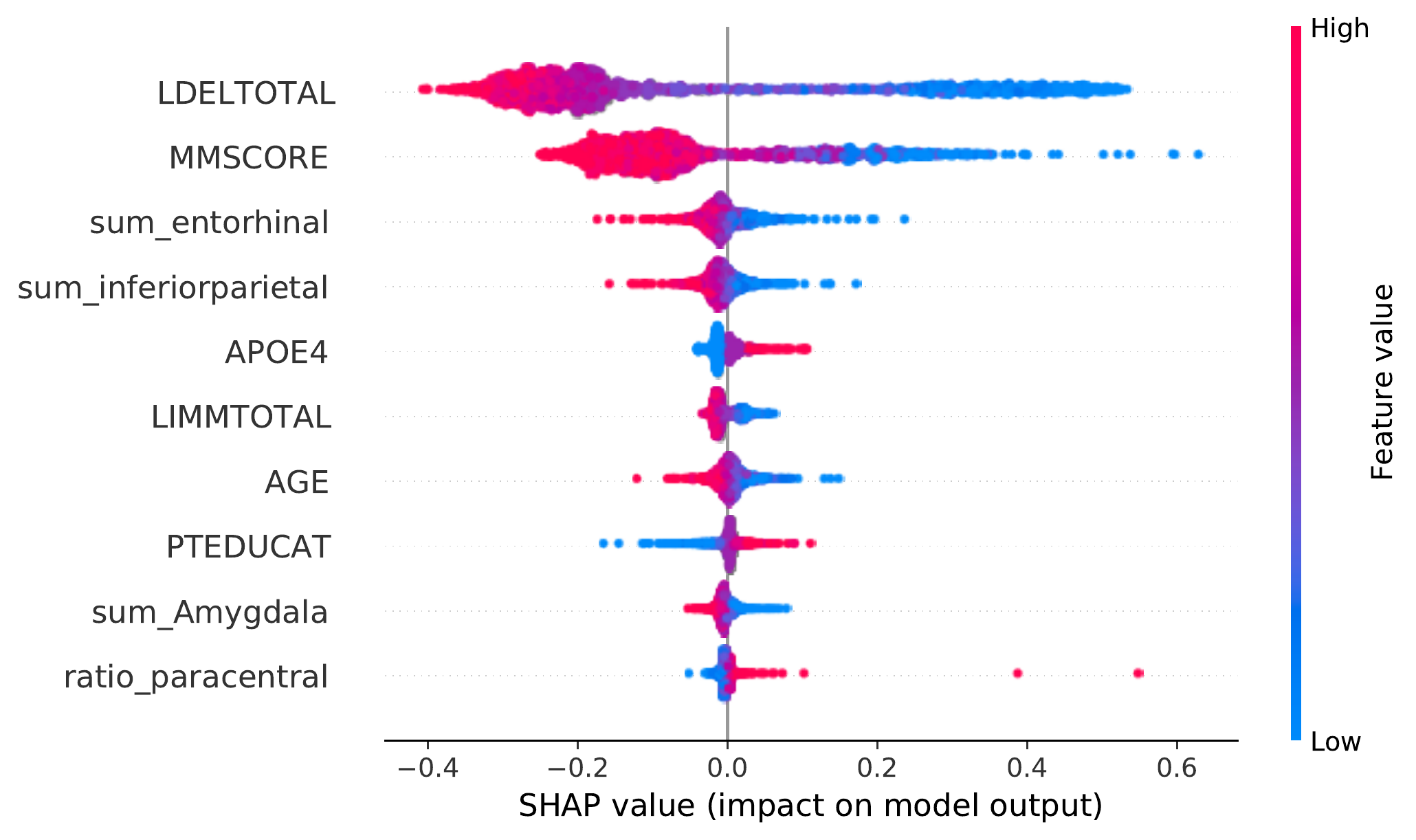}
		\caption{FS-3}
		\label{fig:CNAD_FS-3_svmPoly_rfMean_SHAP}
	\end{subfigure}\hfill
	\caption{SHAP summary plots of the polynomial SVM trained with feature selection for CN vs. AD. The plots visualize the Shapley values of \textit{n=2,266} subjects from the ADNI, AIBL, and OASIS datasets and the ten most important model features. Each subject is represented by a dot. The colors decode the subject's feature expression. High feature expressions are colored in red and small expressions in blue. The model learned that features with high Shapley values increased the patient's AD risk. Each plot shows a model trained on a different feature set}
	\label{fig:FeatureSet_comp_CNAD_svmPoly_rfMean_SHAP}
\end{figure}

As can be seen in Table \ref{Table:ResultsCNAD_FS1}, for CN vs. AD classification, all models achieved the best scores using FS-3. Thus, adding cognitive test results to the dataset improved the overall classification results. 
The SHAP summary plots for the polynomial SVMs trained with feature selection for all three feature sets are shown in Figure \ref{fig:FeatureSet_comp_CNAD_svmPoly_rfMean_SHAP}. SHAP summary plots \cite{Lundberg2018} explain the predictions for the subjects of the entire ADNI, AIBL, and OASIS datasets. Each point plots a Shapley value for a subject and a feature and is colored depending on the feature value. The vertical axis represented the features, ordered by the mean absolute Shapley values and their distribution. The higher a Shapley value is, the more the feature expression increases the probability the model classifies the subject as an AD subject. All SHAP summary plots were limited to the top ten features.

Following the biological processes of AD, small brain volumes~\cite{10.1016/j.dadm.2015.11.002,10.1371/journal.pone.0066367,10.1038/s41598-018-29295-9,10.1038/nrneurol.2009.215}, large ventricular volumes~\cite{10.1016/j.neuroimage.2004.03.040,10.1212/01.WNL.0000110315.26026.EF}, the presence of ApoE$\epsilon$4 alleles \cite{10.1126/science.8346443,10.1073/pnas.90.5.1977,10.1136/jnnp.2010.231555}, and bad performances in cognitive test scores were expected to have a pathogenic effect on the disease progression. The left hemisphere was expected to be more affected by atrophy than the right \cite{10.3390/s21030778} one. However, some investigations for MCI subjects also showed the right hippocampus was more affected \cite{10.1002/hbm.23772}. This asymmetry occurs primarily in the hippocampi and amygdalae \cite{10.1093/brain/aww243,10.1016/j.neuroimage.2015.01.032}.
For FS-1 and FS-2, the most important feature was the sum of the left and right amygdalae. Consistently with the previously mentioned atrophy patterns \cite{10.1016/j.dadm.2015.11.002,10.1371/journal.pone.0066367,10.1038/s41598-018-29295-9,10.1038/nrneurol.2009.215}, small volumes of the amygdalae (colored in blue) increased the probability the model classifies a subject as an AD subject. Large amygdala volumes (colored in red) were associated with CN subjects. The second most important feature for both models was the sum of the left and right entorhinal cortex. The model, trained using FS-1, learned that large volumes (colored in red) of the amygdalae, the entorhinal cortices, and the inferior parietal lobules had protective effects on the disease progression (negative Shapley values). Those assocications correspond to previous research \cite{10.1016/j.dadm.2015.11.002,10.1371/journal.pone.0066367,10.1038/s41598-018-29295-9,10.1038/nrneurol.2009.215}. The model additionally learned that a large difference between the left and right cortex volume (colored in red) was associated with CN. Large differences were reached if the volume of the left hemisphere was larger than the right one. The same observation applies to the ratio of the left and right paracentral lobules. Considering the sociodemographic features show, that the model learned, young age (colored in blue) was associated with disease progression. However, the summary of the ADNI dataset in Table \ref{Table:ADNIBaseline} shows the mean age of the CN group was younger than the AD group. Additionally, the model learned that females (colored in blue) and subjects with low education more likely develop AD. 

FS-2 added the number of ApoE$\epsilon$4 alleles to FS-1. The additional feature was the third most important feature of this model. The model learned that a large number of ApoE$\epsilon$4 alleles (no ApoE$\epsilon$4 alleles are colored in blue, one ApoE$\epsilon$4 allele is colored in purple, two ApoE$\epsilon$4 alleles are colored in red) led to an increased risk. Previous research also identified ApoE$\epsilon$4 as an AD risk factor \cite{10.1126/science.8346443,10.1073/pnas.90.5.1977,10.1136/jnnp.2010.231555}. Biologically plausible associations \cite{10.1016/j.dadm.2015.11.002,10.1371/journal.pone.0066367,10.1038/s41598-018-29295-9,10.1038/nrneurol.2009.215} were noted for the summed volumes of the left and right amygdalae, the entorhinal cortices, and the inferior parietal lobules. The ratio of the left and right paracentral lobule volumes showed an increased risk of AD if the left hemisphere was smaller than the right one. The same applies to the differences in cortex volumes.

FS-3 added the results of three cognitive tests to FS-2. The LDELTOTAL cognitive test score achieved the best feature importance for this model, followed by the MMSCORE. The LIMMTOTAL score achieved the sixth-best feature importance. For all cognitive test scores, the model associated good scores (colored in red) with healthy subjects. The third most important feature was the summed volume of the entorhinal cortices. Consistently with AD atrophy, small volumes (colored in blue) were associated with AD progression. The same applied to the sum of the inferior parietal lobules and the amygdalae. Similar to the FS-1 and FS-2 models, the FS-3 model also learned young age (colored in blue) was associated with AD, although, the mean age of the ADNI-CN group was younger than the mean age of the ADNI-AD group. 

As can be seen in Table \ref{Table:Results_CNMCI_FS-1}, for the CN vs. MCI classification, FS-3 outperformed FS-1 and FS-2 for the ADNI and AIBL performance scores. The best accuracies for OASIS were reached for FS-3, whereas FS-2 models outperform those models for F1-Score, balanced accuracy, AUROC, and MCC.

For the MCI vs. AD task, which is summarized in Table \ref{Table:Results_MCIAD_FS1}, the same applied to all ADNI and AIBL scores. For the OASIS dataset, the best accuracy and F1-Score were reached by FS-1 and the best balanced accuracy, AUROC, and MCC for FS-3.

The results for the sMCI vs. pMCI classification are shown in Table \ref{Table:ResultssMCIpMCI_FS1}. Those results show that FS-3 outperformed FS-1 and FS-2 for all ADNI scores. For the AIBL dataset, the best accuracy, balanced accuracy, F1-Score, and MCC were achieved for FS-2, whereas the best AUROC was reached for FS-3. 

To indicate whether the differences in ADNI test accuracies between the three feature sets are statistically significant, a Friedman test \cite{10.1080/01621459.1937.10503522} ($p-value < 0.05$) was executed. For this investigation, the results of Table \ref{Table:ForwardSelectionCNAD}, Table \ref{Table:ForwardSelectionCNMCI}, Table \ref{Table:ForwardSelectionMCIAD}, and Table \ref{Table:ForwardSelectionsMCIpMCI} are summarized, resulting in 48 observations per feature set (six different models, two feature selection methods, and four tasks). The p-value of $2.2\cdot 10^{-16}$ indicated statistically significant differences between the feature sets. To identify, which feature sets differed from each other, a pairwise Wilcoxon signed rank test ($p-value < 0.05$) with Bonferroni adjustment was executed. A summary of the results is given in Table \ref{Table:Wilcox_FS_Comp}. The results FS-3 achieved significantly differed from FS-1 and FS-2. The FS-1 and FS-2 results showed no statistically significant differences. 

\begin{table}[h]	\caption{P-values of the pairwise Wilcoxon signed rank test ($p-value < 0.05$) with Bonferroni adjustment to compare the differences in ADNI test accuracies between the three feature sets. Statistically significant results are highlighted in bold}
	\label{Table:Wilcox_FS_Comp}
	\begin{center}
			\begin{tabular}{r|rrr}
				\toprule
				&FS-1&FS-2&FS-3\\\midrule
				FS-1&-&-&-\\
				FS-2&0.95&-&-\\
				FS-3&$\mathbf{<0.001}$&$\mathbf{<0.001}$&-\\
				\botrule
		\end{tabular}
	\end{center}
\end{table}


\subsection{Reproducibility}
In this work, all models were trained using the ADNI dataset. Data from AIBL and OASIS subjects were used to test model reproducibility. 

\begin{figure}
	\includegraphics[width=1\textwidth]{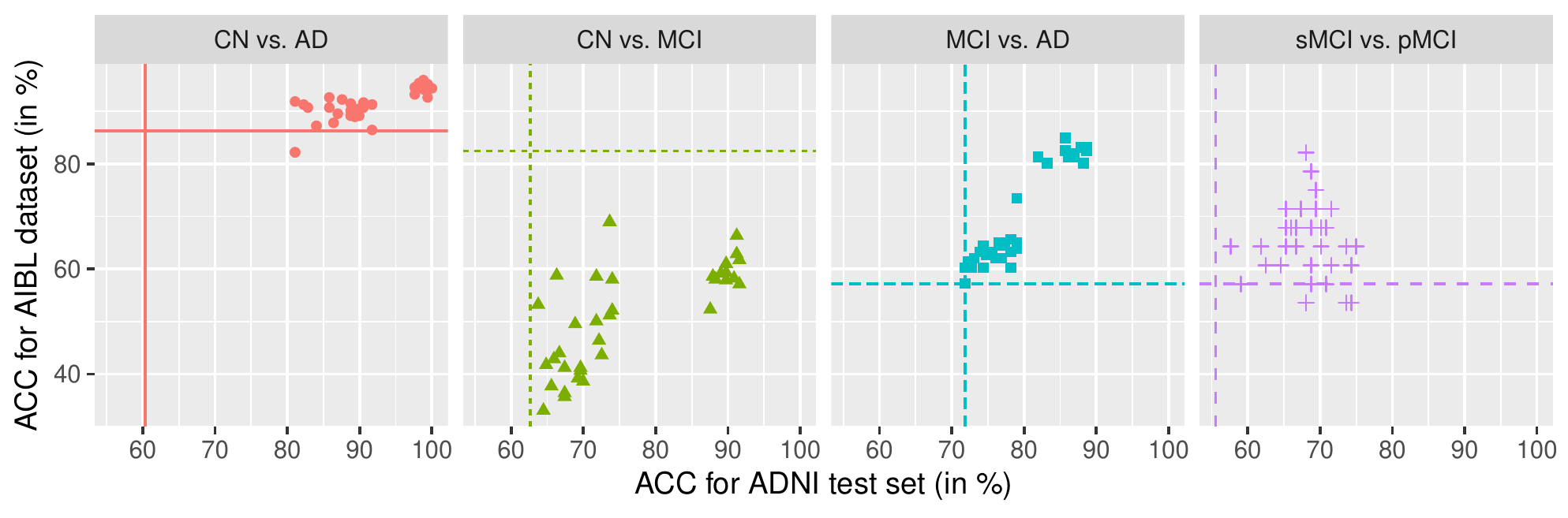}
	\caption{Plot showing the accuracies achieved for the independent ADNI test set and the AIBL dataset for all models described in Table \ref{Table:ResultsCNAD_FS1}, Table \ref{Table:Results_CNMCI_FS-1}, Table \ref{Table:Results_MCIAD_FS1}, and Table \ref{Table:ResultssMCIpMCI_FS1}. The no information rates for all classification tasks are visualized as horizontal lines for AIBL and as vertical lines for ADNI}
	\label{fig:CorrelationAIBL} 
\end{figure}

For all classification tasks, most models achieved worse results for AIBL and OASIS in comparison to the independent ADNI test set. 
The AIBL accuracies are plotted against the ADNI accuracies for all previously described models in Figure \ref{fig:CorrelationAIBL}. Overall, the AIBL accuracies were worse than those achieved for ADNI. The CN vs. AD classification models achieved the best accuracies for ADNI and AIBL. The worst AIBL accuracies were achieved for CN vs. MCI classification, where all models reached AIBL accuracies worse than the no information rate. For the remaining classification tasks, most models reached AIBL accuracies better than the no information rate. For the sMCI vs. pMCI classification, no correlation between ADNI and AIBL accuracies was observed. 

\begin{figure}
	\includegraphics[width=1\textwidth]{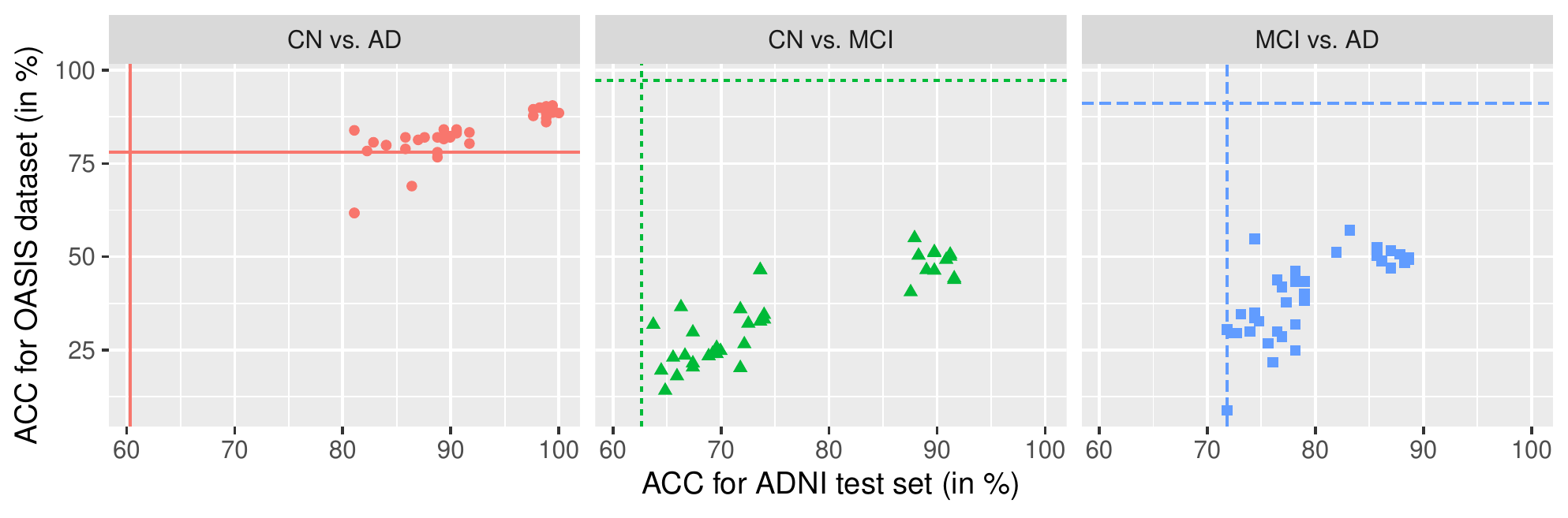}
	\caption{Plot showing the accuracies achieved for the independent ADNI test set and the OASIS dataset for all models described in Table \ref{Table:ResultsCNAD_FS1}, Table \ref{Table:Results_CNMCI_FS-1}, Table \ref{Table:Results_MCIAD_FS1}, and Table \ref{Table:ResultssMCIpMCI_FS1}. The no information rates for all classification tasks are visualized as horizontal lines for OASIS and as vertical lines for ADNI}
	\label{fig:CorrelationOASIS} 
\end{figure}

In Figure \ref{fig:CorrelationOASIS}, the OASIS accuracies of all previously described models are plotted against their ADNI accuracies. Similar to the AIBL results, the overall OASIS accuracies were worse than those achieved for ADNI. The best results for OASIS were achieved for CN vs. AD classification. Those models mainly reached accuracies better than the no information rate. The OASIS no information rates for the remaining classification tasks were larger than 90~\% and all classification models trained for the ADNI dataset performed worse. However, the most frequent classes in OASIS and ADNI differed from each other for those classification tasks. For the OASIS dataset, the worst accuracy was achieved for MCI vs. AD classification. Reasons for the worse OASIS performances were, for example, differing MRI protocols and differences in the subject selection process.

\begin{figure}
	\centering
	\begin{subfigure}[b]{0.49\textwidth}
		\centering
		\includegraphics[width=\textwidth]{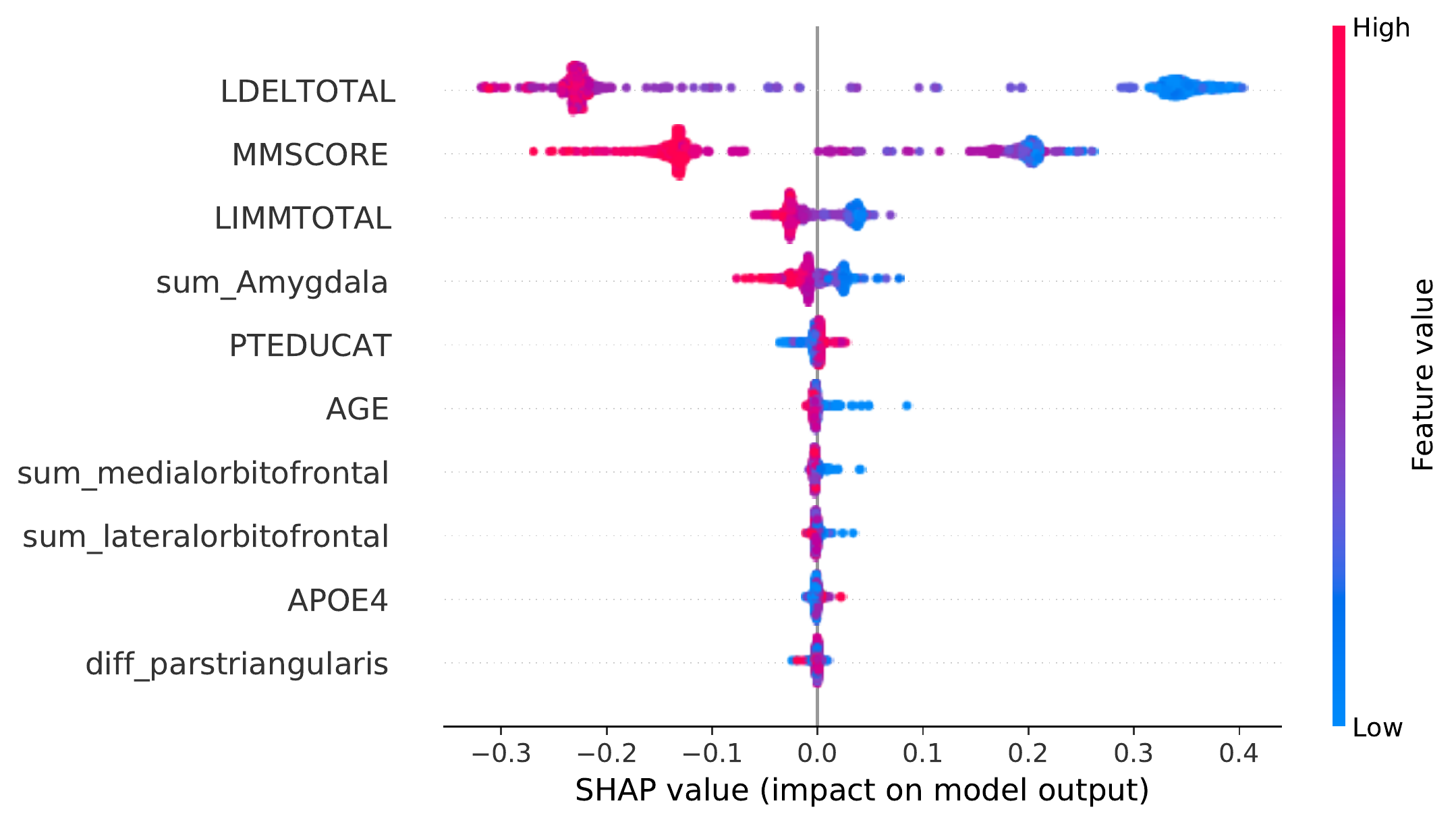}
		\caption{ADNI training and test set (\textit{n=847})}
		\label{fig:CNAD_FS-3_RF_RFMean_SHAP}
	\end{subfigure}
	\hfill
	\begin{subfigure}[b]{0.49\textwidth}
		\centering
		\includegraphics[width=\textwidth]{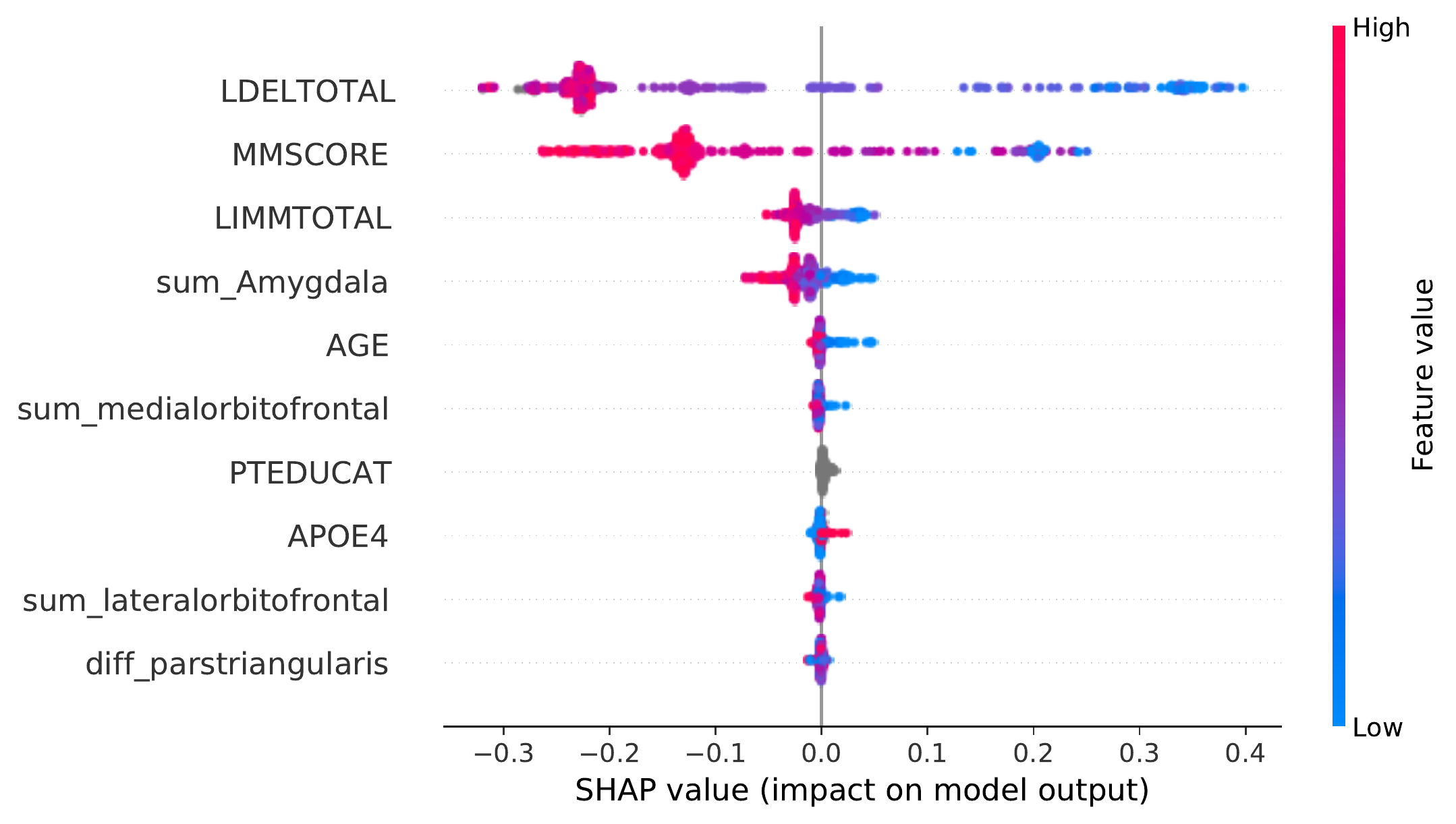}
		\caption{AIBL dataset (\textit{n=517})}
		\label{fig:CNAD_FS-3_RF_RFMean_SHAP_AIBL}
	\end{subfigure}\\\vspace{4pt} 
	
	\begin{subfigure}[b]{0.49\textwidth}
		\centering
		\includegraphics[width=\textwidth]{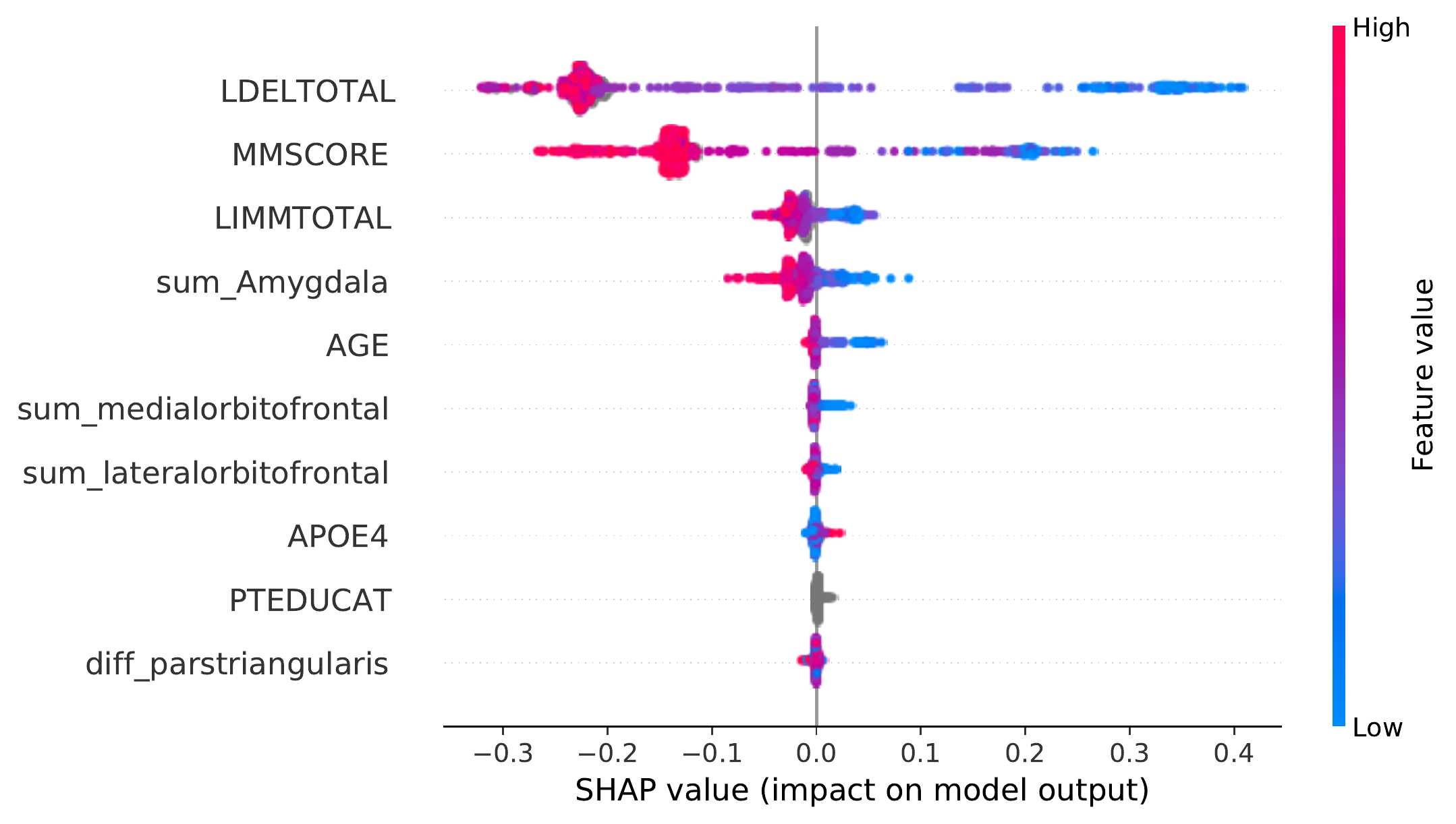}
		\caption{OASIS dataset (\textit{n=902})}
		\label{fig:CNAD_FS-3_RF_RFMean_SHAP_OASIS}
	\end{subfigure}
	\caption{SHAP summary plots of the RF trained with FS-3 and feature selection for CN vs. AD classification. The plots visualize the Shapley values of subjects from the ADNI, AIBL, and OASIS datasets and the ten most important model features. Each subject is represented by a dot. The colors decode the feature values of the subject. High feature values are colored in red whereas small feature values are colored in blue. The model learned that features with high Shapley values increased the patient's risk to develop AD. Each plot shows the results on a different dataset	
	}
	\label{fig:Reproducibility_Comparison_RF_CNAD_SHAP}
\end{figure}

To compare the model predictions for the three datasets, SHAP summary plots were visualized for the individual datasets in Figure \ref{fig:Reproducibility_Comparison_RF_CNAD_SHAP}. Those plots show the Shapley values for the RF trained with feature selection and FS-3, which was trained for CN vs. AD classification. For all three datasets, the three most important features were the cognitive test scores, and bad scores were associated with disease progression. Those cognitive test scores were followed by the volumetric features, of the amygdalae, medial orbitofrontal cortices, and pars triangularis, as well as the AGE, the number of APOE$\epsilon$4 alleles, and the number of education years in slightly differing orders. For all volumetric features, biologically plausible associations \cite{10.1016/j.dadm.2015.11.002,10.1371/journal.pone.0066367,10.1038/s41598-018-29295-9,10.1038/nrneurol.2009.215} were learned. The number of education years was not available in AIBL and OASIS, those scores are therefore colored in grey.

\begin{figure}
	\centering
	\begin{subfigure}[b]{0.49\textwidth}
		\centering
		\includegraphics[width=\textwidth]{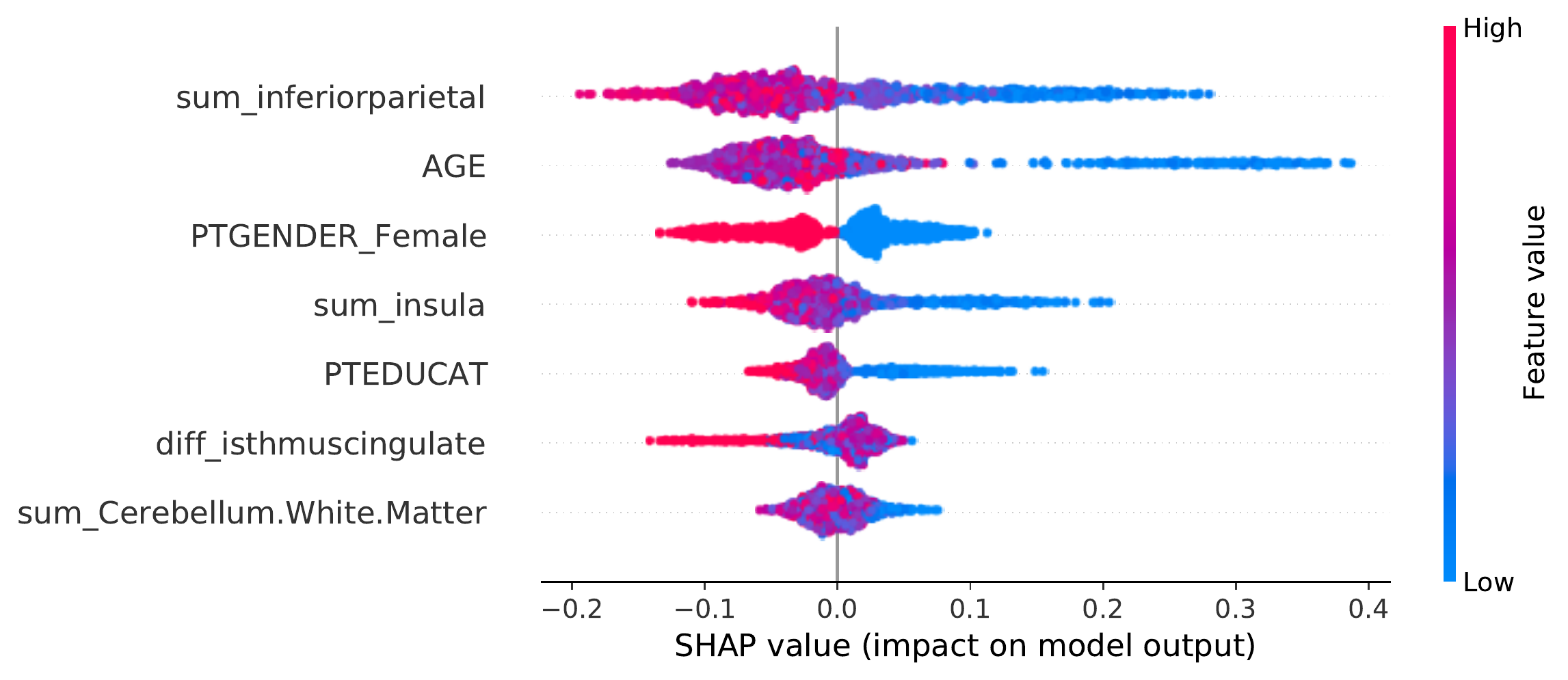}
		\caption{ADNI training and test set (\textit{n=1,365})}
		\label{fig:CNMCI_FS-1_RF_RFMean_SHAP}
	\end{subfigure}
	\hfill
	\begin{subfigure}[b]{0.49\textwidth}
		\centering
		\includegraphics[width=\textwidth]{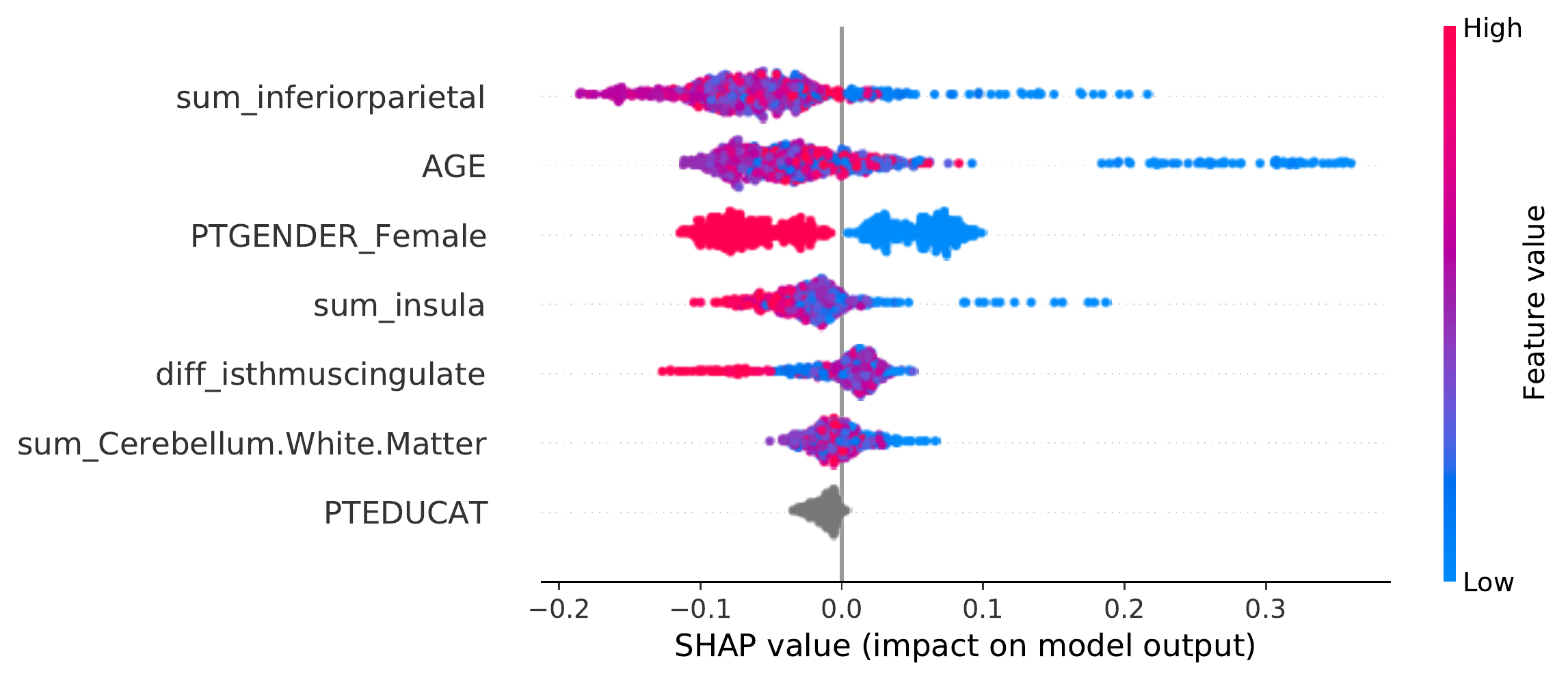}
		\caption{AIBL dataset (\textit{n=541})}
		\label{fig:CNMCI_FS-1_RF_RFMean_SHAP_AIBL}
	\end{subfigure}\\\vspace{4pt} 
	
	\begin{subfigure}[b]{0.49\textwidth}
		\centering
		\includegraphics[width=\textwidth]{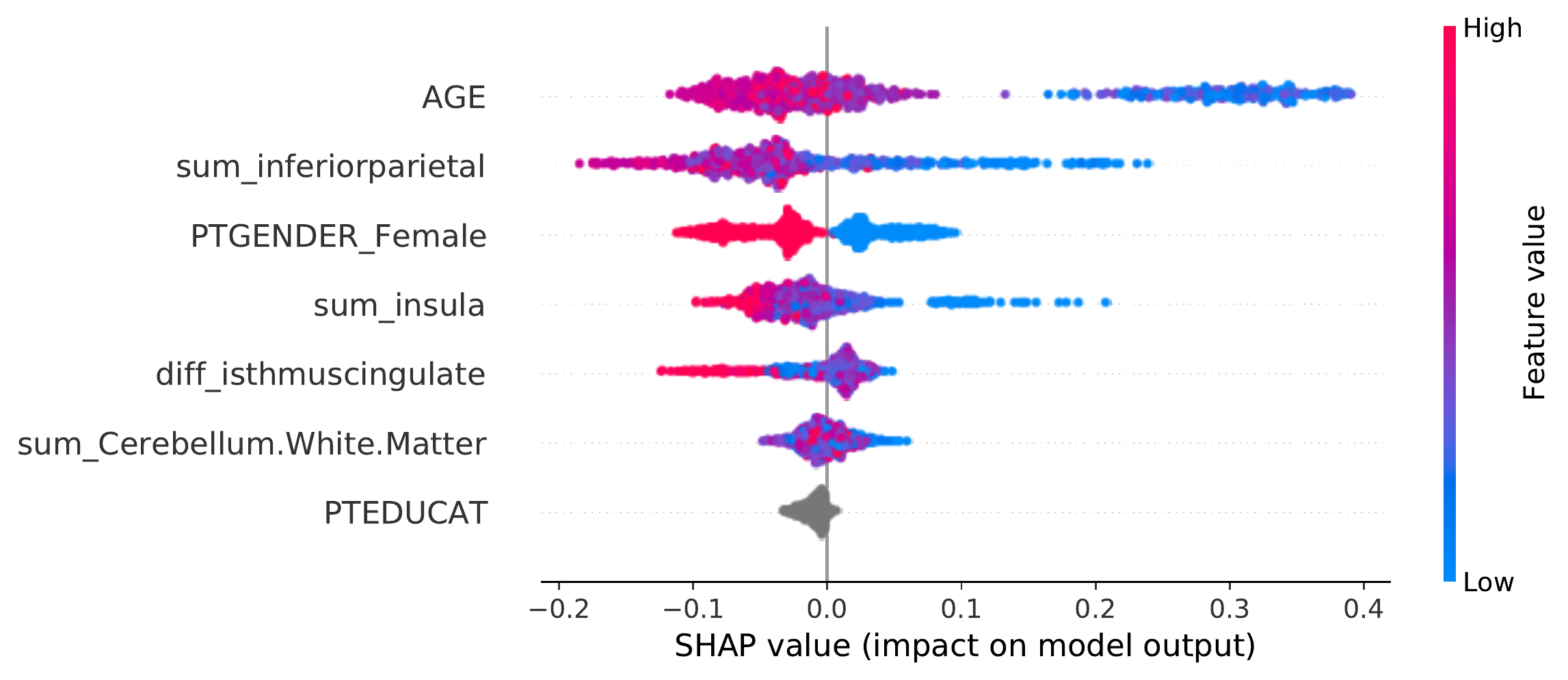}
		\caption{OASIS dataset (\textit{n=723})}
		\label{fig:CNMCI_FS-1_RF_RFMean_SHAP_OASIS}
	\end{subfigure}
	\caption{SHAP summary plots of the RF trained with FS-1 and feature selection for CN vs. MCI classification. The plots visualize the Shapley values of subjects from the ADNI, AIBL, and OASIS datasets and the ten most important model features. Each subject is represented by a dot. The colors decode the feature values of the subject. High feature values are colored in red whereas small feature values are colored in blue. The model learned that features with high Shapley values increased the patient's risk to develop MCI. Each plot shows the results on a different dataset}
	\label{fig:Reproducibility_Comparison_RF_CNMCI_SHAP}
\end{figure}

SHAP summary plots for the RF trained with feature selection for CN vs. MCI classification based on FS-1 are shown in Figure \ref{fig:Reproducibility_Comparison_RF_CNMCI_SHAP}. The figure contains subplots for all three datasets. Overall, the Shapley values for this model were asymmetric. The positive Shapley values show stronger amplitudes than the negative ones. One explanation for this behavior might be, that the MCI class was more frequent in the ADNI training dataset.
For the ADNI and AIBL dataset, the most important feature was the sum of the inferior parietal lobules followed by the age and gender. The model learned that small brain volumes, young age, and male gender increased the risk to develop MCI. The volumetric observations correspond to previous research \cite{10.1016/j.dadm.2015.11.002,10.1371/journal.pone.0066367,10.1038/s41598-018-29295-9,10.1038/nrneurol.2009.215}. The volume of the inferior parietal lobules was the second most important feature for the OASIS dataset. Age was the most important feature for OASIS and the second most important feature for ADNI and AIBL. The model learned young age was associated with disease progression. It can be noted in Table \ref{Table:ADNIBaseline} that the mean age of CN subjects is older than the mean age of MCI subjets in the ADNI dataset but not in the AIBL (Table \ref{Table:AIBLBaseline}) and OASIS (Table \ref{Table:OASISBaseline}) datasets. The differences observed in the datasets might cause problems in model reproducibility. The feature representing the years of education was in the fifth position for the ADNI dataset. That information was not available in OASIS and AIBL and was thus colored in grey. Consistently, this feature was the least important one for both datasets. Overall, the ranking of the feature importance differed for all models. 

\subsection{Classification Model}\label{SEC:ClassificationModel}

\begin{figure}
	\includegraphics[width=1\textwidth]{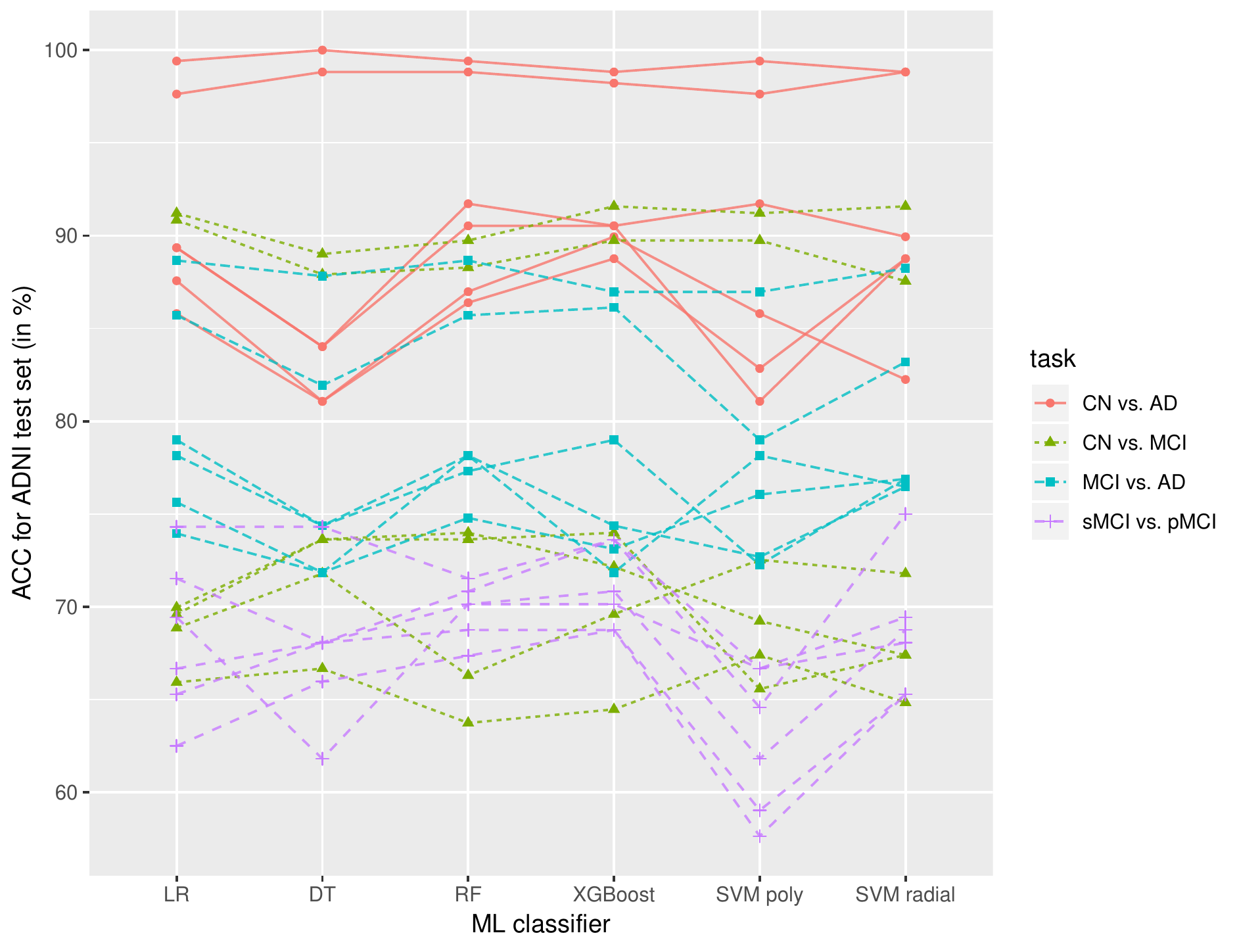}
	\caption{Line plot showing the accuracies achieved for the independent ADNI test set dependently on the classification tasks and the ML model. The plot includes all 216 models described in Table \ref{Table:ResultsCNAD_FS1}, Table \ref{Table:Results_CNMCI_FS-1}, Table \ref{Table:Results_MCIAD_FS1}, and Table \ref{Table:ResultssMCIpMCI_FS1}. For each ML classifier, 36 models were included}
	\label{fig:Boxplot_ADNI_Accuracy_By_MLModel} 
\end{figure}

In this research, six ML models were trained to compare their results to each other. A line plot of the accuracies achieved for the independent ADNI test set dependently on the classification task and the ML model is shown in Figure \ref{fig:Boxplot_ADNI_Accuracy_By_MLModel}. For the sMCI vs. pMCI classification, it can be seen, that the performance variance is smaller for RF and XGBoost models in comparison to the remaining ML models. In addition, the polynomial SVMs achieved worse results for this classification task. Overall, the DT models were often outperformed by RF and XGBoost classifiers. The LR models outperformed the DTs in many cases, except for the CN vs. MCI classification. Overall, no ML model outperformed the remaining models.

To indicate whether the differences in ADNI test accuracies between the ML methods are statistically significant, a Friedman test ($p-value < 0.05$) was executed. For this investigation, the results of Table \ref{Table:ForwardSelectionCNAD}, Table \ref{Table:ForwardSelectionCNMCI}, Table \ref{Table:ForwardSelectionMCIAD}, and Table \ref{Table:ForwardSelectionsMCIpMCI} are summarized, resulting in 24 observations per feature set (three feature sets, two feature selection methods, and four tasks). The p-value of $0.006$ indicated statistically significant differences between the ML models. A pairwise Wilcoxon signed rank test ($p-value < 0.05$) with Bonferroni adjustment was executed, to identify, which model performances differed from each other. However, the results, summarized in Table \ref{Table:Wilcox_model_Comp}, show that there are no statistically significant differences between ML models. 

\begin{table}[h]	\caption{P-values of the pairwise Wilcoxon signed rank test ($p-value < 0.05$) with Bonferroni adjustment to compare the differences in ADNI test accuracies between the six ML models. Statistically significant results are highlighted in bold.}
	\label{Table:Wilcox_model_Comp}
	\begin{center}
			\begin{tabular}{r|rrrrrr}
				\toprule
				&LR&DT&RF&XGBoost&SVM poly&SVM radial\\\midrule
				LR&-&-&-&-&-&-\\
				DT&0.622&-&-&-&-&-\\
				RF&1.000&0.087&-&-&-&-\\
				XGBoost&1.000&0.141&1.000&-&-&-\\
				SVM poly&0.081&1.000&0.081&0.111&-&-\\
				SVM radial&1.000&1.000&1.000&1.000&0.402&-\\
			
			\botrule
			\end{tabular}
		\end{center}
	\end{table}

\begin{figure}
	\centering
	\begin{subfigure}[b]{0.49\textwidth}
		\centering
		\includegraphics[width=\textwidth]{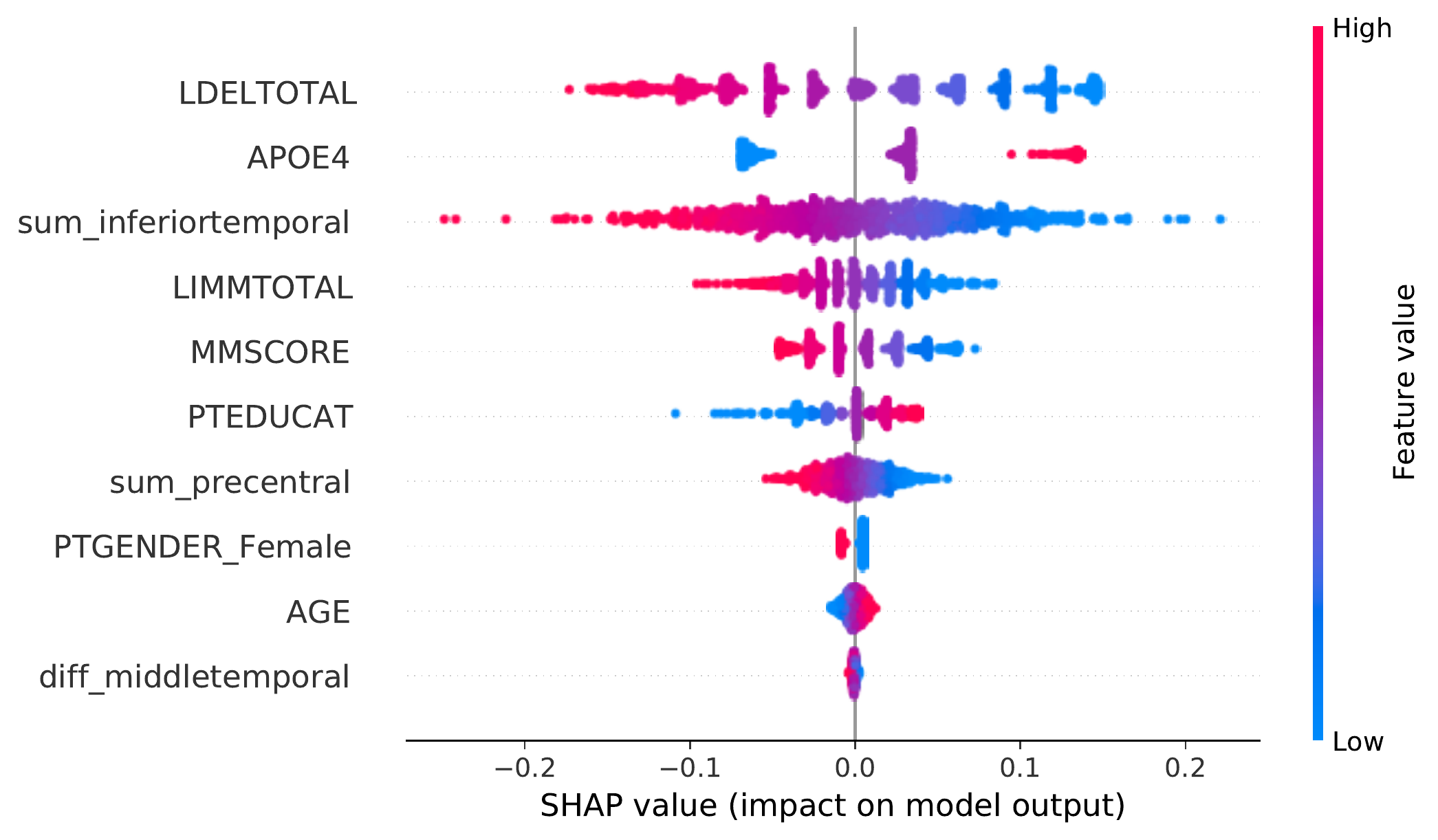}
		\caption{LR}
		\label{fig:sMCIpMCI_FS-3_LR_RFMean_SHAP}
	\end{subfigure}\hfill
\begin{subfigure}[b]{0.49\textwidth}
	\centering
	\includegraphics[width=\textwidth]{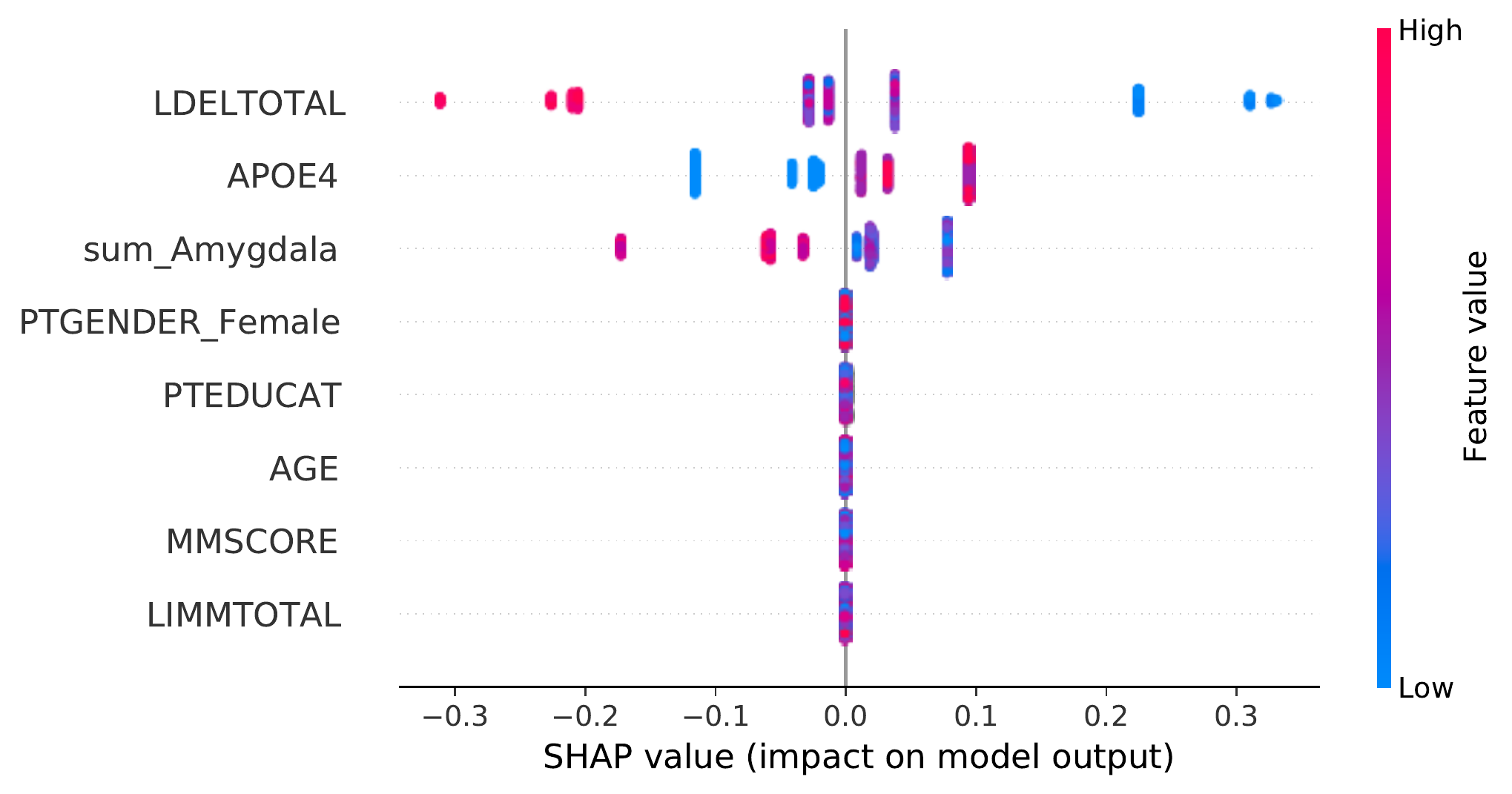}
	\caption{DT}
	\label{fig:sMCIpMCI_FS-3_DT_RFMean_SHAP}
\end{subfigure}\\\vspace{4pt}
	\begin{subfigure}[b]{0.49\textwidth}
		\centering
		\includegraphics[width=\textwidth]{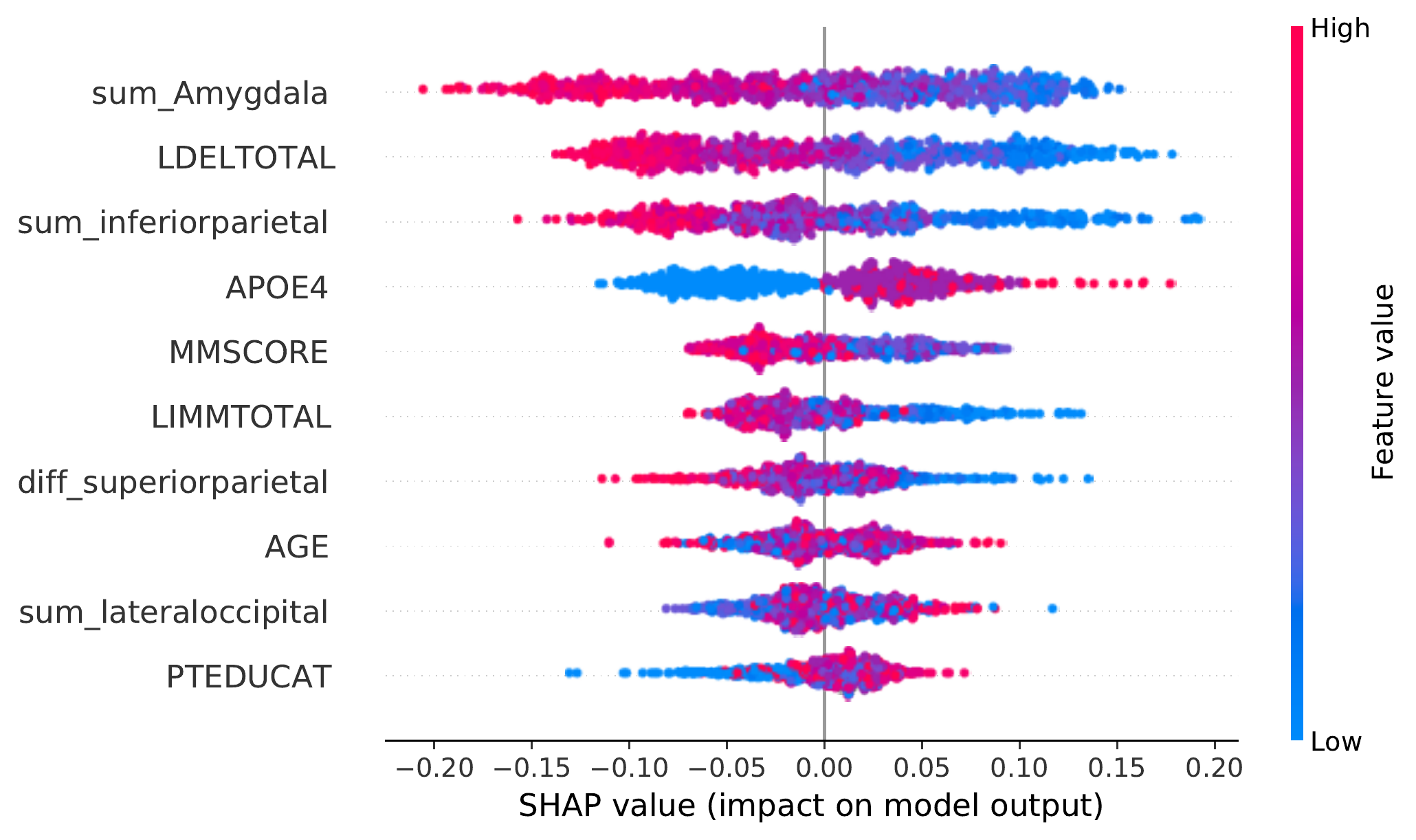}
		\caption{RF}
		\label{fig:sMCIpMCI_FS-3_RF_RFMean_SHAP}
	\end{subfigure}
	\hfill
	\begin{subfigure}[b]{0.49\textwidth}
		\centering
		\includegraphics[width=\textwidth]{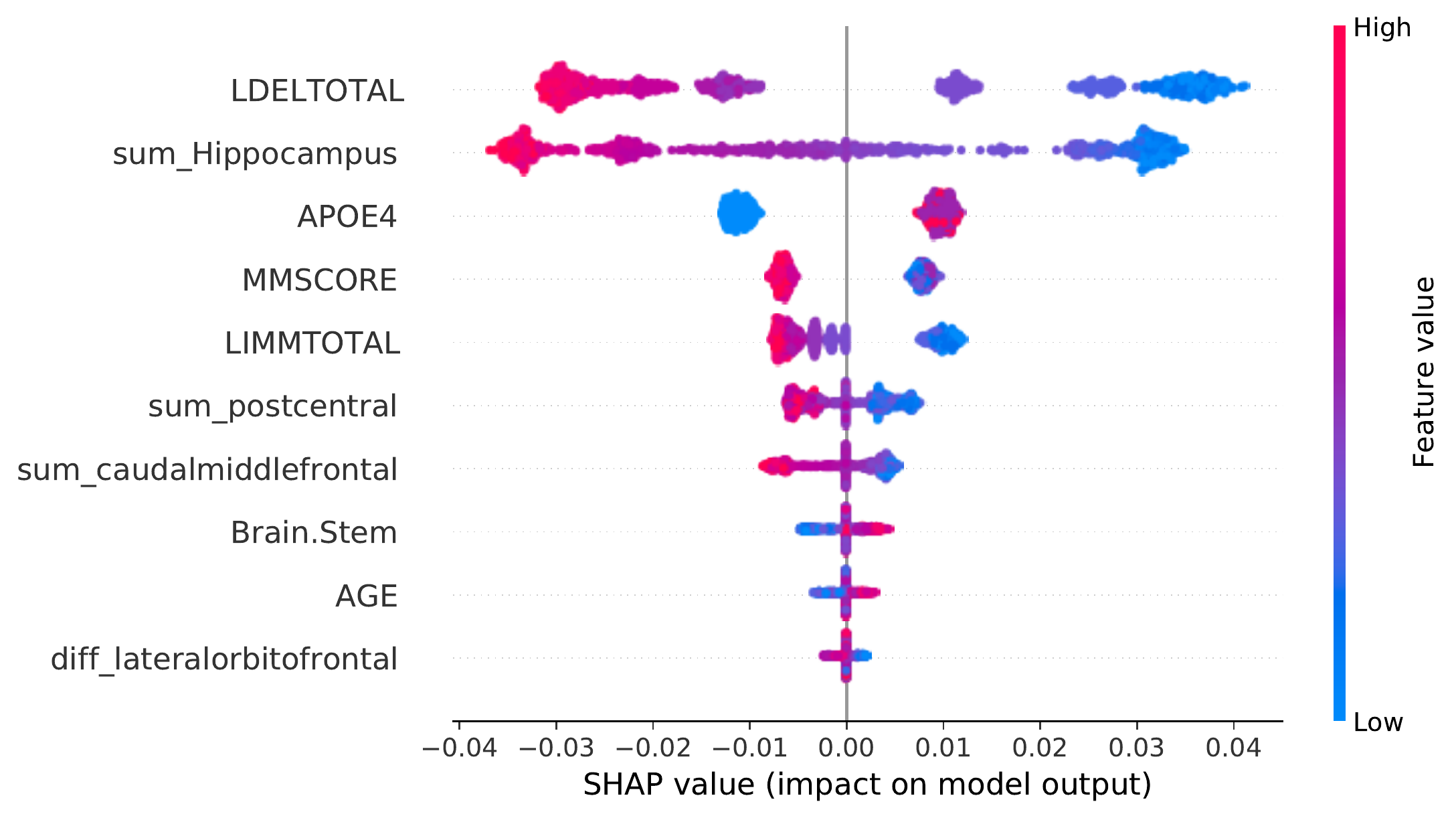}
		\caption{XGBoost}
		\label{fig:sMCIpMCI_FS-3_XGBoost_RFMean_SHAP}
	\end{subfigure}\\\vspace{4pt} 
	\begin{subfigure}[b]{0.49\textwidth}
		\centering
		\includegraphics[width=\textwidth]{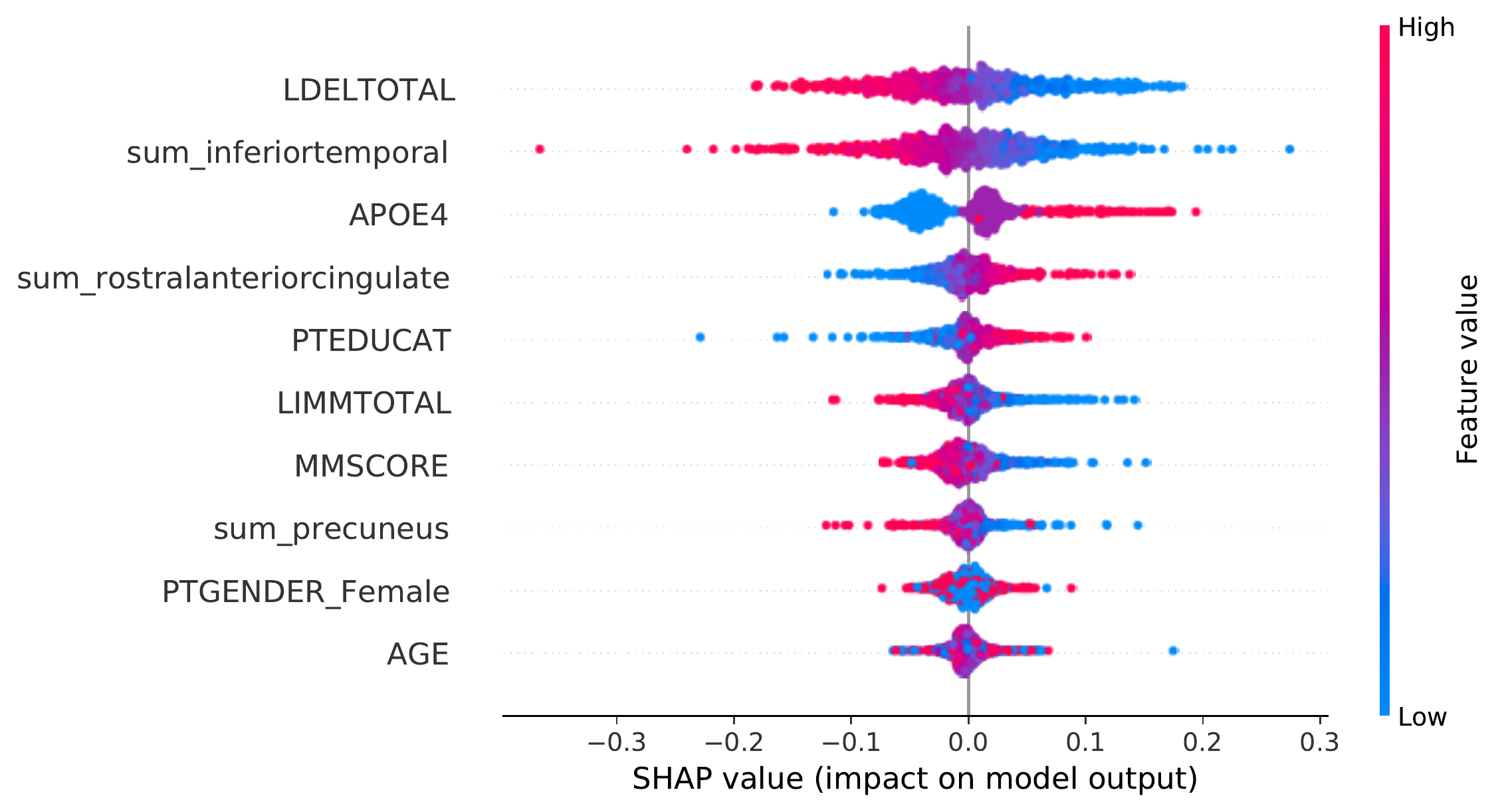}
		\caption{Polynomial SVM}
		\label{fig:sMCIpMCI_FS-3_SVMPoly_RFMean_SHAP}
	\end{subfigure}\hfill
	\begin{subfigure}[b]{0.49\textwidth}
		\centering
		\includegraphics[width=\textwidth]{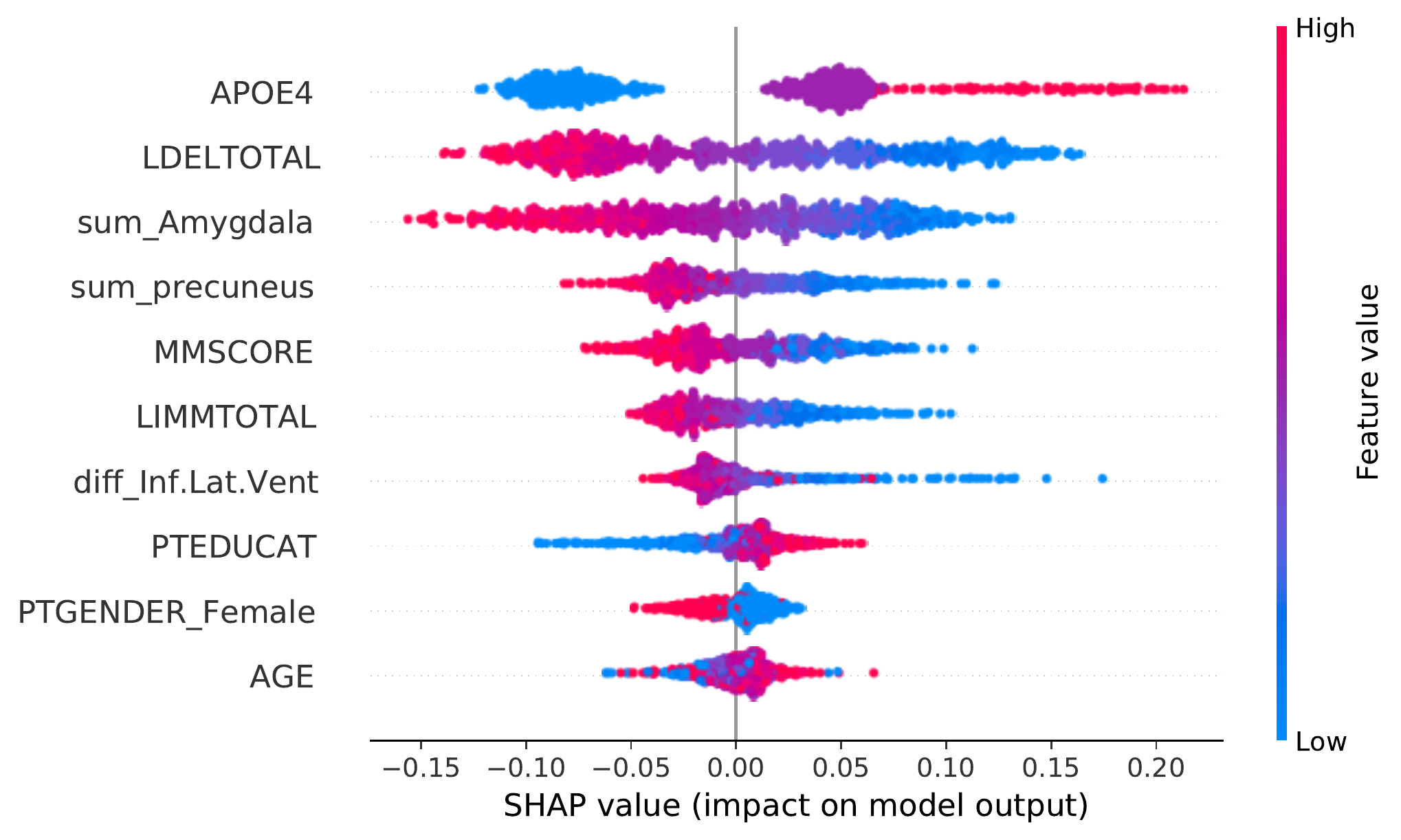}
		\caption{Radial SVM}
		\label{fig:sMCIpMCI_FS-3_SVMRadial_RFMean_SHAP}
	\end{subfigure}\\\vspace{4pt}
	\caption{SHAP summary plots for sMCI vs. pMCI classification trained using FS-3, feature selection, and multiple ML models. The Shapley values of \textit{n=747} ADNI and AIBL subjects and the ten most important model features are shown. Each dot represents a subject. The colors decode the feature expressions. High expressions are colored in red and small ones in blue. The model learned that features with high Shapley values increased the patient's AD risk}
	\label{fig:Comparison_ClassificationModels_sMCI_FS-3_RFMean_SHAP}
\end{figure}

To visualize ML model differences, Figure \ref{fig:Comparison_ClassificationModels_sMCI_FS-3_RFMean_SHAP} shows SHAP summary plots for all six models. All models were trained using FS-3 with feature selection to distinguish between sMCI and pMCI subjects. The feature selection results in slightly different features within all models. Overall, the Shapley values had the largest deviance for the DT and the SVM with a polynomial kernel, followed by the LR model and the RF. The most important feature for all models except for the RF and the radial SVM was the LDELTOTAL cognitive test score. For this test score, all models associated small feature values (colored in blue) with disease progression. For the radial SVM and the RF, LDELTOTAL was the second most important feature. The most important feature in the RF model was the sum of the left and right amygdalae. The model learned, that large brain volumes decreased the patient's risk to develop AD. This observation is biologically plausible \cite{10.1016/j.dadm.2015.11.002,10.1371/journal.pone.0066367,10.1038/s41598-018-29295-9,10.1038/nrneurol.2009.215}. The sum of the amygdala volumes was the third most important feature in the DT and the radial SVM. The number of ApoE$\epsilon$4 alleles was the most important feature for the radial SVM. The model learned that ApoE$\epsilon$4 is an AD risk factor, and the presence of ApoE$\epsilon$4 alleles is associated with AD progression. The number of ApoE$\epsilon$4 alleles is the second most important feature for the DT, and the LR, the third most important feature for the XGBoost model and the polynomial SVM, and the fourth most important feature for the RF. In this comparison, all models except for the DT and the polynomial SVM had at least one asymmetry feature within its top ten features. The decision tree only depended on three features, namely the LDELTOTAL cognitive test score, the number of ApoE$\epsilon$4 alleles, and the hippocampus volume. Most associations, the models learned, were biologically plausible. The radial RF showed two features with a biologically implausible association \cite{10.1016/j.dadm.2015.11.002,10.1371/journal.pone.0066367,10.1038/s41598-018-29295-9,10.1038/nrneurol.2009.215}. The model learned, that high volumes of the lateral occipital sulci, as well as a high number of education years, are associated with disease progression. Those features are ranked as the ninth and tenth important features in this model. Surprisingly the association of the education feature was also learned for the SVMs and the LR. For the polynomial SVM, the summed volumes of the rostral anterior cingulate cortices show a biologically implausible \cite{10.1016/j.dadm.2015.11.002,10.1371/journal.pone.0066367,10.1038/s41598-018-29295-9,10.1038/nrneurol.2009.215} association. Overall, biological plausibility should only be expected for high-performing models.

\begin{figure}
	\centering
		\begin{subfigure}[b]{0.49\textwidth}
		\centering
		\includegraphics[width=\textwidth]{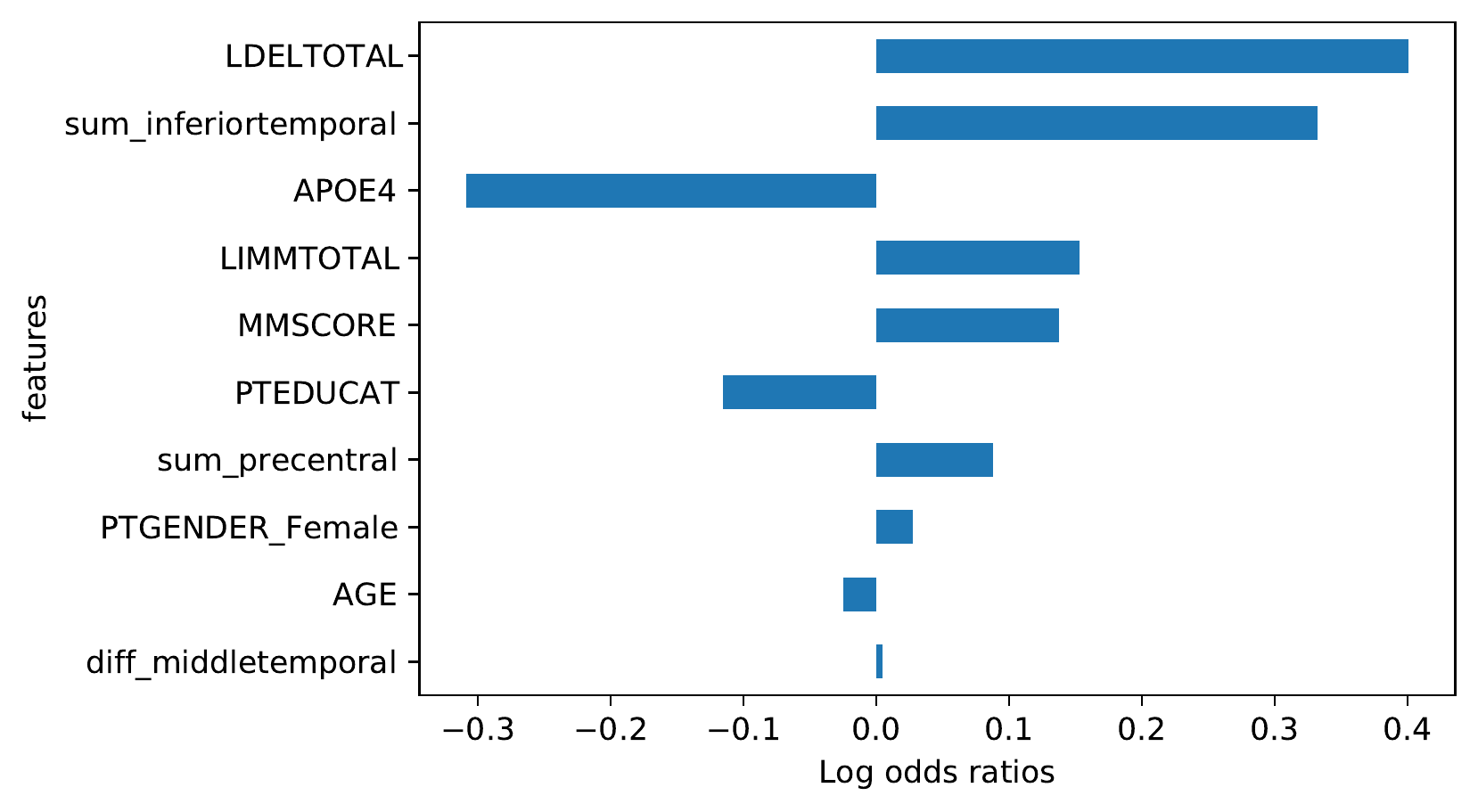}
		\caption{LR}
		\label{fig:FIP_LR_Model}
	\end{subfigure}\hfill
		\begin{subfigure}[b]{0.49\textwidth}
		\centering
		\includegraphics[width=\textwidth]{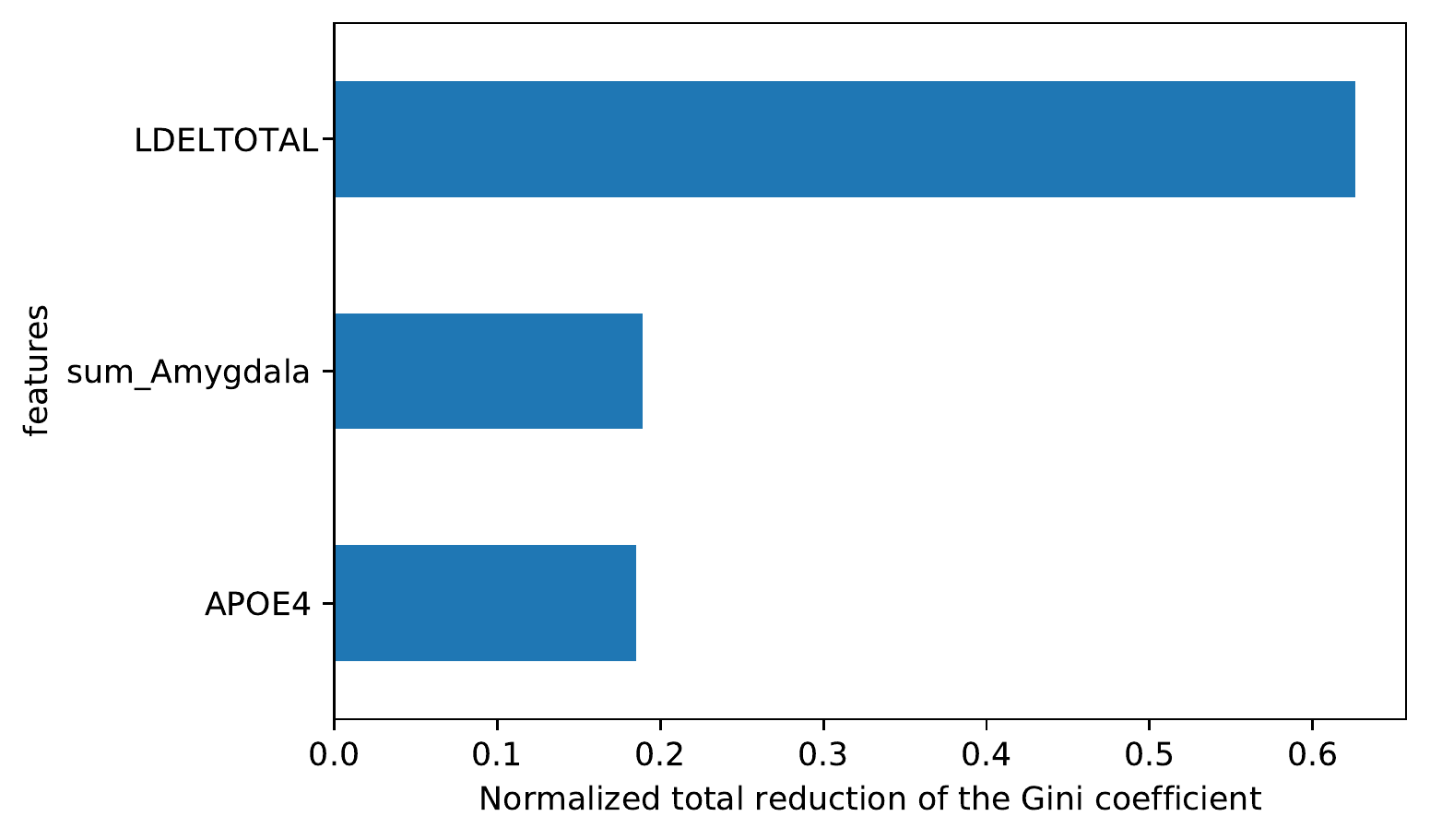}
		\caption{DT}
		\label{fig:FIP_DT_Model}
	\end{subfigure}\\\vspace{4pt}
	\begin{subfigure}[b]{0.49\textwidth}
		\centering
		\includegraphics[width=\textwidth]{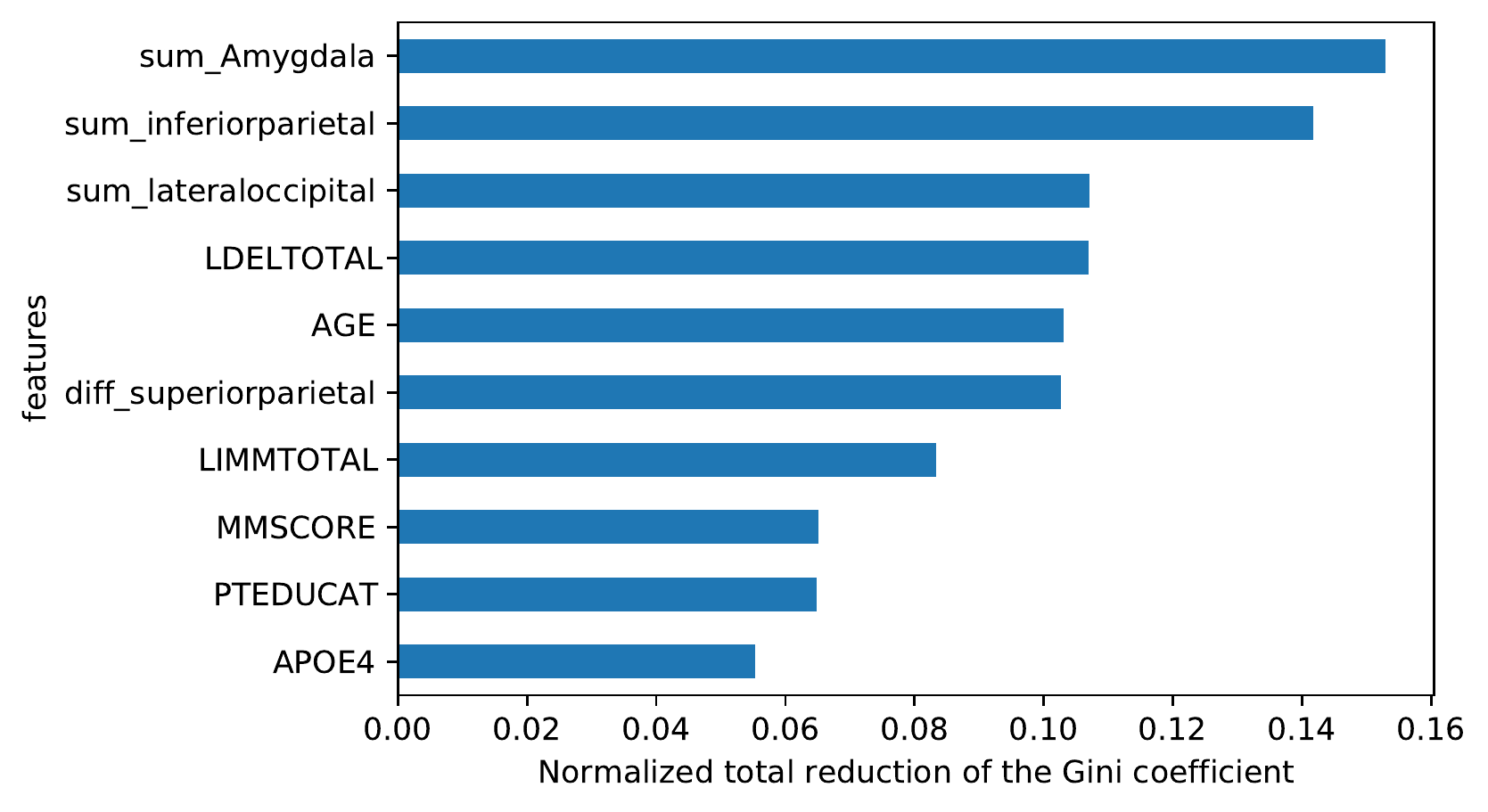}
		\caption{RF}
		\label{fig:FIP_RF_Model}
	\end{subfigure}
	\hfill
	\begin{subfigure}[b]{0.49\textwidth}
		\centering
		\includegraphics[width=\textwidth]{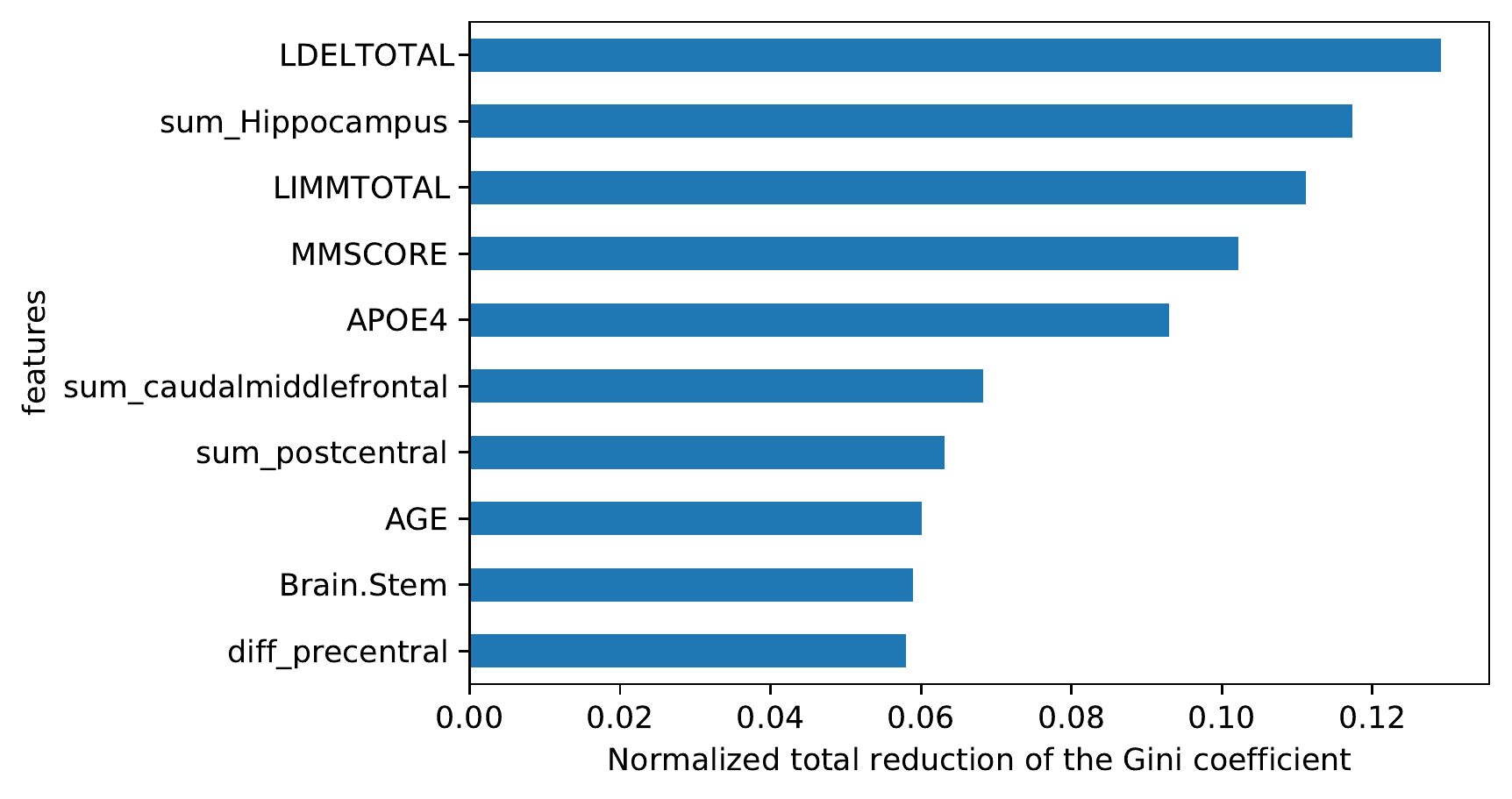}
		\caption{XGBoost}
		\label{fig:FIP_XGBoost_Model}
	\end{subfigure}
	\\\vspace{4pt} 
	\caption{Natural feature importance plots of the RF, XGBoost and DT models and log odd's ratios for LR models trained to distinguish between sMCI and pMCI subjects using FS-3 and feature selection. Each plot shows a different classification model}
	\label{fig:Comparison_ClassificationModels_sMCI_FS-3_RFMean_FIS}
\end{figure}

\begin{figure}
	\centering
	\begin{subfigure}[b]{0.49\textwidth}
		\centering
		\includegraphics[width=\textwidth]{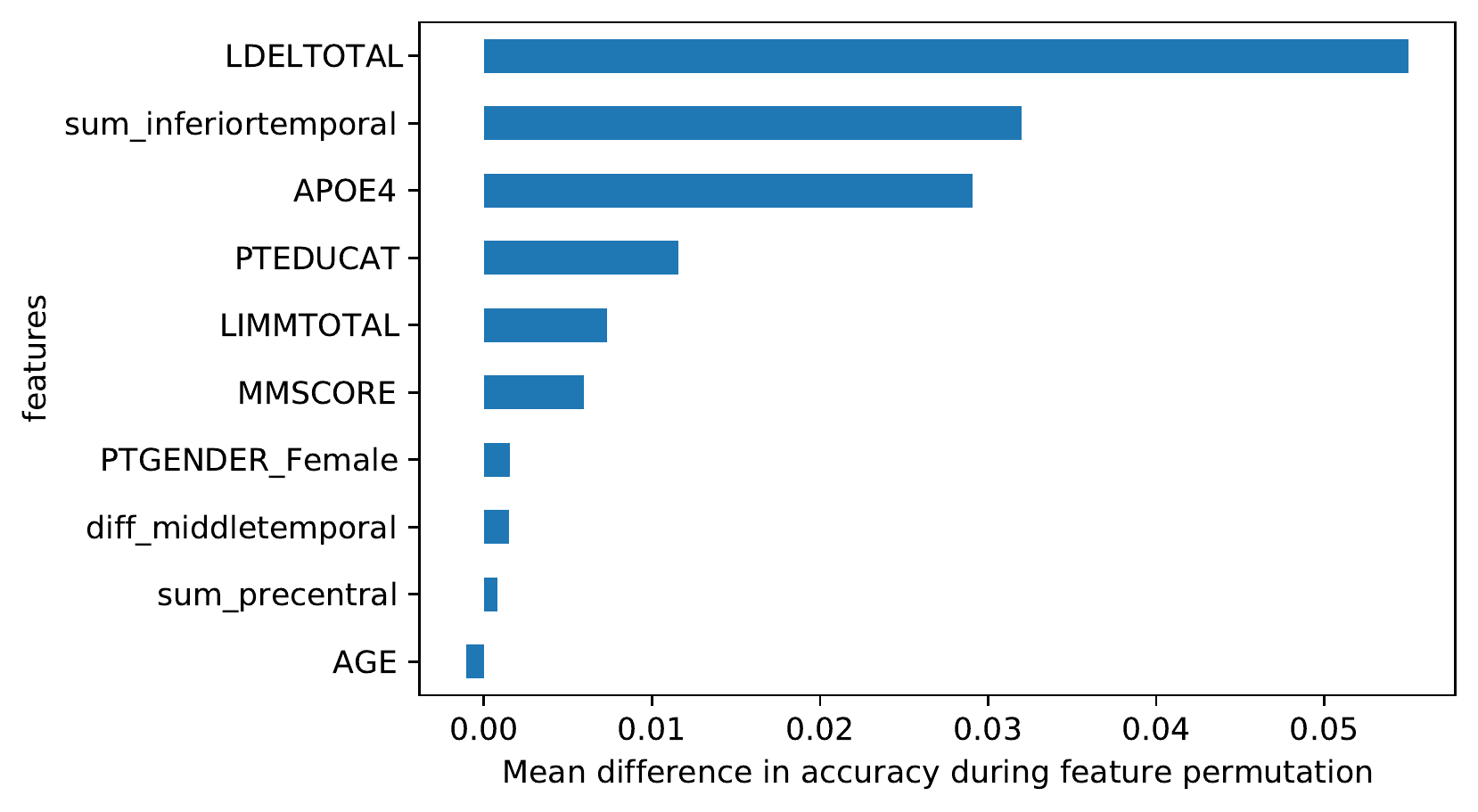}
		\caption{LR}
		\label{fig:PIP_LR_Model}
	\end{subfigure}\hfill
\begin{subfigure}[b]{0.49\textwidth}
	\centering
	\includegraphics[width=\textwidth]{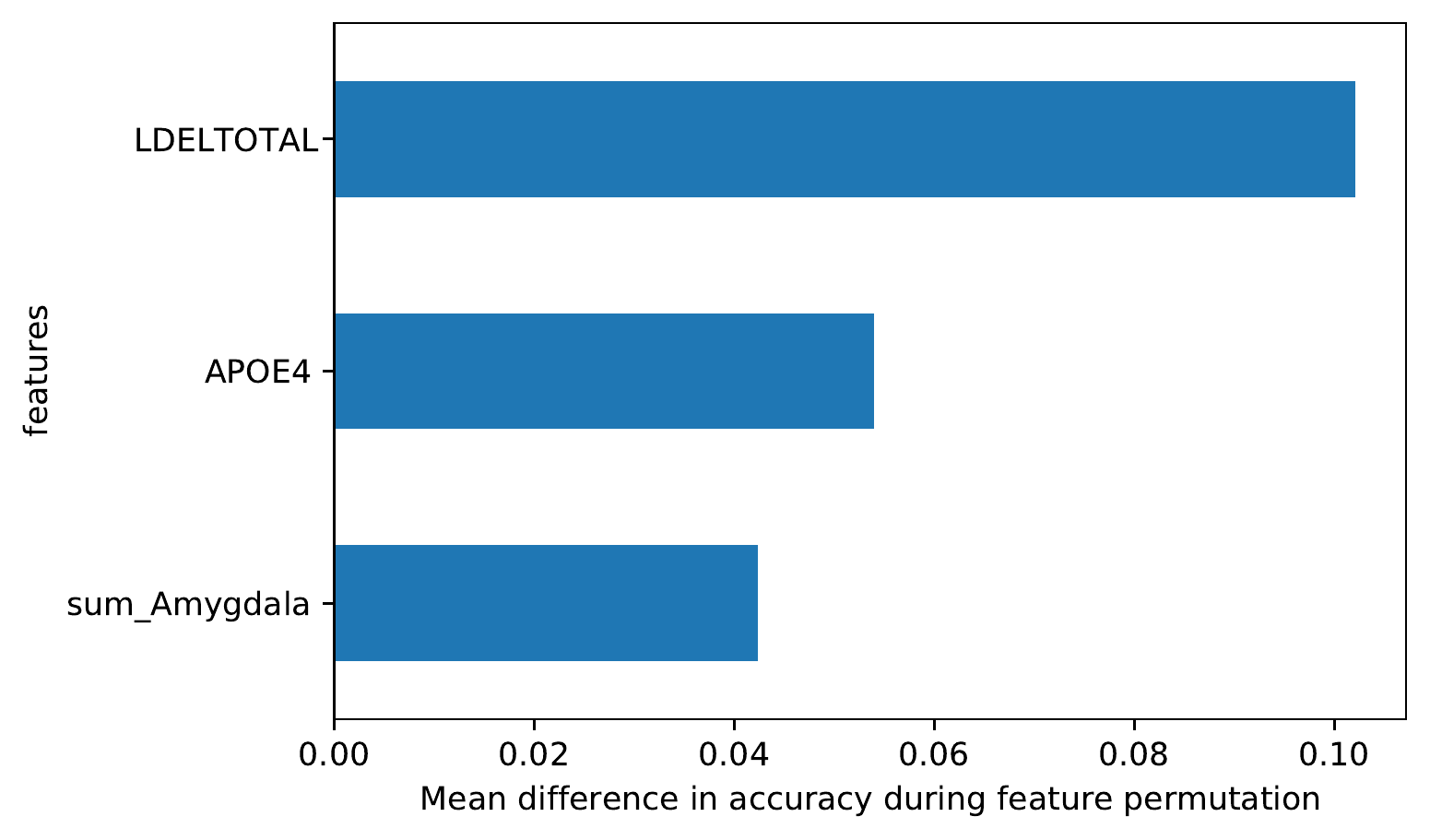}
	\caption{DT}
	\label{fig:PIP_DT_Model}
\end{subfigure}\\\vspace{4pt}
	\begin{subfigure}[b]{0.49\textwidth}
		\centering
		\includegraphics[width=\textwidth]{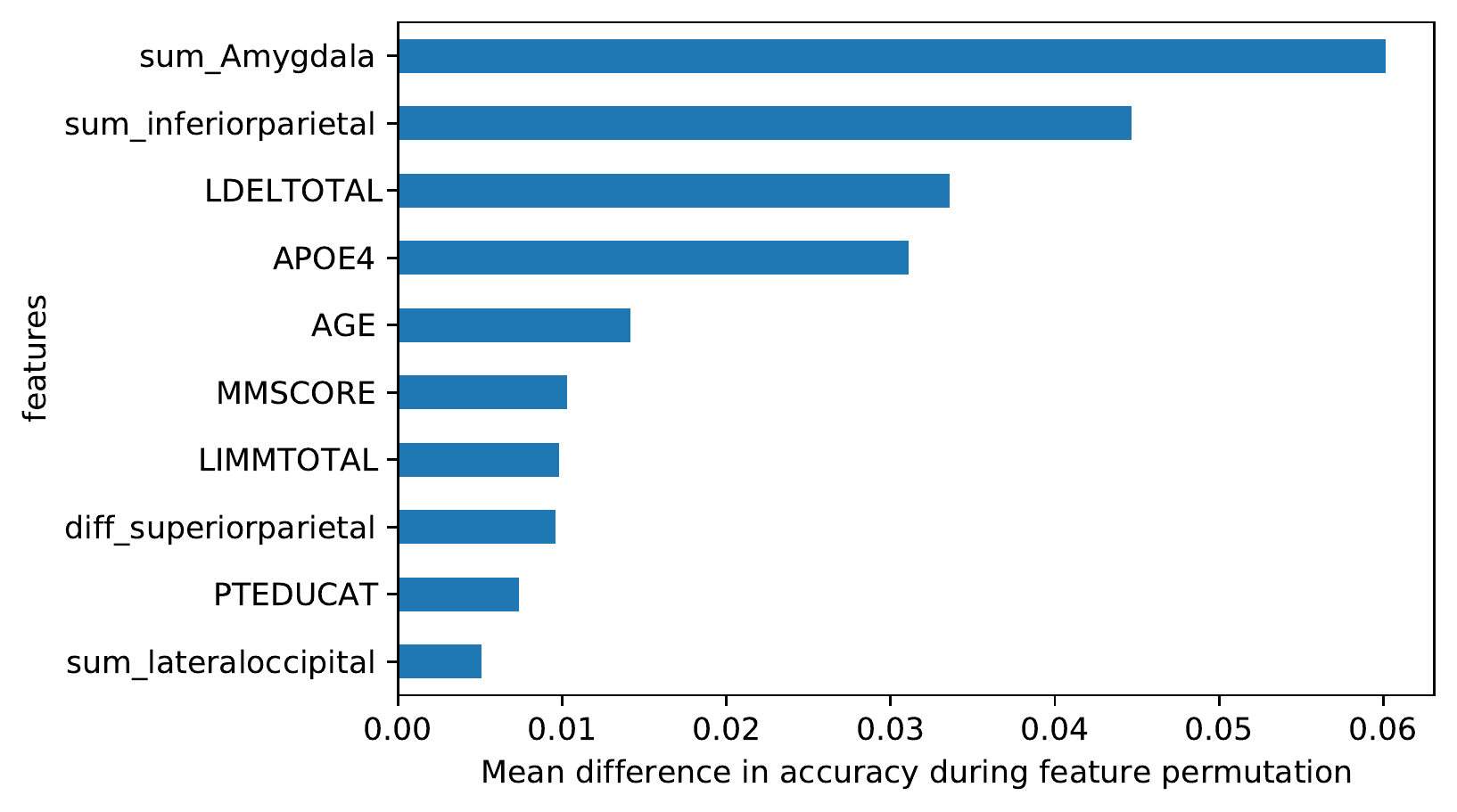}
		\caption{RF}
		\label{fig:PIP_RF_Model}
	\end{subfigure}
	\hfill
	\begin{subfigure}[b]{0.49\textwidth}
		\centering
		\includegraphics[width=\textwidth]{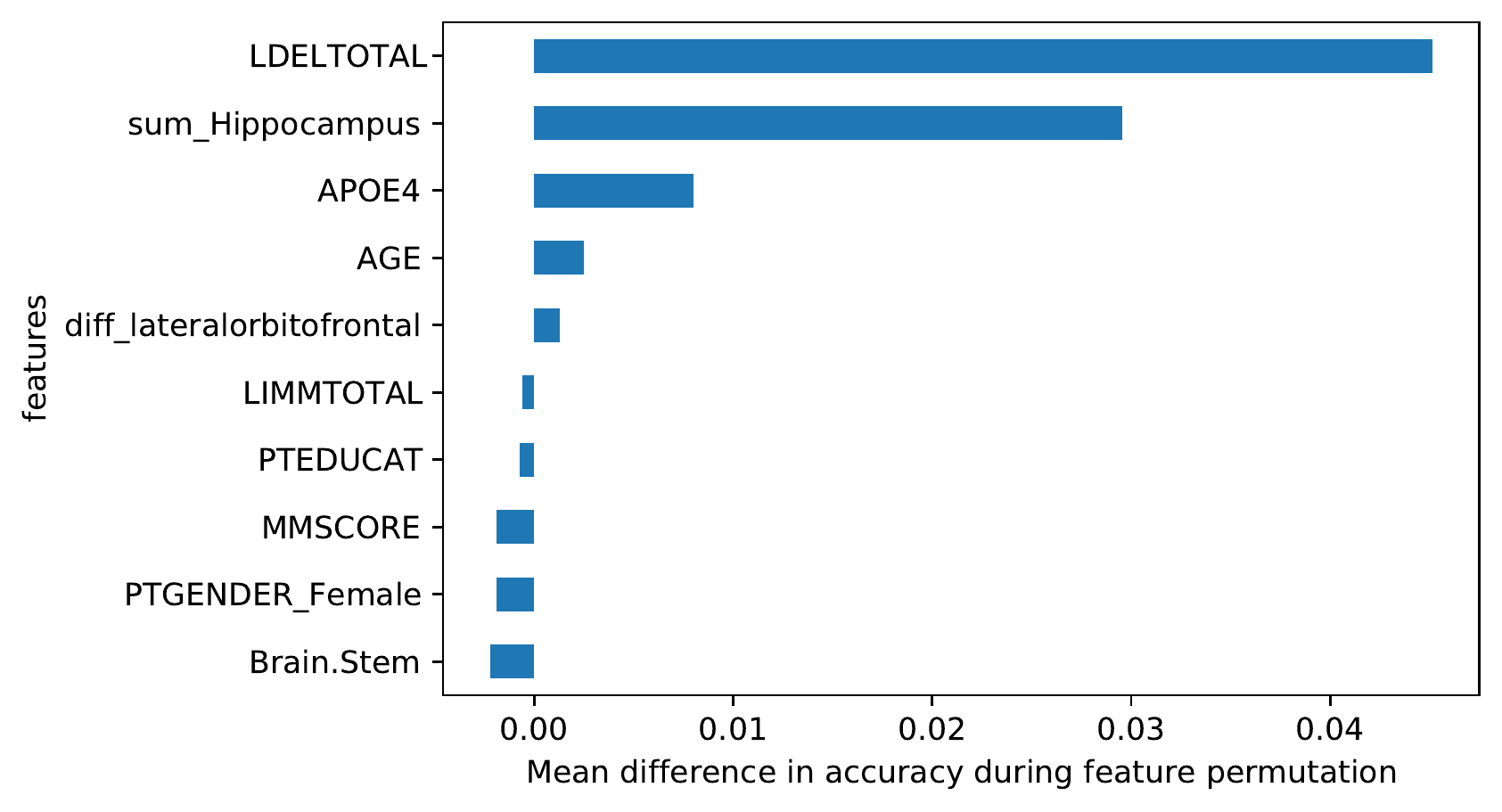}
		\caption{XGBoost}
		\label{fig:PIP_XGBoost_Model}
	\end{subfigure}
	\\\vspace{4pt} 
\begin{subfigure}[b]{0.49\textwidth}
	\centering
	\includegraphics[width=\textwidth]{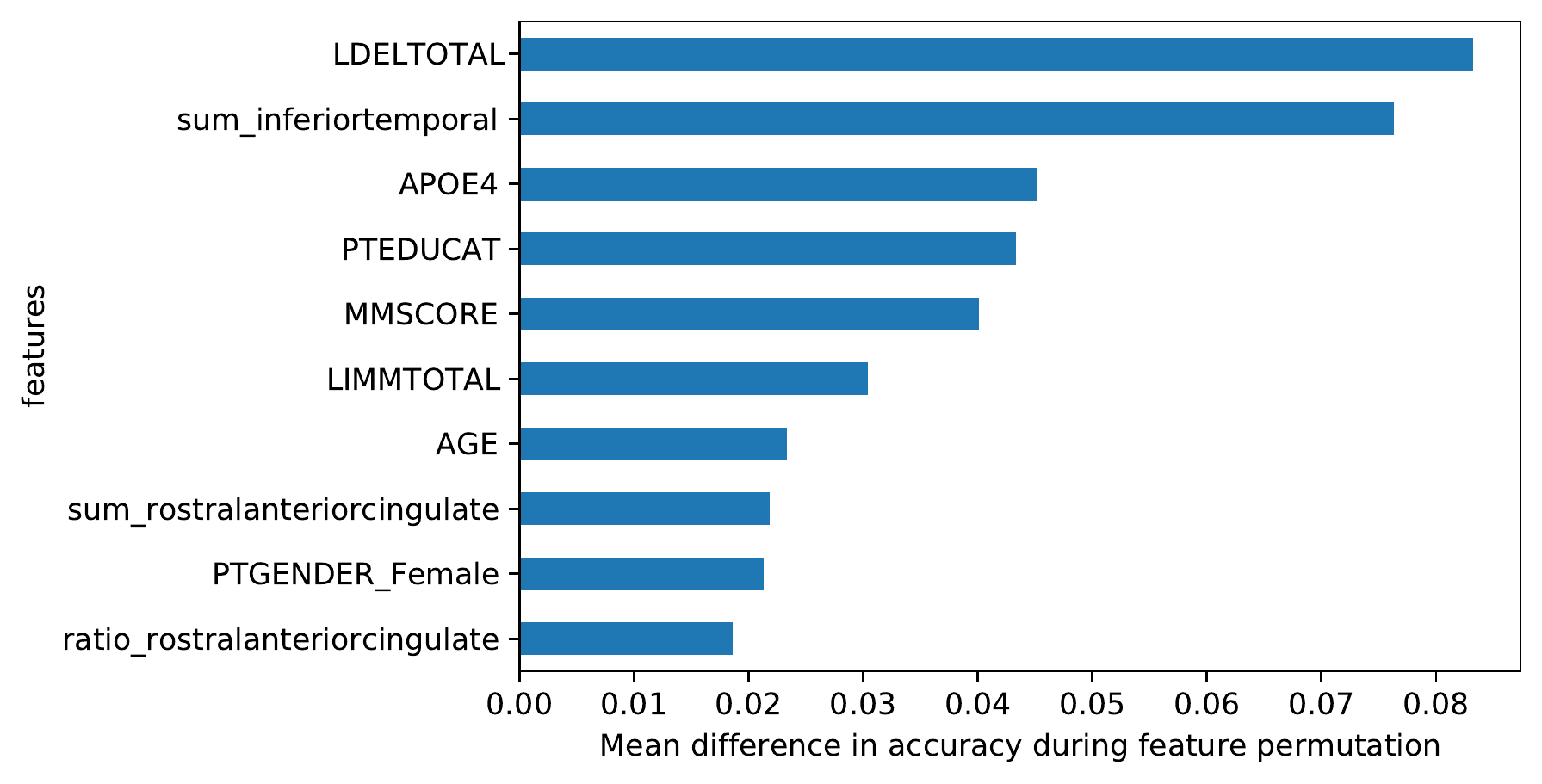}
	\caption{SVM poly}
	\label{fig:PIP_SVMPoly_Model}
\end{subfigure}
\hfill
\begin{subfigure}[b]{0.49\textwidth}
	\centering
	\includegraphics[width=\textwidth]{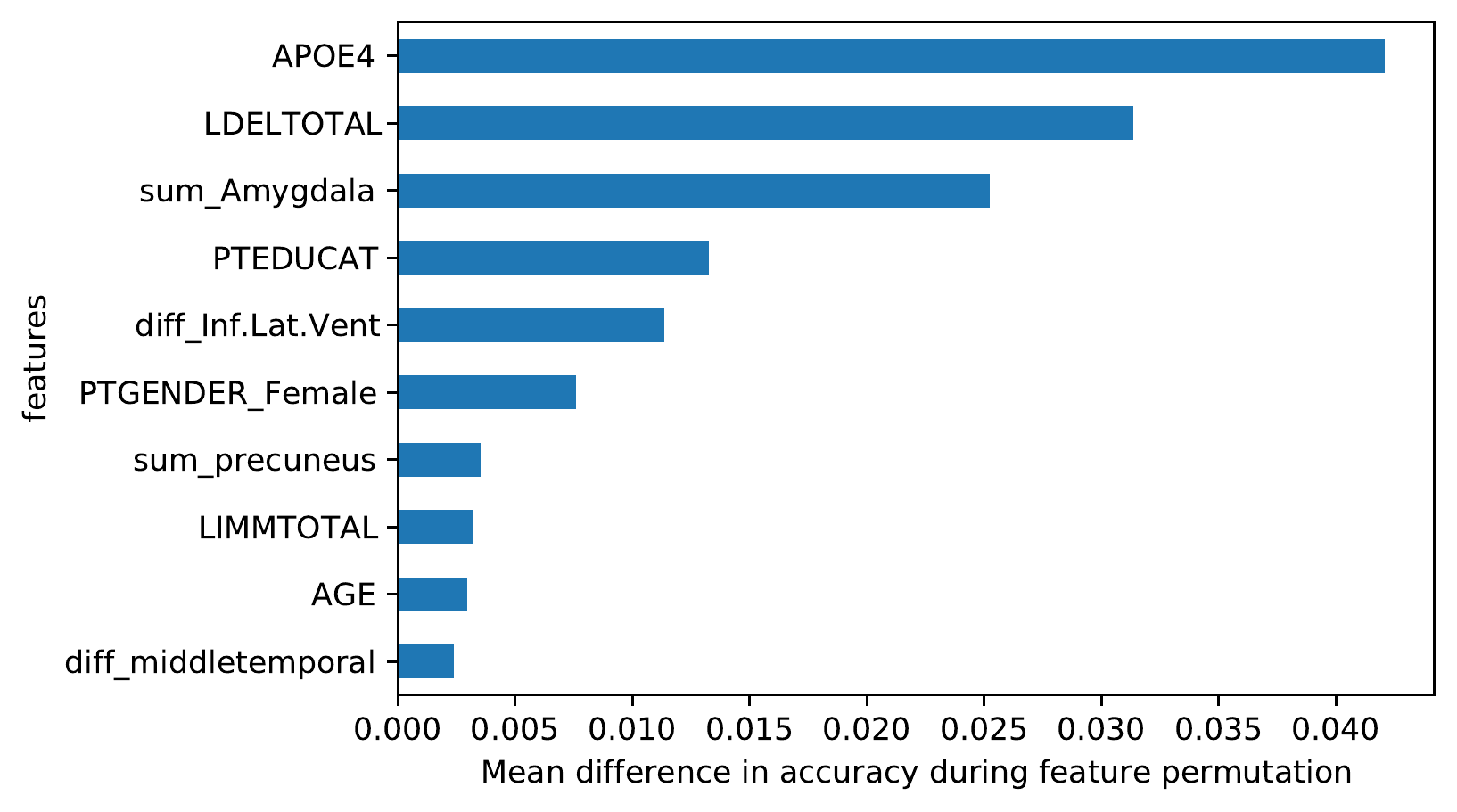}
	\caption{SVM radial}
	\label{fig:PIP_SVMRBF_Model}
\end{subfigure}\hfill

	\caption{Permutation importance plots of all six ML models trained to distinguish between sMCI and pMCI subjects using FS-3 and feature selection. Each plot shows a different classification model}
	\label{fig:Comparison_ClassificationModels_sMCI_FS-3_RFMean_PIP}
\end{figure}

As a comparison, Figure \ref{fig:Comparison_ClassificationModels_sMCI_FS-3_RFMean_FIS} visualizes the natural feature importance for the RF and XGBoost models, and the log odd's ratios for the LR model (ordered by the absolute log odd's ratio), and Figure \ref{fig:Comparison_ClassificationModels_sMCI_FS-3_RFMean_PIP} shows the permutation importance of all models. The most important features of all natural feature importance plots and all permutation importance plots correspond to the SHAP summary plots. 

\begin{figure}
	\begin{center}
	\includegraphics[width=0.7\textwidth]{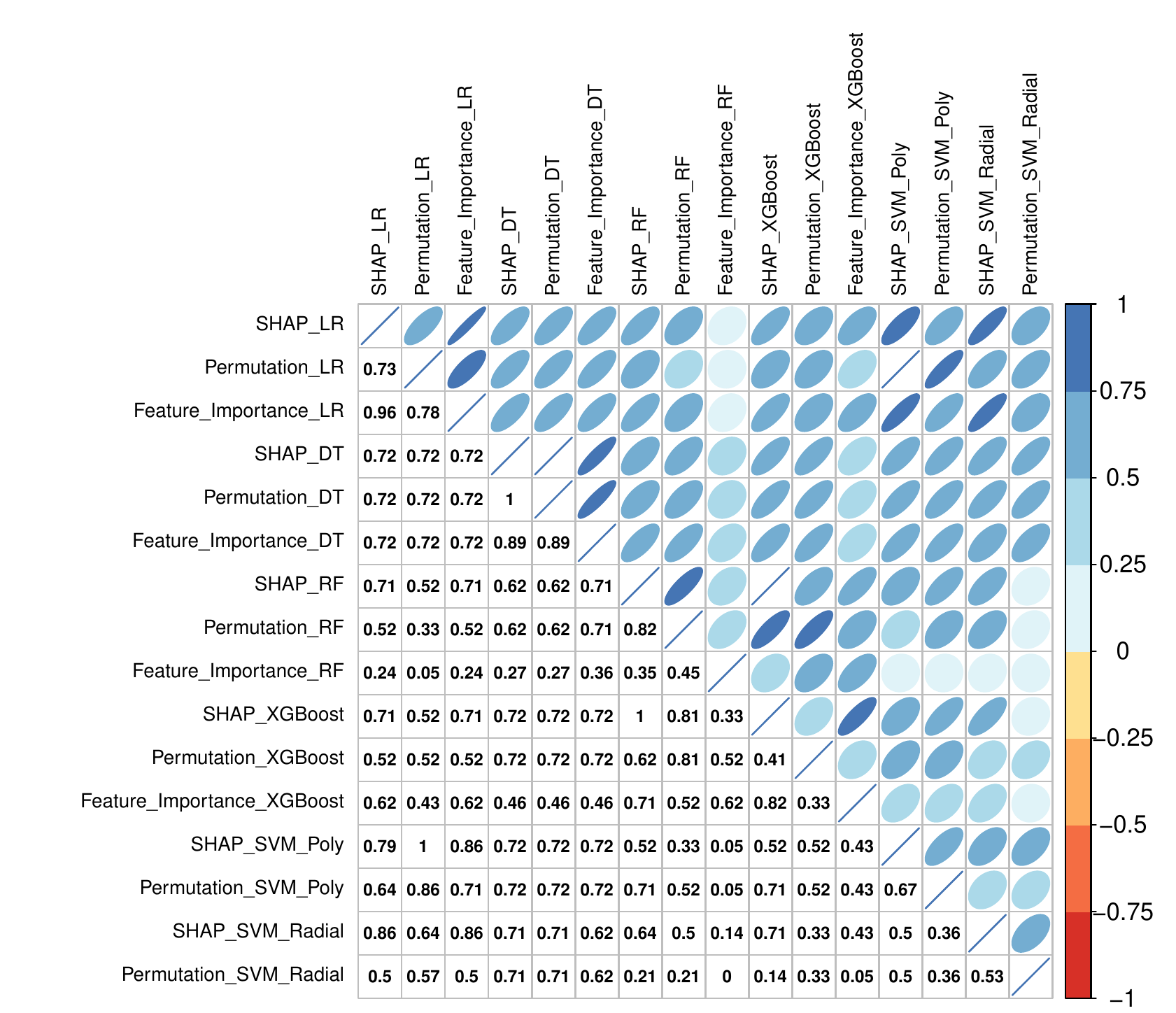}
	\caption{Plot showing Kendall's tau correlation between feature importances of all SHAP models, permutation importance of all models, and natural XGBoost and RF feature importances, as well as log odd's ratios of the LR models for FS-3, feature selection, sMCI vs. pMCI classification and the ADNI and AIBL datasets (\textit{n=747})}
	\label{fig:Correlation_plot_natural_feature_importances} 
	\end{center}
\end{figure}

The Kendall's tau rank correlation \cite{10.1093/biomet/30.1-2.81} between feature rankings for all SHAP models, natural XGBoost and RF feature importances, absolute log odd's ratios of the LR model, and permutation importance of all models is shown in Figure \ref{fig:Correlation_plot_natural_feature_importances}. Due to the forward feature selection, the different models are trained on slightly different MRI features and the correlation was calculated for pairwise complete observations. However, the sociodemographic data, the number of ApoE$\epsilon$4 alleles, and the cognitive test scores were used to train all models. As the features within a specific model are identical, firstly the SHAP values, the permutation importance, and the feature importances are compared for each individual model. The SHAP values of the RF model and the permutation importance of the RF have a correlation coefficient of 0.82. The natural feature importance of the RF is only moderately correlated to the permutation importance of the RF (0.45) and weakly correlated to the RF SHAP values (0.35). The XGBoost SHAP values showed a very strong correlation of 0.82 to the natural XGBoost feature importance and a moderate correlation of 0.41 to the XGBoost permutation importance. The DT selected three features in all methods leading to a perfect correlation between the DT SHAP values, and the permutation importance as well as a very strong correlation of 0.89 for the DT SHAP values and the natural feature importances. The SHAP values of the polynomial SVM showed a strong correlation to the permutation importance (0.67) of the same model. A moderate correlation of 0.53 was reached for the SHAP values of the radial SVM and the the permutation importance of the same model. The SHAP LR values are strongly correlated (0.73) to the permutation importances and very strong correlated to the log odds (0.96).

As previously mentioned, the features within the different ML models differed which makes the comparison of inter- and intra- model correlations difficult. Considering the inter-model correlations, a perfect correlation of 1 was reached between the SHAP values of the RF and the SHAP values of the XGBoost model, as well as the permutation importance of the LR and the SHAP values of the polynomial SVM.

The execution times of the different ML models, the SHAP algorithm, and the permutation importance calculation are summarized in Table \ref{Table:ExecutionTime}. All experiments were executed on an NVIDIA\textsuperscript{\textregistered} DGX-1\footnote{DGX-1: \url{https://www.nvidia.com/en-us/data-center/dgx-1/}, Accessed 2022-05-01} supercomputer. The execution environment was an NVIDIA\textsuperscript{\textregistered}-optimized\footnote{NVIDIA\textsuperscript{\textregistered}-Docker: \url{https://github.com/NVIDIA/nvidia-docker}, Accessed 2022-05-01} Docker\footnote{Docker: \url{https://www.docker.com/}, Accessed 2022-05-01} \cite{merkel2014docker} container, running a Deepo\footnote{Deepo: \url{https://github.com/ufoym/deepo}, Accessed 2022-05-01} image. The results showed, that except for the RF model all models were trained in less than 1 s. The mean training time during CV was 5.66 s for the RF. The RF model was trained using twelve features. The long training time of the RF model was also reflected in the SHAP algorithm and the permutation importance. To compute the SHAP values of one subject, the RF model consumes 47.79 s, whereas the radial SVM which achieved the second slowest time requires only 13.51 s per subject. This results in an execution time of approximately 10 h to calculate the RF SHAP values of the entire dataset. The execution time for permutation importance was approximately 1 h for the RF model and 8 min for the XGBoost model reaching the second-longest execution time. Overall, it has to be mentioned, that SHAP value calculation is a time-intensive process. However, the times presented can only be used as an orientation, and optimation is possible by for example clustering the background subjects of the SHAP algorithm. In this work, the samples of the entire training dataset were used as background subjects. The SHAP execution time depended on the number of features in a dataset, the time needed for model inference, the number of background subjects and the number of subjects that should be explained.  
\begin{table}[h]	\caption{Execution times of the different ML models and explainability methods for FS-3, feature selection, sMCI vs. pMCI classification, and the ADNI and AIBL datasets (\textit{n=747})}
	\label{Table:ExecutionTime}
	\begin{center}
			\begin{tabular}{r|r|r|r|r|r}
				\toprule
			ML model&\# model&Mean CV&Mean SHAP&Global&Global time\\
			&features&training&time / subject&SHAP time&permutation\\
			&&time &(in h)&($n=747$)&importance\\
			&&(in h)&&(in h)&(in h)\\\midrule
			
			RF&12&00:00:05.66& 00:00:47.79&09:54:59.00&00:54:19.27\\
			XGBoost&13&00:00:00.18& 00:00:04.37&00:54:22.00& 00:07:36.85\\
			DT&8&00:00:00.01&00:00:01.18&00:14:41.00&00:01:39.61\\
			SVM Poly&11&00:00:00.15& 00:00:08.09&01:40:44.00&00:05:05.77\\
			SVM Radial&11&00:00:00.19& 00:00:13.51&02:48:15.00&00:07:33.02\\
			LR&10&00:00:00.02& 00:00:03.89&00:48:23.00& 00:02:04.32\\
				
				\botrule
			\end{tabular}
		\end{center}
	\end{table}
\subsection{Feature Dependency and Shapley Values}
As feature correlations reduce the validity of explainability methods \cite{10.1007/978-3-030-43823-4_17,10.1016/0031-3203(79)90049-9}, the previously explained SHAP summary plots are all generated using feature selection to avoid strong feature correlations in the dataset. Feature correlations can also make explainability more difficult \cite{10.1016/0031-3203(79)90049-9} and may lead to biologically implausible explanations. The original dataset without feature selection contains many correlated features. To compare the explanations for such a dataset \cite{Pekala2021} developed a method to consolidate correlated features to aspects and compute permutation and SHAP importances for those aspects. First, correlated features of the entire training dataset are identified using Spearman rank correlation coefficients. Hierarchical agglomerative clustering \cite{10.1016/0031-3203(79)90049-9} was used to create a dendrogram. In this work, a threshold of $H=0.5$ determining the least correlated features in a group, filtered the resulting aspects from the dendrogram. The permutation and SHAP importances are computed by jointly permuting all features in an aspect. This work uses the python package dalex v1.4.1 \cite{dalex2021} for implementation. 

The resulting aspects computed for sMCI vs. pMCI classification and FS-3 without feature selection, are shown in Table \ref{Table:Aspects}.  The 161 features of FS-3 are consolidated to 79 aspects. Of those aspects, 14 included an individual feature. Of the remaining 65 aspects, nine included more than two features. As was expected, the differences and ratios of the same region are often correlated. At least one pair of ratio and difference for the same region was included within 49 aspects. Aspect\_34 included four regions within the medial temporal lobe. Previous research showed, that those regions are important for the detection of AD progression \cite{10.1093/brain/aww243,10.1016/j.neuroimage.2015.01.032}. Aspect\_30 consolidated three ventricular regions. Previous research found that ventricular enlargement was associated with AD progression \cite{10.1016/j.neuroimage.2004.03.040,10.1212/01.WNL.0000110315.26026.EF}. Aspect\_46 included the cognitive test scores LIMMTOTAL and LDELTOTAL, and aspect\_45 included the eTIV and the gender.

\begin{longtable}{lp{7cm}}
	\caption{Aspects extracted with hierarchical agglomerative clustering for Spearman rank correlation and a threshold $H=0.5$ for sMCI vs. pMCI classification and FS-3 without feature selection.} 
	\label{Table:Aspects}\\
					Aspect&Features\\
					\hline
					
					aspect\_1& [diff\_Caudate, ratio\_Caudate]\\
					aspect\_2& [diff\_Putamen, ratio\_Putamen]\\
					aspect\_3& [diff\_inferiortemporal,  ratio\_inferiortemporal]\\
					aspect\_4& [diff\_parahippocampal,  ratio\_parahippocampal]\\
					aspect\_5& [diff\_entorhinal, ratio\_entorhinal]\\
					aspect\_6& [diff\_Lateral.Ventricle, ratio\_Lateral.Ventricle]\\
					aspect\_7& [diff\_Hippocampus, ratio\_Hippocampus]\\
					aspect\_8& [diff\_Inf.Lat.Vent, ratio\_Inf.Lat.Vent]\\
					aspect\_9& [diff\_temporalpole, ratio\_temporalpole]\\
					aspect\_10& [diff\_Amygdala, ratio\_Amygdala]\\
					aspect\_11& [diff\_posteriorcingulate,  ratio\_posteriorcingulate]\\
					aspect\_12& [diff\_CerebralWhiteMatter,  ratio\_CerebralWhiteMatter]\\
					aspect\_13& [diff\_Cortex, ratio\_Cortex]\\
					aspect\_14& [diff\_middletemporal,  ratio\_middletemporal]\\
					aspect\_15& [diff\_lingual, ratio\_lingual]\\
					aspect\_16& [diff\_cuneus, ratio\_cuneus]\\
					aspect\_17& [diff\_pericalcarine,  ratio\_pericalcarine]\\
					aspect\_18& [diff\_Thalamus.Proper, ratio\_Thalamus.Proper]\\
					aspect\_19& [diff\_Pallidum, ratio\_Pallidum]\\
					aspect\_20& [diff\_VentralDC, ratio\_VentralDC]\\
					aspect\_21& [diff\_Accumbens.area, ratio\_Accumbens.area]\\
					aspect\_22& [sum\_caudalanteriorcingulate,  sum\_rostralanteriorcingulate]\\
					aspect\_23& [sum\_frontalpole,  sum\_lateralorbitofrontal,  sum\_medialorbitofrontal,  sum\_parsorbitalis,  sum\_rostralmiddlefrontal]\\
					aspect\_24& [sum\_parsopercularis,  sum\_parstriangularis]\\
					aspect\_25& [sum\_insula,  sum\_superiortemporal,  sum\_transversetemporal]\\
					aspect\_26& [sum\_bankssts, sum\_inferiorparietal]\\
					aspect\_27& [sum\_fusiform,  sum\_inferiortemporal,  sum\_middletemporal]\\
					aspect\_28& [sum\_precuneus,  sum\_superiorparietal,  sum\_supramarginal]\\
					aspect\_29& [sum\_caudalmiddlefrontal,  sum\_paracentral,  sum\_postcentral,  sum\_precentral,  sum\_superiorfrontal,  sum\_Cortex]\\
					sum\_posteriorcingulate& [sum\_posteriorcingulate]\\
					aspect\_30& [X3rd.Ventricle,  sum\_Inf.Lat.Vent,  sum\_Lateral.Ventricle]\\
					sum\_Accumbens.area& [sum\_Accumbens.area]\\
					AGE& [AGE]\\
					CSF& [CSF]\\
					sum\_CerebralWhiteMatter& [sum\_CerebralWhiteMatter]\\
					aspect\_31& [Brain.Stem,  sum\_Cerebellum.Cortex,  sum\_Cerebellum.White.Matter]\\
					aspect\_32& [sum\_Thalamus.Proper, sum\_VentralDC]\\
					aspect\_33& [sum\_Pallidum, sum\_Putamen]\\
					aspect\_34& [sum\_entorhinal,  sum\_parahippocampal,  sum\_Amygdala,  sum\_Hippocampus]\\
					sum\_temporalpole& [sum\_temporalpole]\\
					aspect\_35& [sum\_cuneus,  sum\_lingual,  sum\_pericalcarine]\\
					sum\_isthmuscingulate& [sum\_isthmuscingulate]\\
					sum\_lateraloccipital& [sum\_lateraloccipital]\\
					aspect\_36& [diff\_fusiform, ratio\_fusiform]\\
					aspect\_37& [diff\_lateraloccipital,  ratio\_lateraloccipital]\\
					aspect\_38& [diff\_supramarginal,  ratio\_supramarginal]\\
					aspect\_39& [diff\_inferiorparietal,  ratio\_inferiorparietal]\\
					aspect\_40& [diff\_superiorparietal,  ratio\_superiorparietal]\\
					aspect\_41& [diff\_precuneus, ratio\_precuneus]\\
					aspect\_42& [diff\_bankssts, ratio\_bankssts]\\
					aspect\_43& [diff\_superiortemporal,  ratio\_superiortemporal]\\
					aspect\_44& [diff\_parsorbitalis,  ratio\_parsorbitalis]\\
					sum\_vessel& [sum\_vessel]\\
					aspect\_45& [EstimatedTotalIntraCranial, PTGENDER\_Female]\\
					sum\_Caudate& [sum\_Caudate]\\
					aspect\_46& [LIMMTOTAL, LDELTOTAL]\\
					MMSCORE& [MMSCORE]\\
					PTEDUCAT& [PTEDUCAT]\\
					aspect\_47& [diff\_parstriangularis,  ratio\_parstriangularis]\\
					aspect\_48& [diff\_parsopercularis,  ratio\_parsopercularis]\\
					aspect\_49& [diff\_caudalmiddlefrontal,  ratio\_caudalmiddlefrontal]\\
					aspect\_50& [diff\_rostralmiddlefrontal,  ratio\_rostralmiddlefrontal]\\
					aspect\_51& [diff\_precentral, ratio\_precentral]\\
					X4th.Ventricle& [X4th.Ventricle]\\
					APOE4& [APOE4]\\
					aspect\_52& [diff\_postcentral, ratio\_postcentral]\\
					aspect\_53& [diff\_transversetemporal,  ratio\_transversetemporal]\\
					aspect\_54& [diff\_rostralanteriorcingulate,  ratio\_rostralanteriorcingulate]\\
					aspect\_55& [diff\_superiorfrontal,  ratio\_superiorfrontal]\\
					aspect\_56& [diff\_caudalanteriorcingulate,  ratio\_caudalanteriorcingulate]\\
					aspect\_57& [diff\_medialorbitofrontal,  ratio\_medialorbitofrontal]\\
					aspect\_58& [diff\_frontalpole, ratio\_frontalpole]\\
					aspect\_59& [diff\_vessel, ratio\_vessel]\\
					aspect\_60& [diff\_lateralorbitofrontal,  ratio\_lateralorbitofrontal]\\
					aspect\_61& [diff\_insula, ratio\_insula]\\
					aspect\_62& [diff\_isthmuscingulate,  ratio\_isthmuscingulate]\\
					aspect\_63& [diff\_paracentral, ratio\_paracentral]\\
					aspect\_64& [diff\_Cerebellum.Cortex, ratio\_Cerebellum.Cortex]\\
					aspect\_65& [diff\_Cerebellum.White.Matter,  ratio\_Cerebellum.White.Matter]\\
					\hline
		
\end{longtable}
 
\begin{figure}
	\centering
	\begin{subfigure}[b]{0.49\textwidth}
		\centering
		\includegraphics[width=\textwidth]{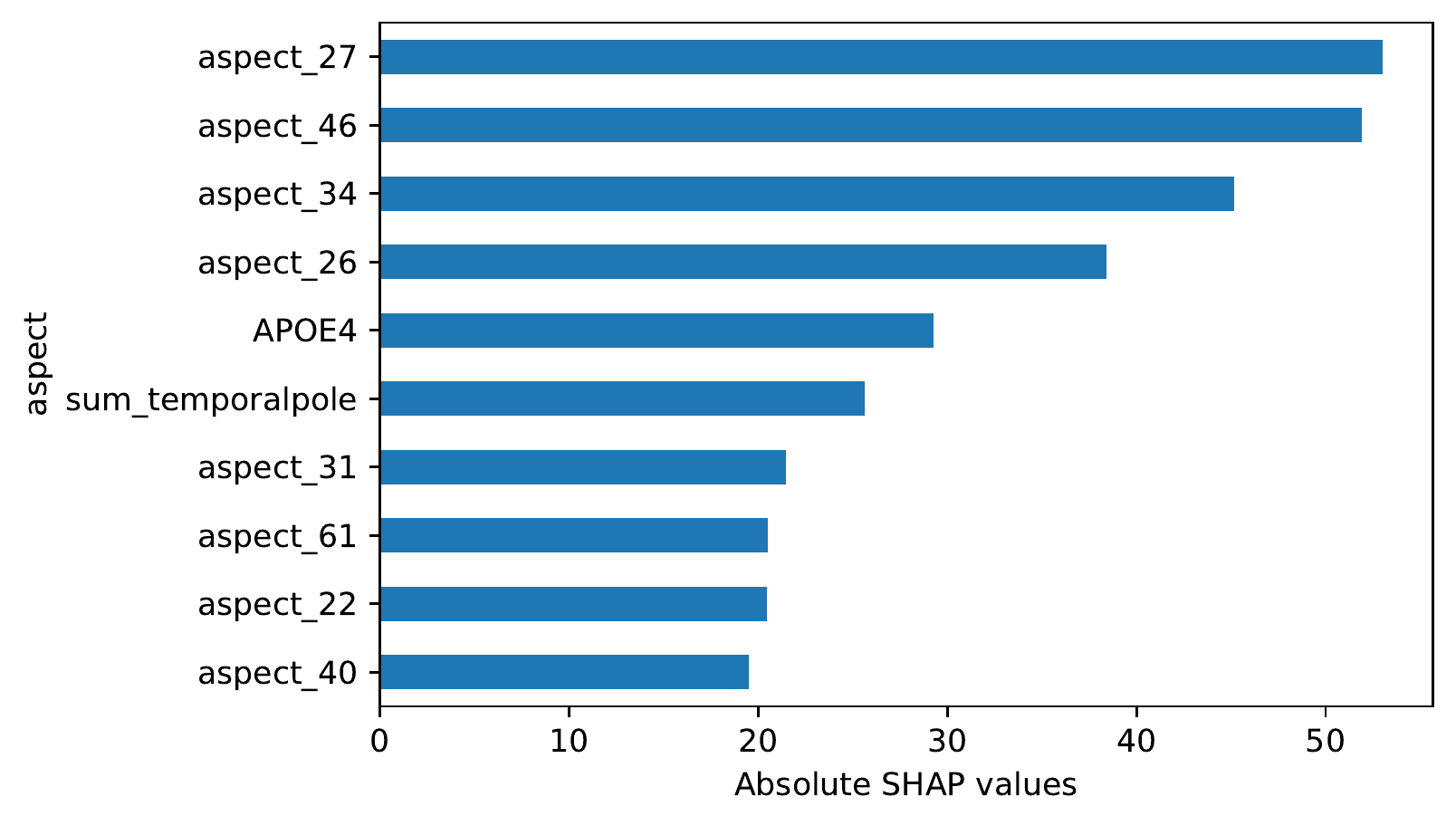}
		\caption{LR}
		\label{fig:SHAP_LR_Model_Aspect}
	\end{subfigure}\hfill 
\begin{subfigure}[b]{0.49\textwidth}
	\centering
	\includegraphics[width=\textwidth]{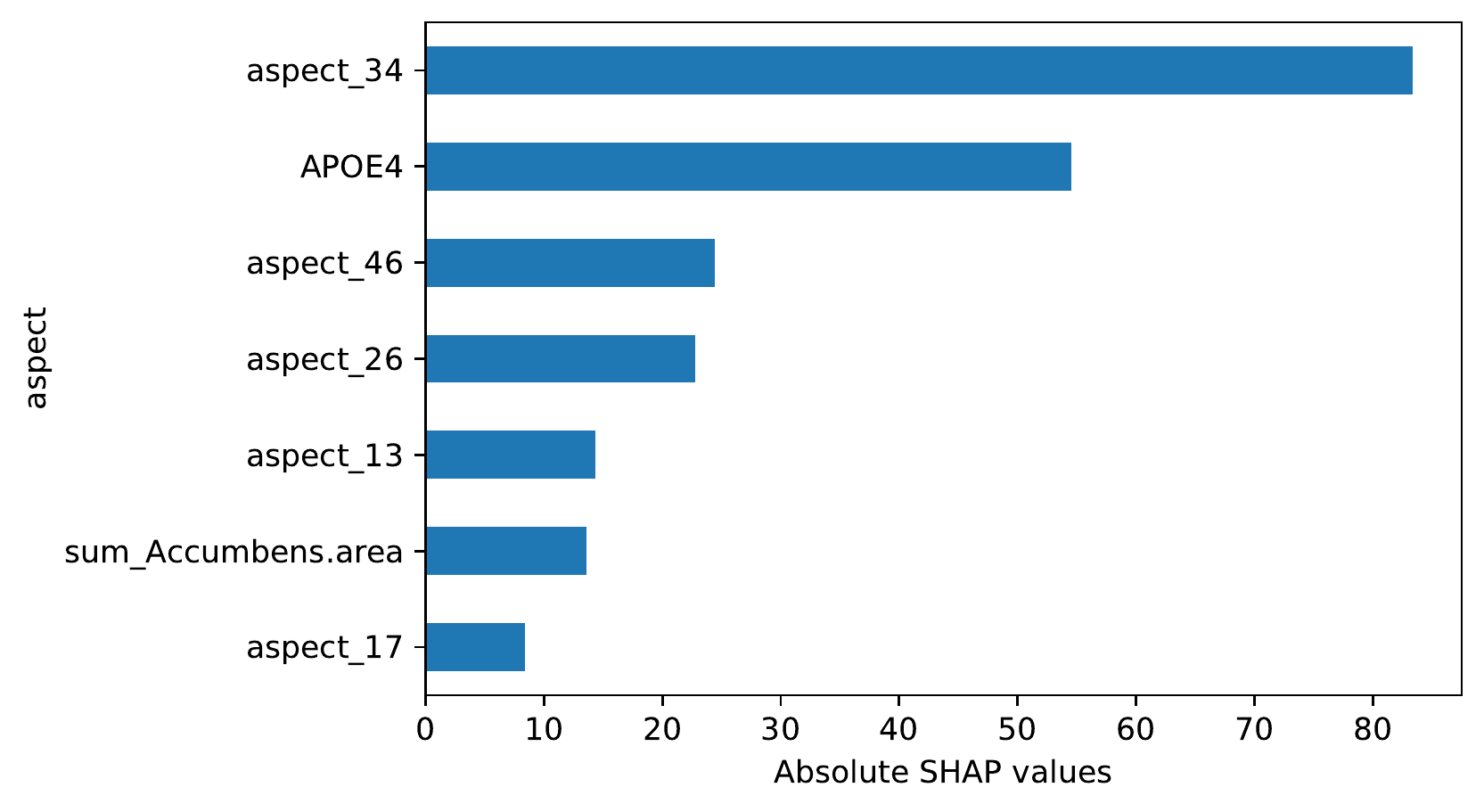}
	\caption{DT}
	\label{fig:SHAP_DT_Model_Aspect}
\end{subfigure}\\\vspace{4pt}
	\begin{subfigure}[b]{0.49\textwidth}
		\centering
		\includegraphics[width=\textwidth]{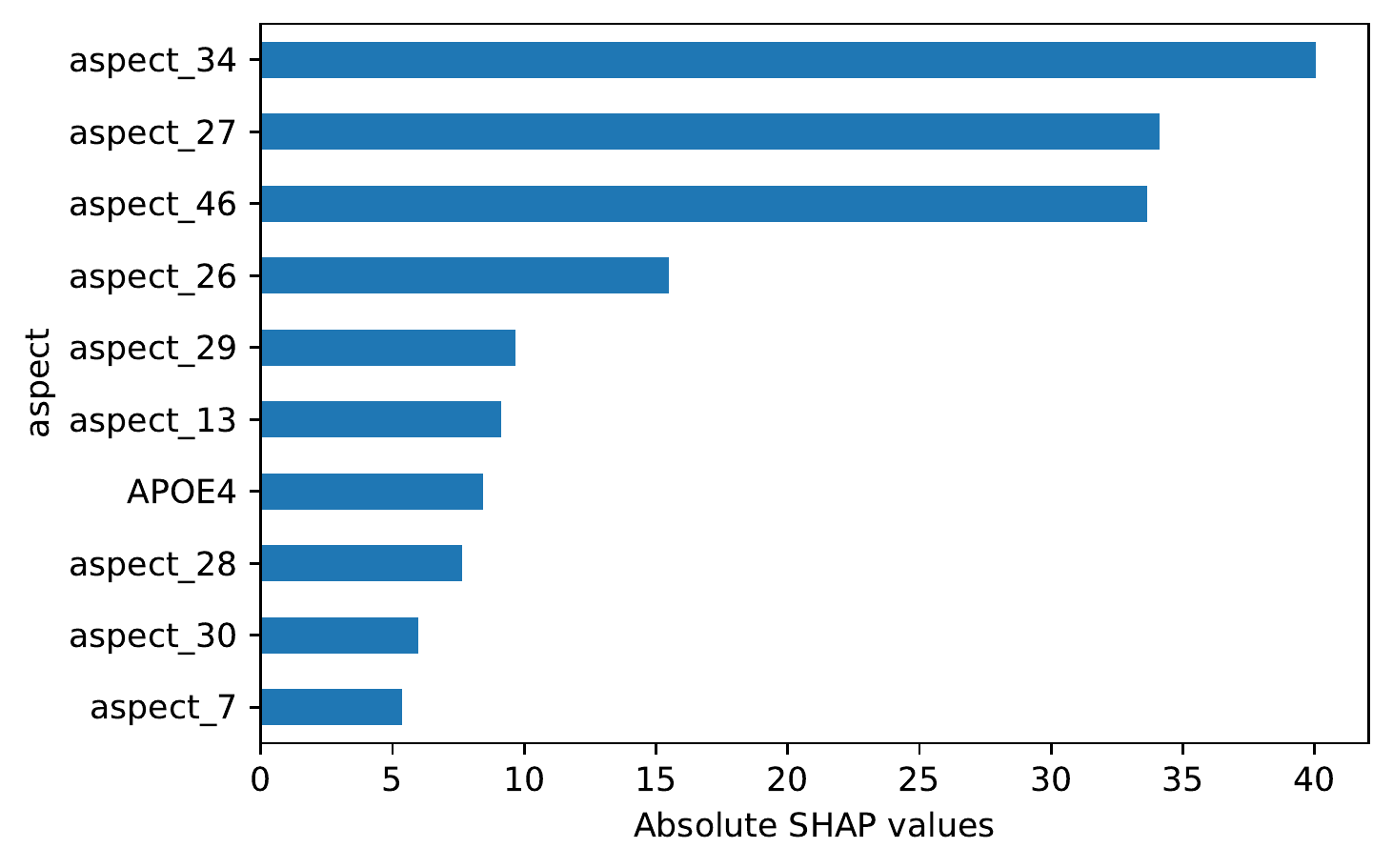}
		\caption{RF}
		\label{fig:SHAP_RF_Model_Aspect}
	\end{subfigure}
	\hfill
	\begin{subfigure}[b]{0.49\textwidth}
		\centering
		\includegraphics[width=\textwidth]{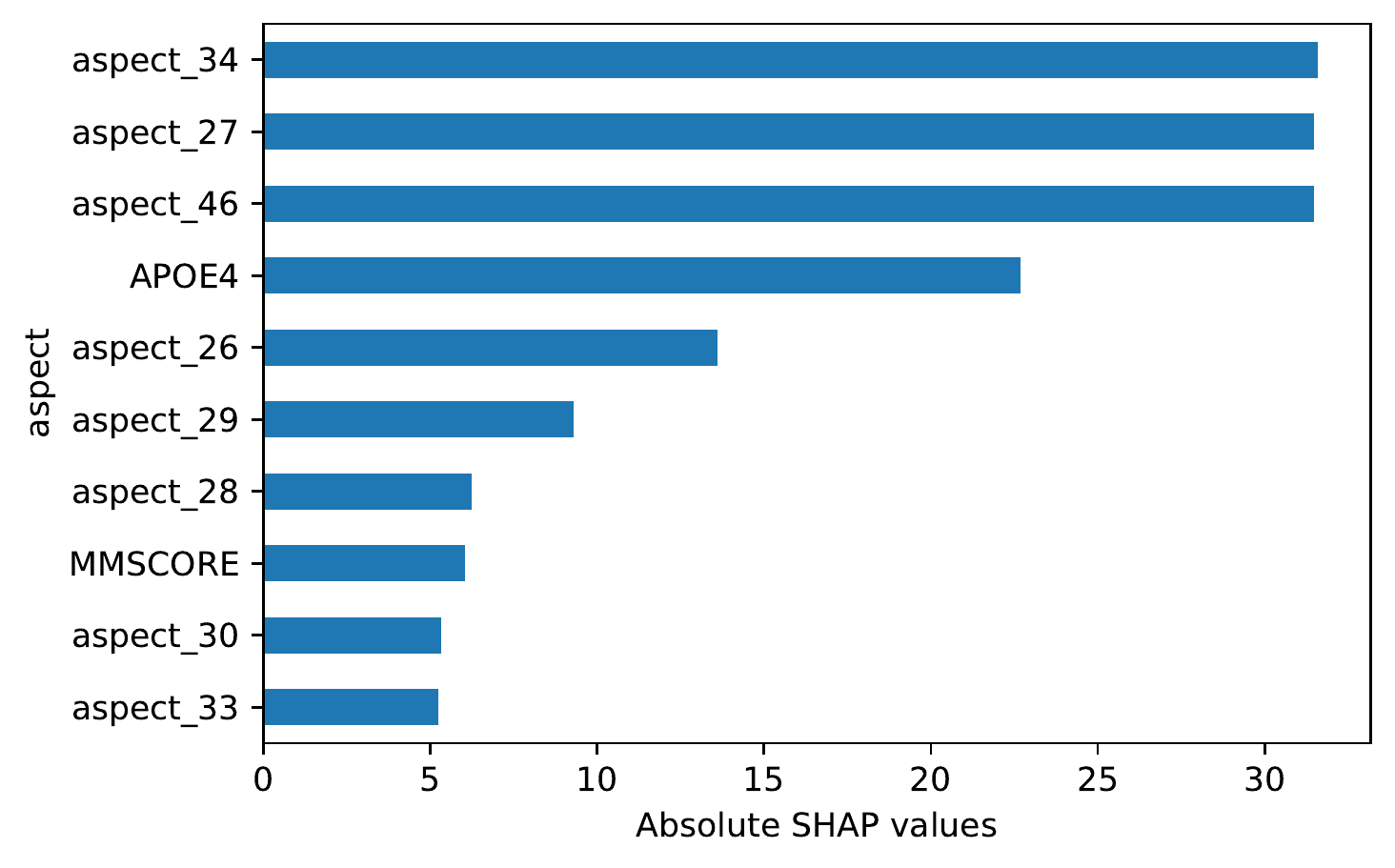}
		\caption{XGBoost}
		\label{fig:SHAP_XGBoost_Model_Aspect}
	\end{subfigure}
	\\\vspace{4pt} 
	\begin{subfigure}[b]{0.49\textwidth}
		\centering
		\includegraphics[width=\textwidth]{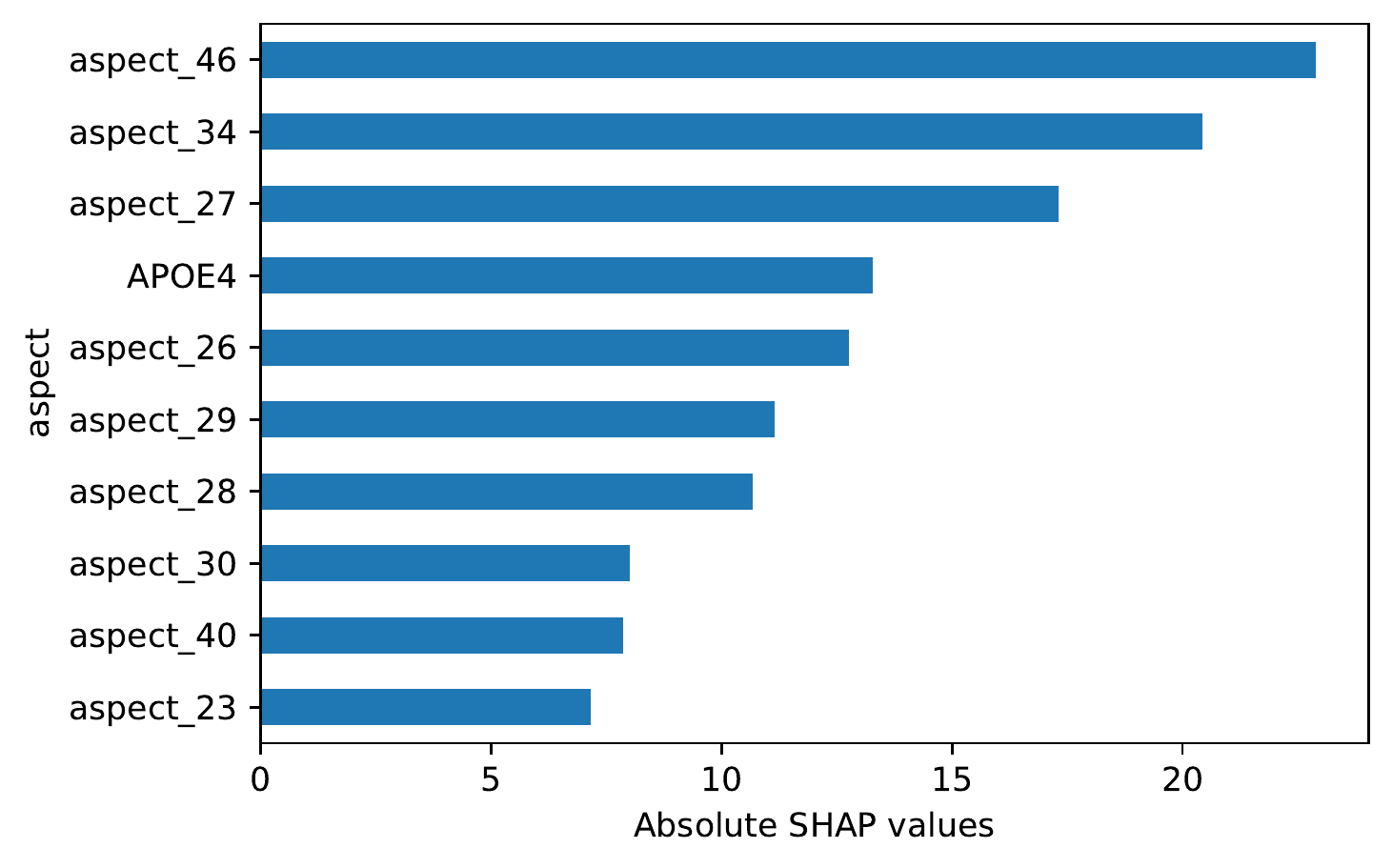}
		\caption{SVM poly}
		\label{fig:SHAP_SVMPoly_Model_Aspect}
	\end{subfigure}
	\hfill
	\begin{subfigure}[b]{0.49\textwidth}
		\centering
		\includegraphics[width=\textwidth]{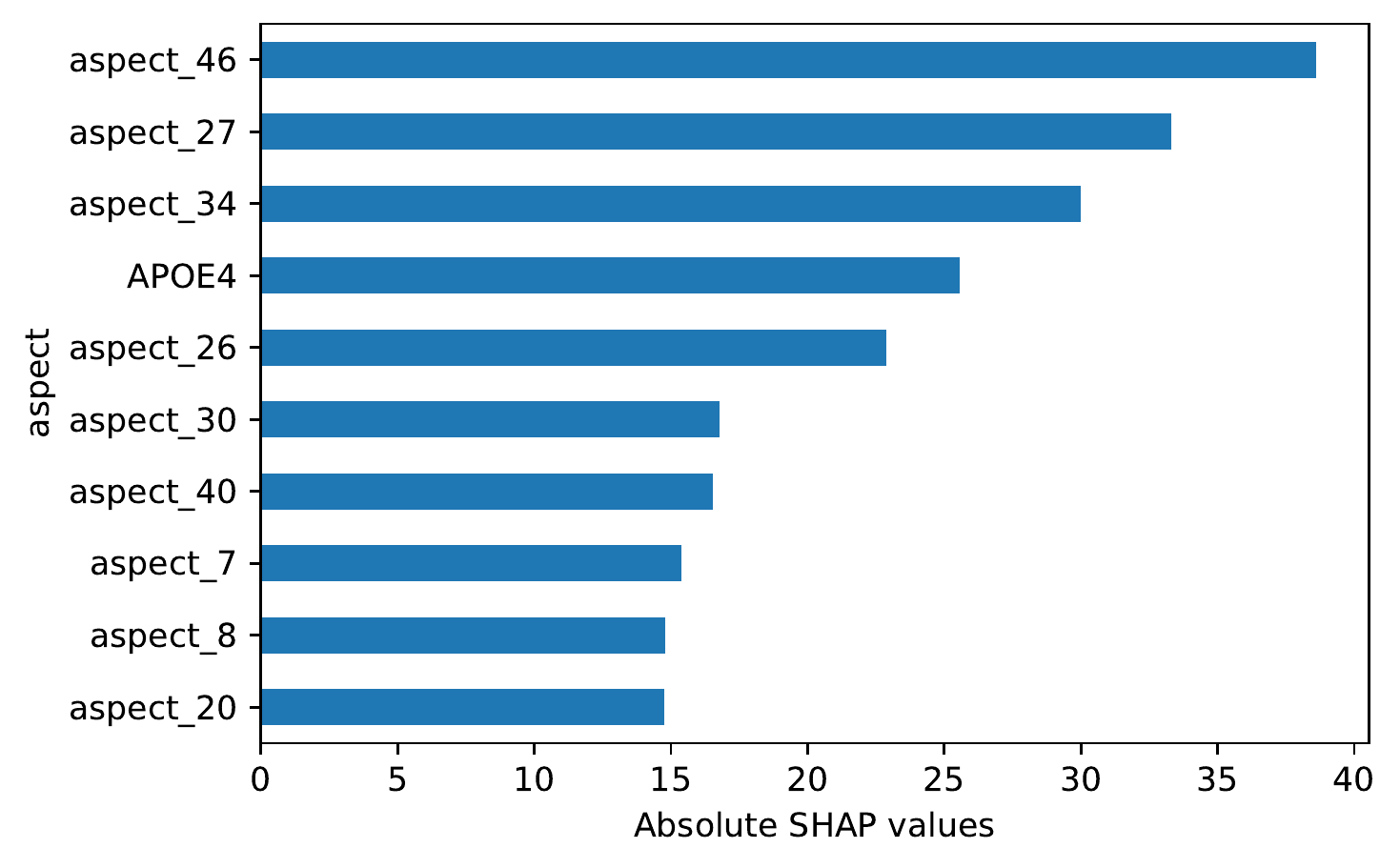}
		\caption{SVM radial}
		\label{fig:SHAP_SVMRBF_Model_Aspect}
	\end{subfigure}
	\caption{SHAP aspect importance plots for all six ML models trained to distinguish between sMCI and pMCI subjects using FS-3 and no feature selection. Each plot shows a different classification model}
	\label{fig:Comparison_ClassificationModels_sMCI_FS-3_RFMean_SHAP_Aspects}
\end{figure}

Using those aspects, the SHAP importances visualized in Figure \ref{fig:Comparison_ClassificationModels_sMCI_FS-3_RFMean_SHAP_Aspects} were computed for the sMCI vs. pMCI classification without feature selection and for all ML models. The most important aspect for the RF, the XGBoost, and the DT was aspect\_34, which consolidated the entorhinal cortices, the parahippocampal gyri, the amygdalae, and the hippocampi. Those brain areas were associated with AD in previous research \cite{10.1093/brain/aww243,10.1016/j.neuroimage.2015.01.032}. Aspect\_34 also was the second most important aspect of the polynomial SVM and the third most important aspect in the LR and radial SVM. The most important aspect for the LR was aspect\_27 which consolidated volumes of the fusiform, the inferior temporal, and the middle temporal gyri. This aspect also reached the second rank for the RF, radial SVM, and XGBoost models as well as the third place for the polynomial SVM. Previous research \cite{10.1371/journal.pone.0048973,10.1093/brain/aww243} showed that those regions are affected in early AD stages. Aspect\_46 consolidated the LDELTOTAL and LIMMTOTAL cognitive test scores and was the most important aspect for both SVMs. This aspect also achieved the third rank for the XGBoost, RF and DT models, and the second rank for the LR.
Overall, the most important aspects of the different models seem to be similar for the ML models.

\begin{figure}
	\centering
	\begin{subfigure}[b]{0.49\textwidth}
		\centering
		\includegraphics[width=\textwidth]{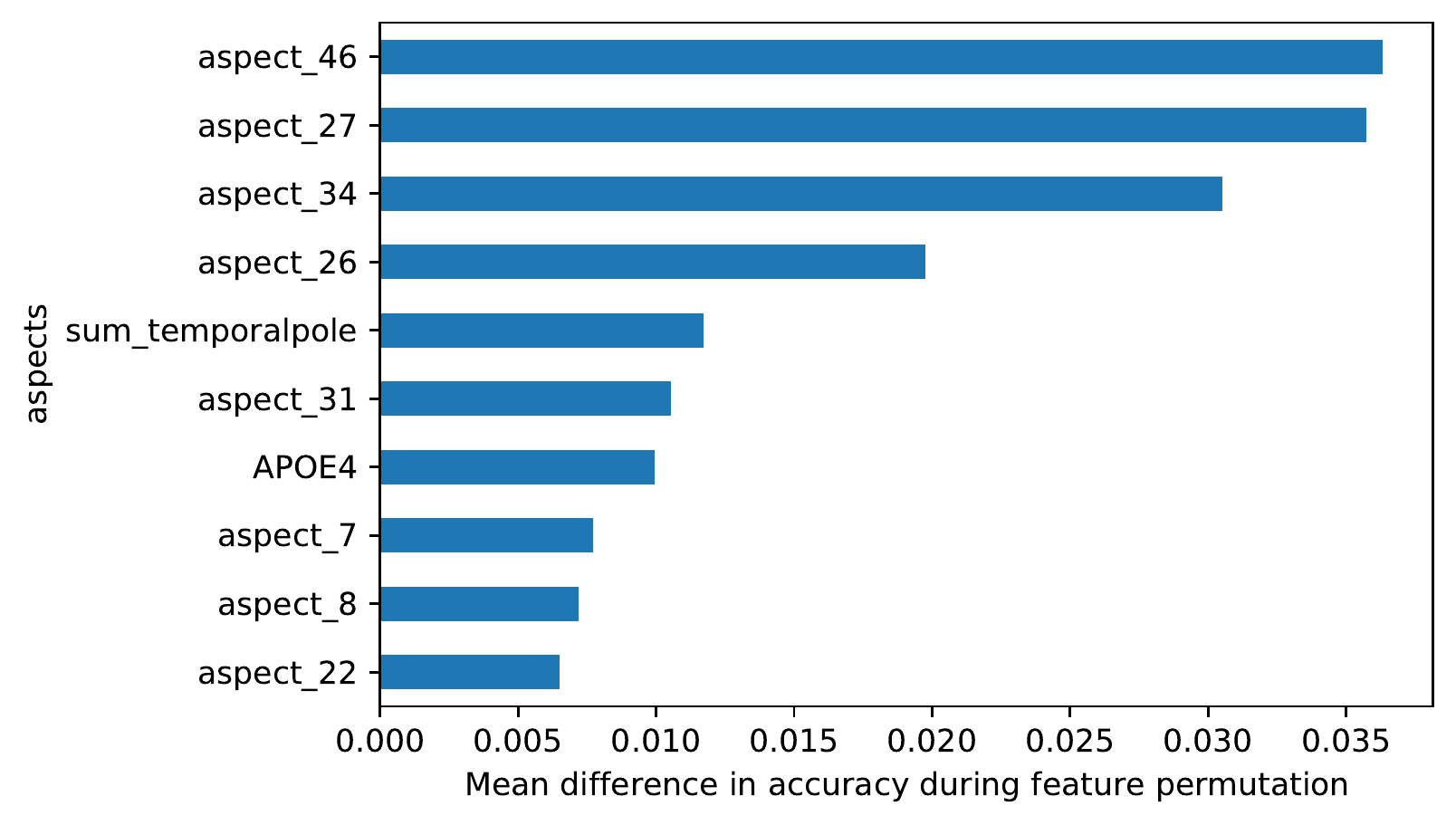}
		\caption{LR}
		\label{fig:PIP_LR_Model_Aspect}
	\end{subfigure}\hfill
\begin{subfigure}[b]{0.49\textwidth}
	\centering
	\includegraphics[width=\textwidth]{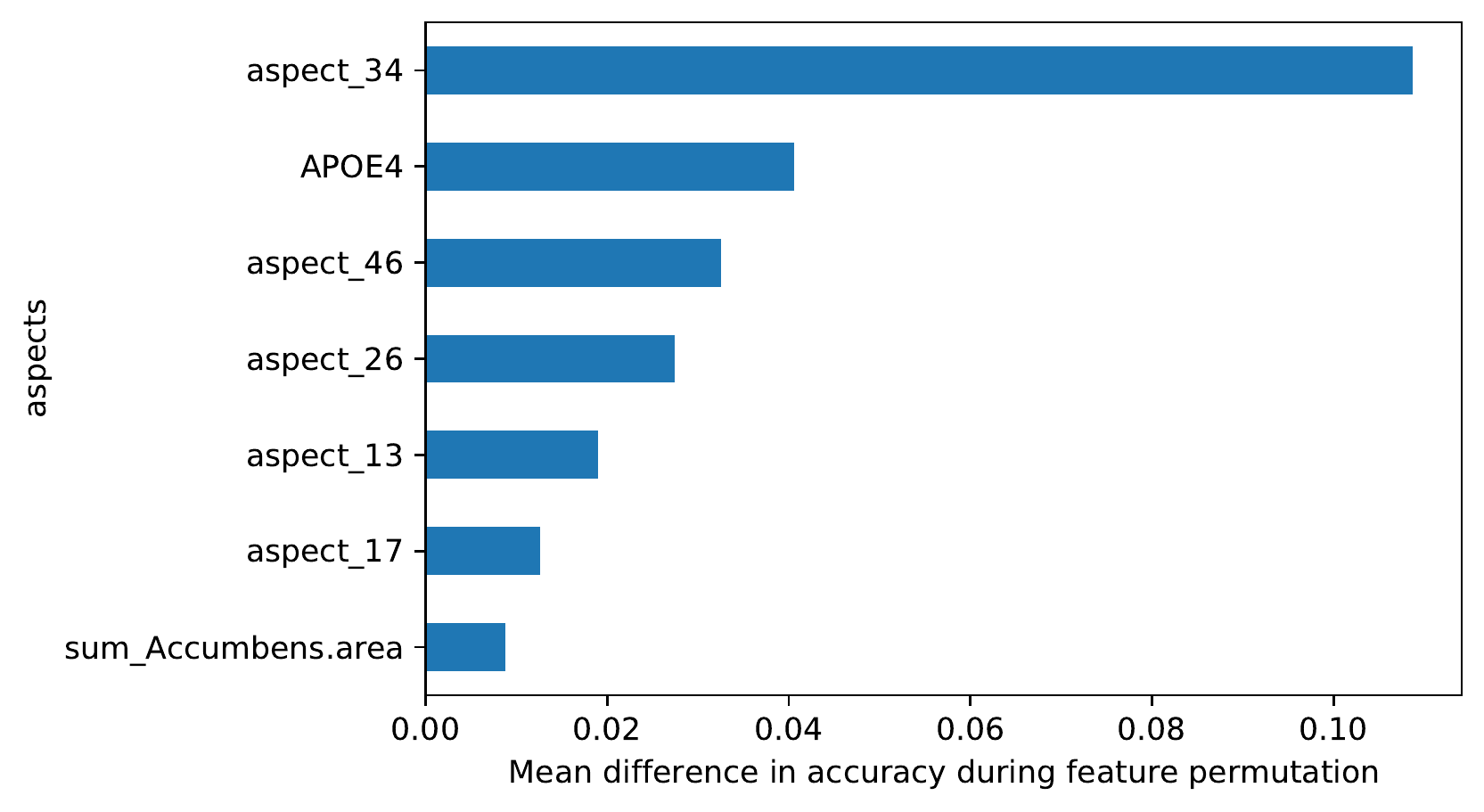}
	\caption{DT}
	\label{fig:PIP_DT_Model_Aspect}
\end{subfigure}\\\vspace{4pt} 
	\begin{subfigure}[b]{0.49\textwidth}
		\centering
		\includegraphics[width=\textwidth]{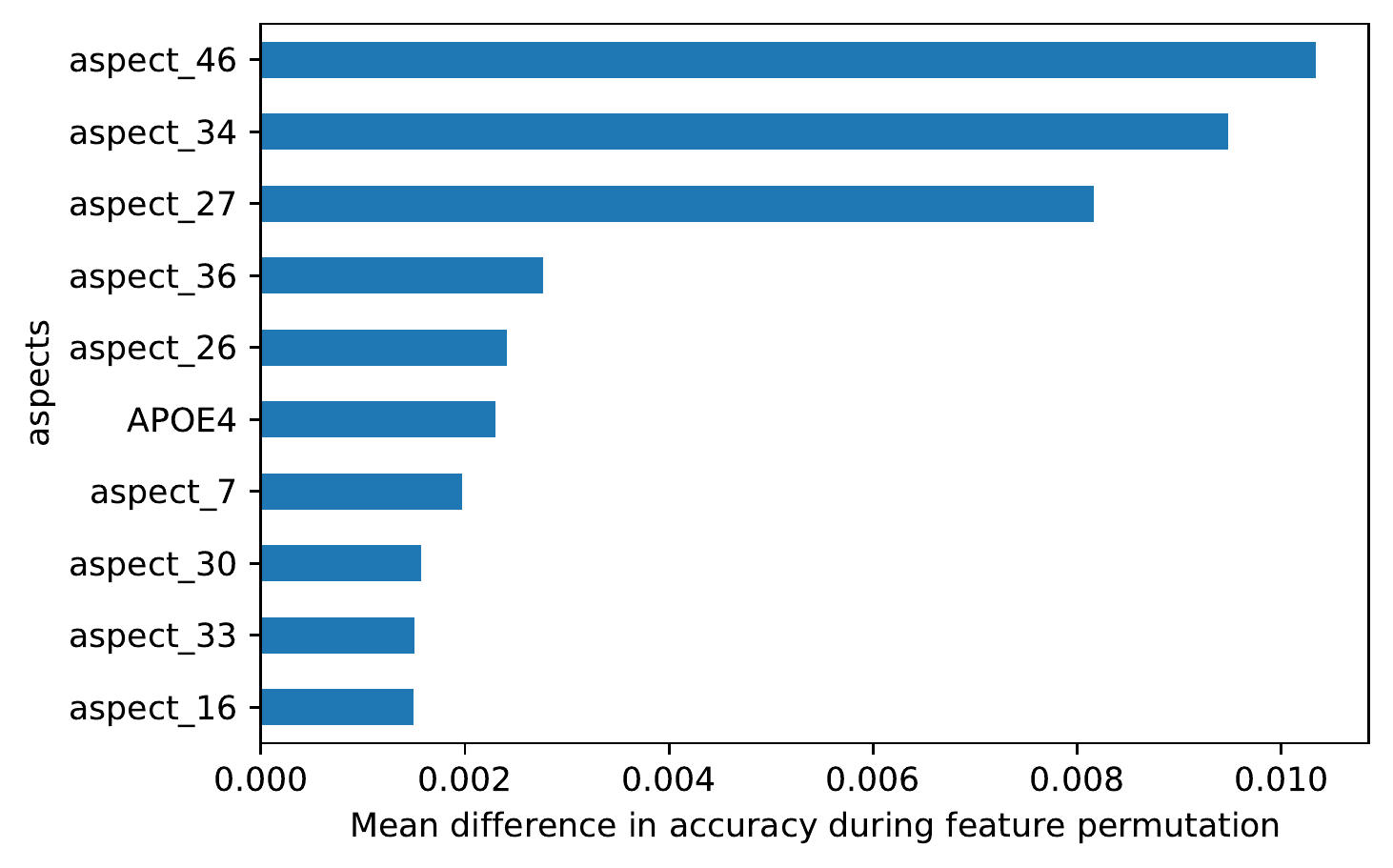}
		\caption{RF}
		\label{fig:PIP_RF_Model_Aspect}
	\end{subfigure}
	\hfill
	\begin{subfigure}[b]{0.49\textwidth}
		\centering
		\includegraphics[width=\textwidth]{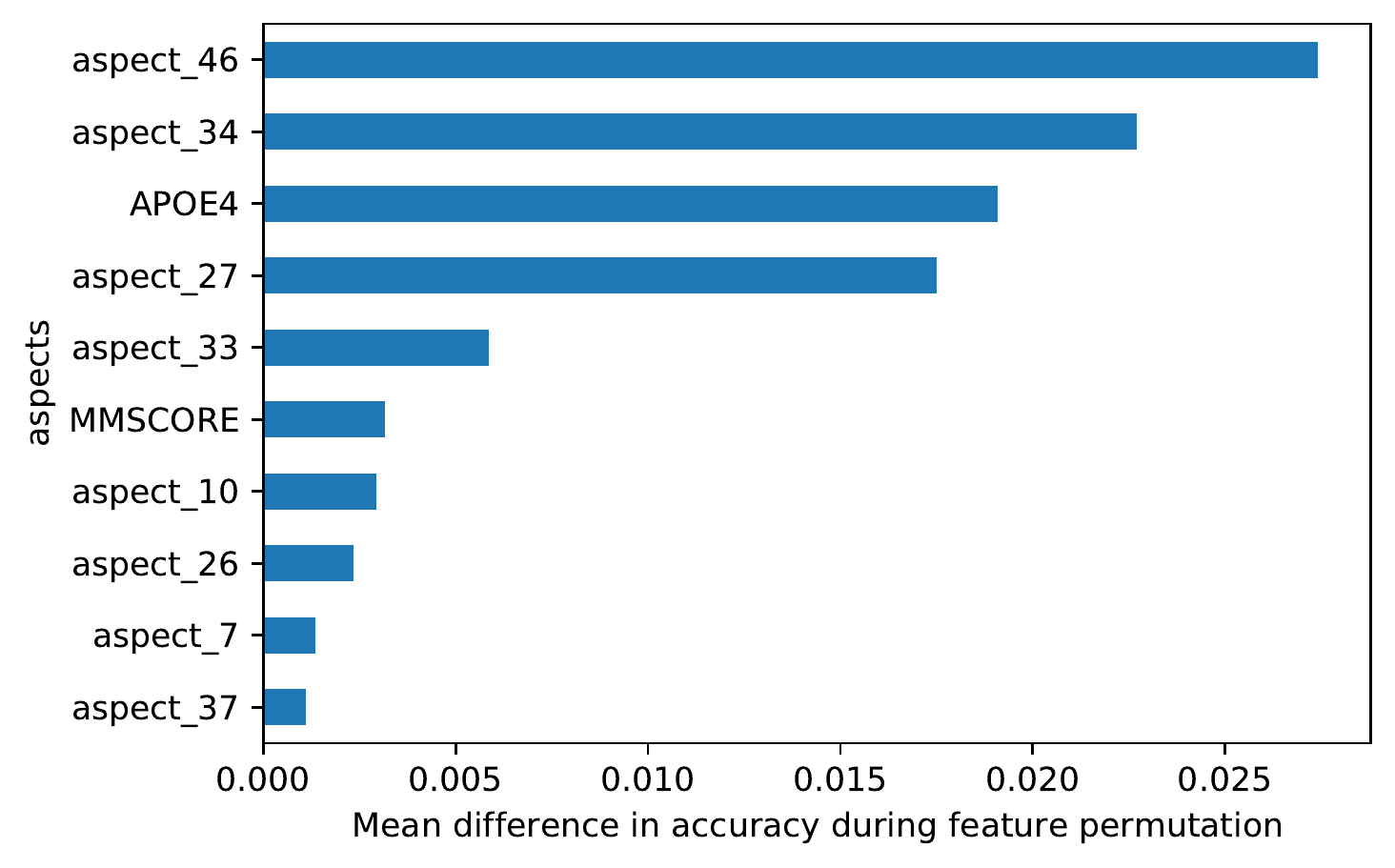}
		\caption{XGBoost}
		\label{fig:PIP_XGBoost_Model_Aspect}
	\end{subfigure}
	\\\vspace{4pt} 
	\begin{subfigure}[b]{0.49\textwidth}
		\centering
		\includegraphics[width=\textwidth]{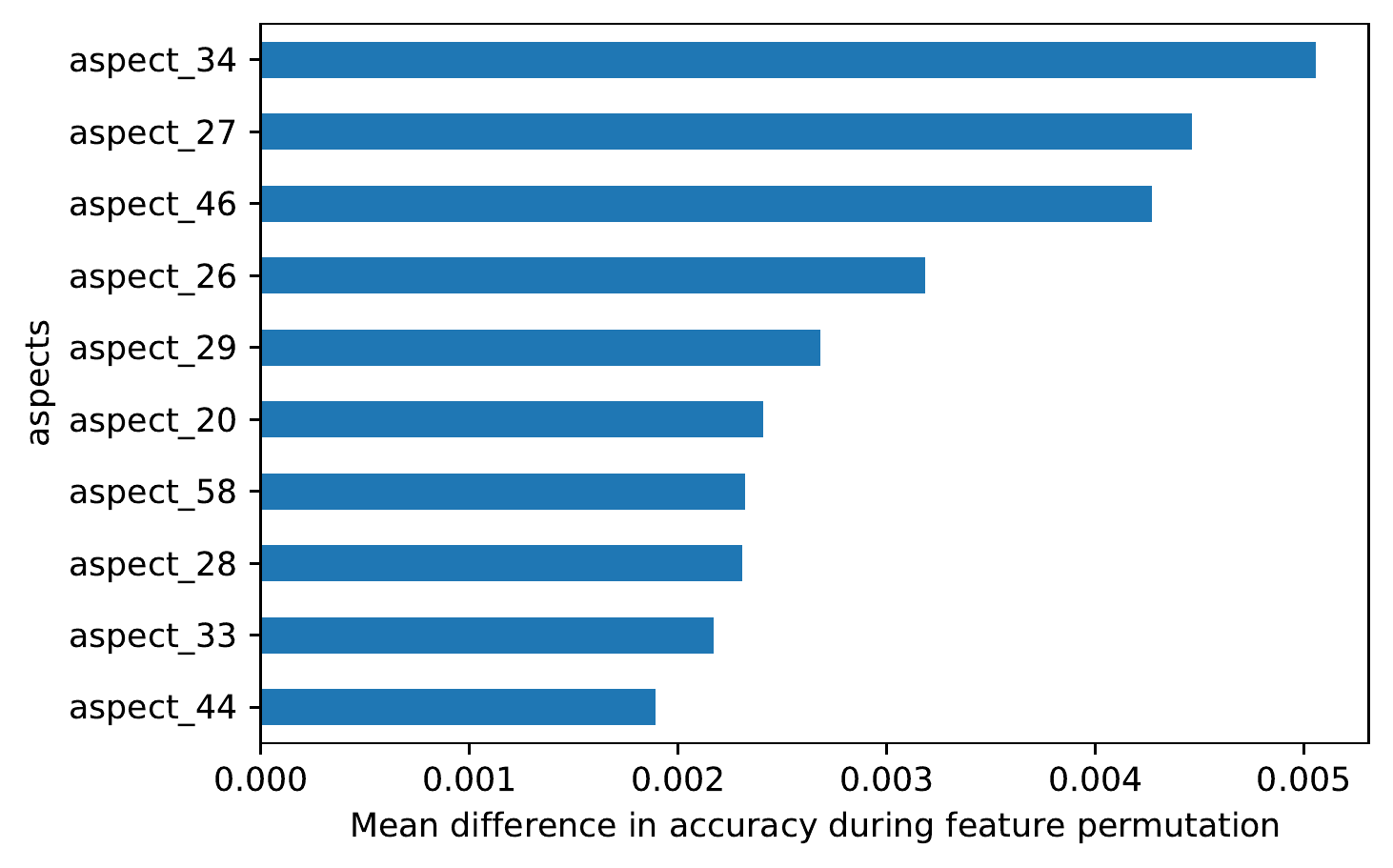}
		\caption{SVM poly}
		\label{fig:PIP_SVMPoly_Model_Aspect}
	\end{subfigure}
	\hfill
	\begin{subfigure}[b]{0.49\textwidth}
		\centering
		\includegraphics[width=\textwidth]{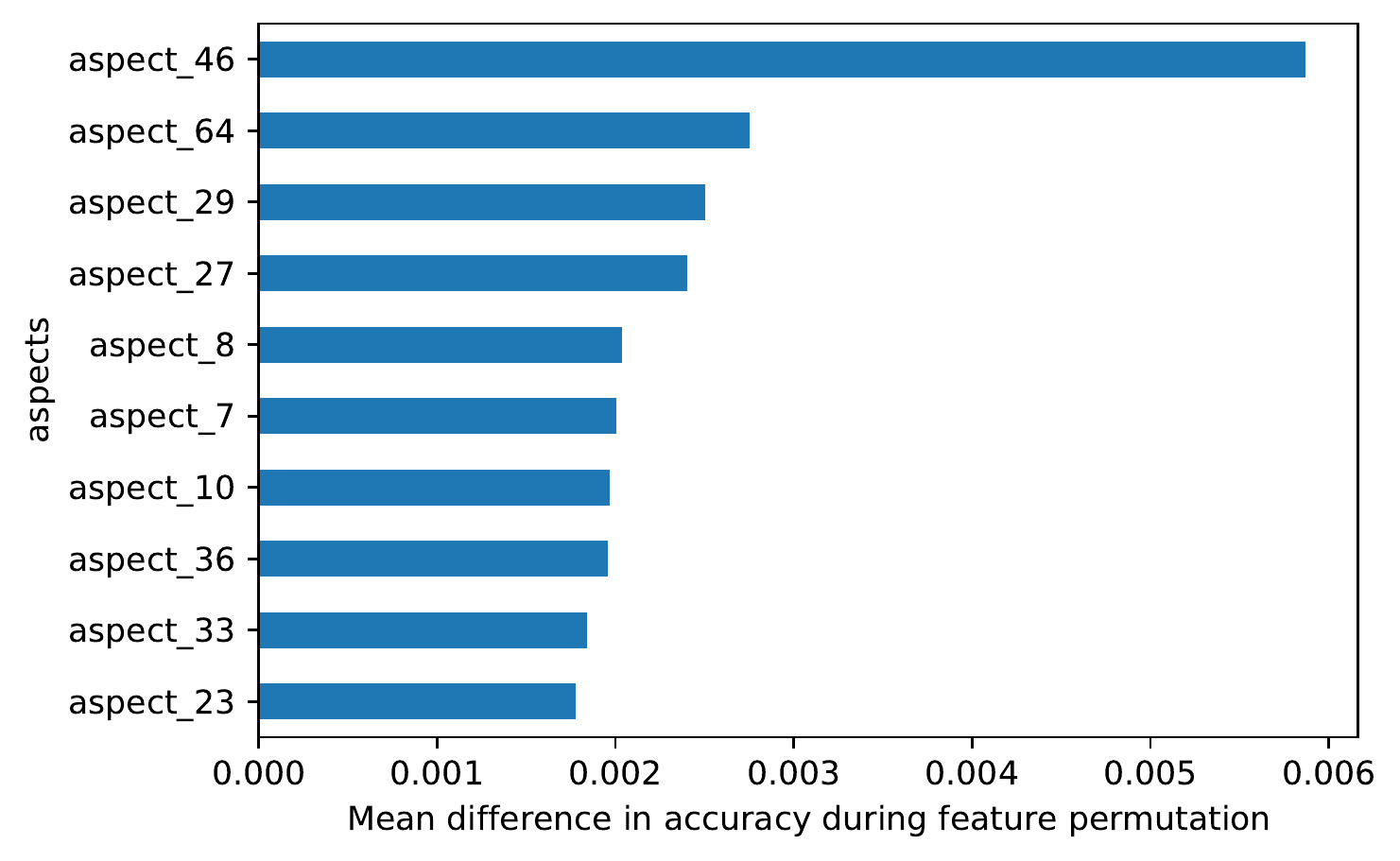}
		\caption{SVM radial}
		\label{fig:PIP_SVMRBF_Model_Aspect}
	\end{subfigure}
	\caption{Permutation aspect importance plots for all six ML models trained to distinguish between sMCI and pMCI subjects using FS-3 and no feature selection. Each plot shows a different classification model}
	\label{fig:Comparison_ClassificationModels_sMCI_FS-3_RFMean_PIP_Aspects}
\end{figure}

Figure \ref{fig:Comparison_ClassificationModels_sMCI_FS-3_RFMean_PIP_Aspects} shows the aspect permutation importance plots of the previously described models. The most important aspects chosen by SHAP importance and permutation importance matched for the DT, and the radial SVM. The most important aspect of the RF and XGBoost models was aspect\_46 which included the cognitive test scores LDELTOTAL and LIMMTOTAL. This aspect reached the third rank using the SHAP method for both models. For the LR, aspect\_46 was also identified as the most important aspect. This aspect reached second place using the SHAP explanations. The highest permutation importance of the SVM was reached for aspect\_34. This aspect reached second place using the SHAP method.

\begin{figure}
	\begin{center}
		\includegraphics[width=0.7\textwidth]{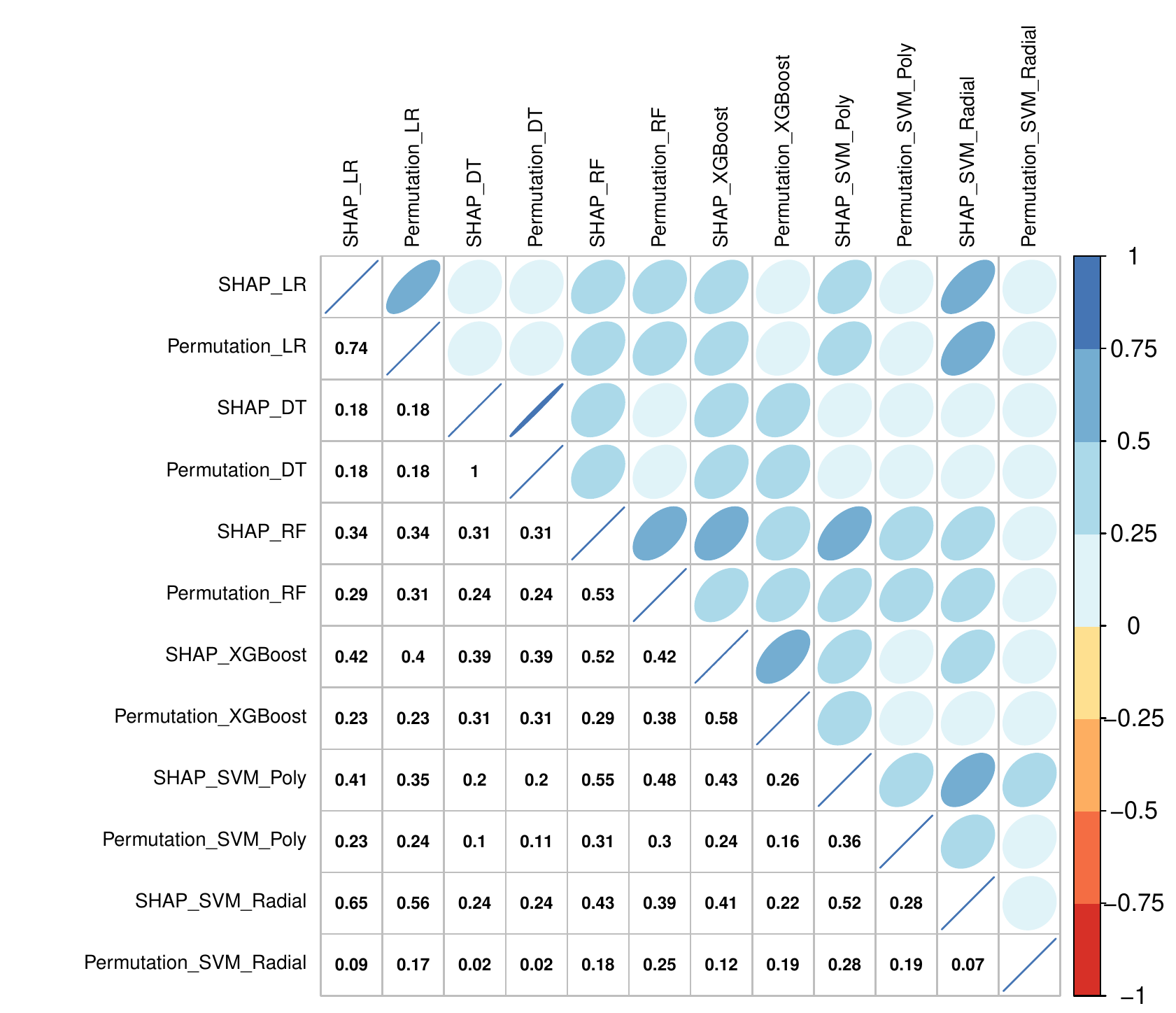}
		\caption{Plot showing Kendall's tau correlation between aspect importances of all SHAP models and permutation importance for FS-3, no feature selection, sMCI vs. pMCI classification, and the ADNI and AIBL datasets (\textit{n=747})}
		\label{fig:Correlation_plot_natural_feature_importances_Aspect} 
	\end{center}
\end{figure}

To investigate the correlation of the feature rankings, between the different methods, Kendall's tau rank correlation between the SHAP feature ranking and the permutation method is visualized in Figure \ref{fig:Correlation_plot_natural_feature_importances_Aspect}. A very strong correlation of 1.00 was observed between the permutation importance and the SHAP values of the DT. A moderate correlation of 0.53 was observed between the SHAP values and the permutation importance of the RF. The XGBoost SHAP rankings also showed a moderate correlation of 0.58 to the permutation importance of the same method. The SHAP values of the polynomial SVM are weakly correlated (0.36) to the model's permutation importance. A very weak correlation of 0.07 was observed between the SHAP values of the radial SVM and their permutation importance rankings. The LR SHAP values are strongly correlated (0.74) to the permutation importance measurements of this model. 

The inter-model correlations of the SHAP values showed a moderate correlation of 0.52 between the polynomial SVM and the radial SVM, as well as a strong correlation (0.65) between the SHAP values of the LR and the radial SVM SHAP values. The SHAP values of the RF was moderately correlated to the XGBoost SHAP values (0.52) and the SHAP values of the polynomial SVM. The SHAP values of the DT and LR model showed a very weak correlation (0.18). Overall the SHAP values of the DT showed only weak correlations to the remaining ML models. The highest correlation of 0.39 was found for the SHAP values of the XGBoost model.

\subsection{Explanations of Individual Predictions}
\begin{figure}
	\centering
	\begin{subfigure}[b]{0.49\textwidth}
		\centering
		\includegraphics[width=\textwidth]{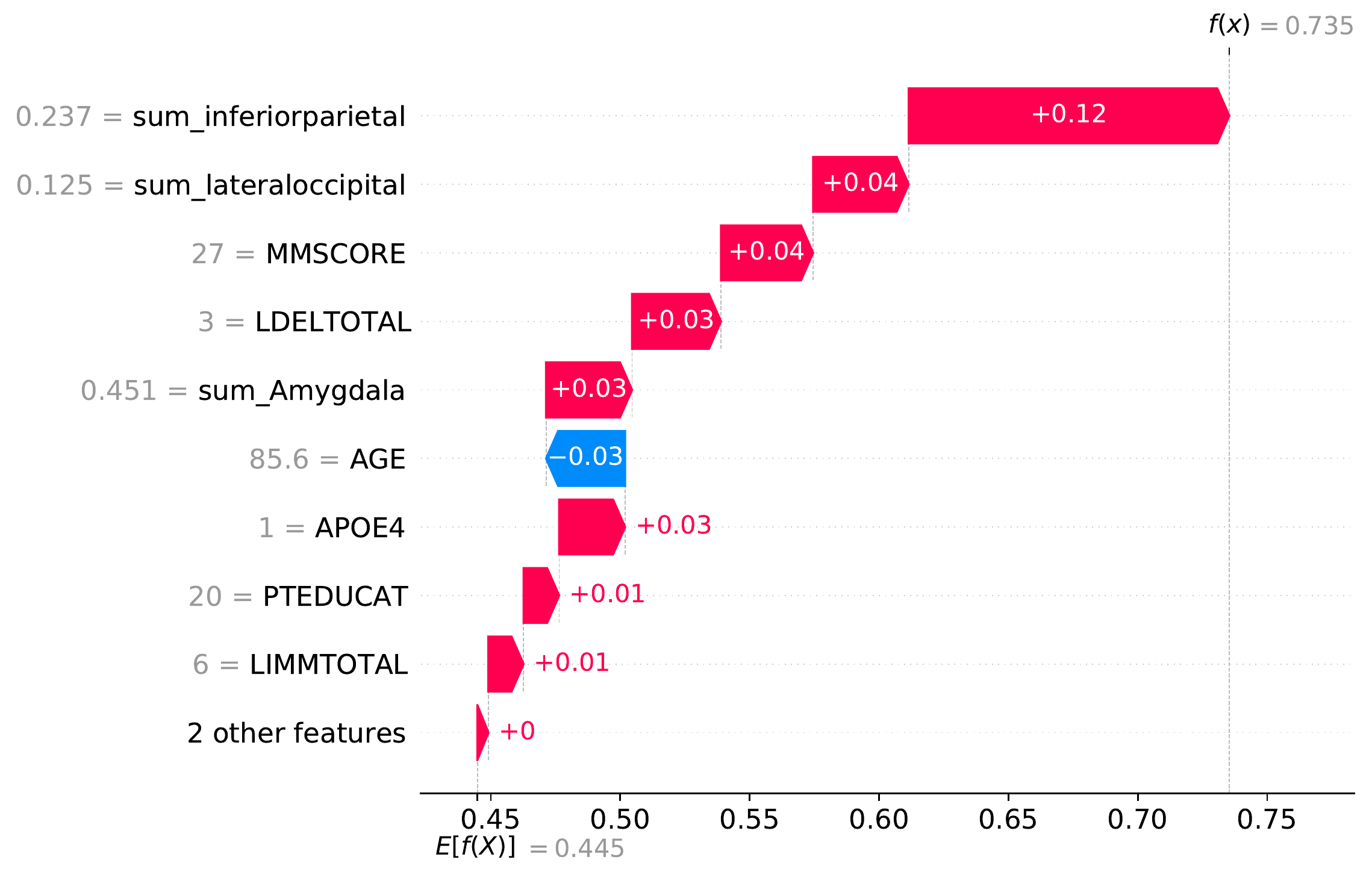}
		\caption{PTID: 027\_S\_1387, diagnosis: pMCI, model prediction value: 0.735}
		\label{fig:SHAP_Force_plot_027_S_0256}
	\end{subfigure} \hfill
	\begin{subfigure}[b]{0.49\textwidth}
		\centering
		\includegraphics[width=\textwidth]{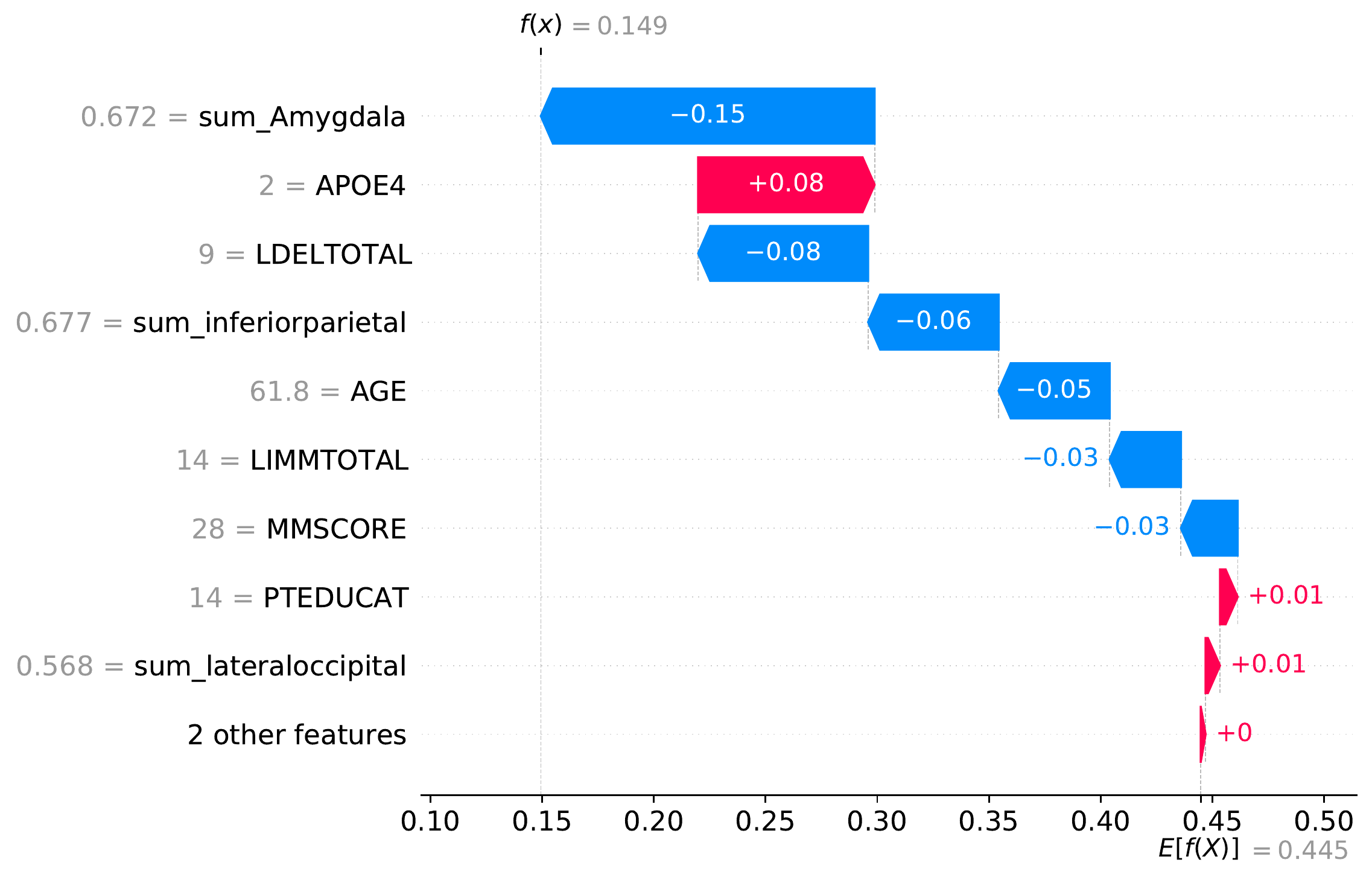}
		\caption{PTID: 037\_S\_4146, diagnosis: sMCI, model prediction value: 0.149}
		\label{fig:SHAP_Force_plot_941_S_4036}
	\end{subfigure}\\\vspace{4pt} 
	\begin{subfigure}[b]{0.49\textwidth}
		\centering
		\includegraphics[width=\textwidth]{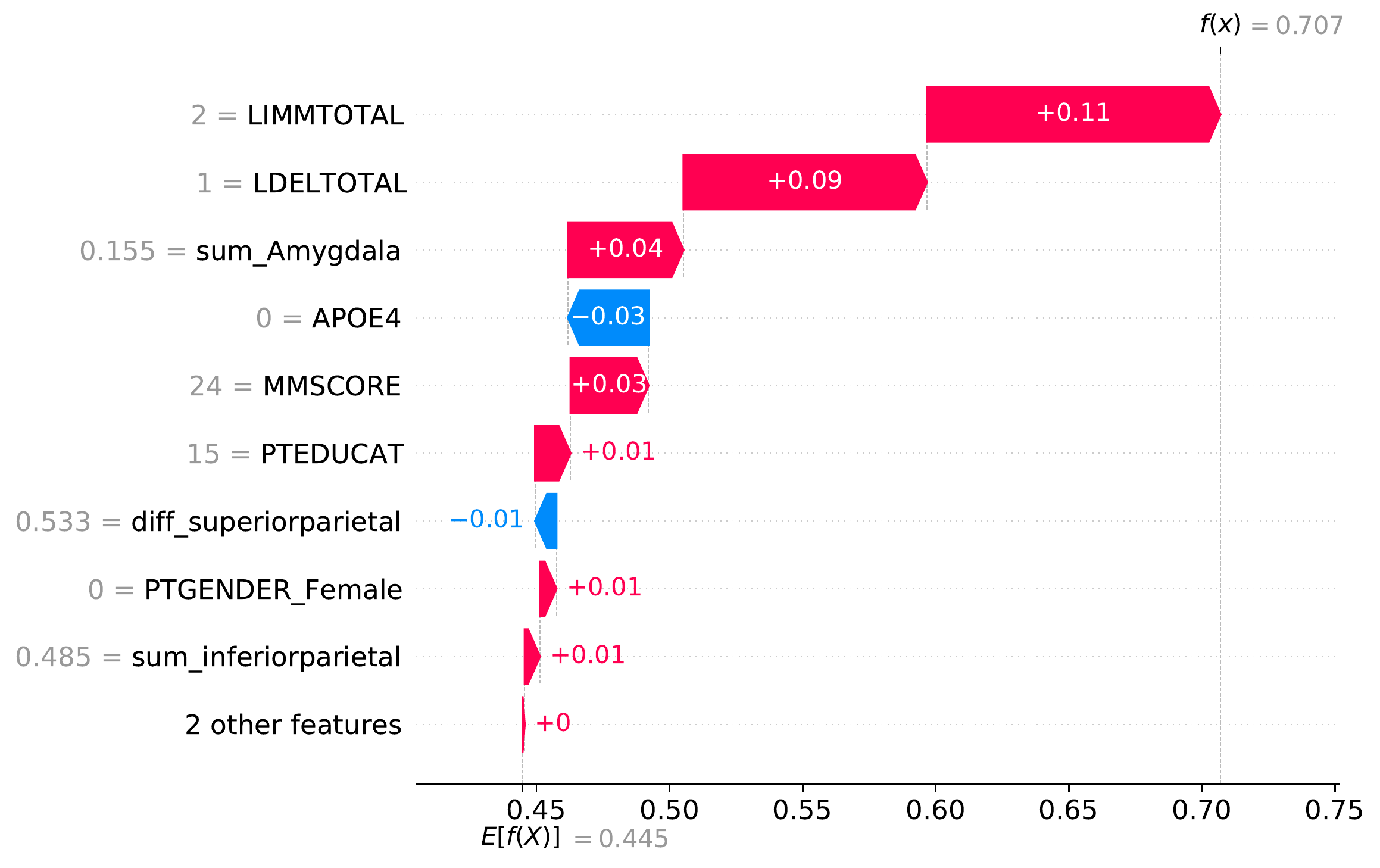}
		\caption{PTID: 036\_S\_4430, diagnosis: pMCI subject which reverted to MCI, model prediction value: 0.707}
		\label{fig:SHAP_Force_plot_036_S_4430}
	\end{subfigure} 
	\begin{subfigure}[b]{0.49\textwidth}
		\centering
		\includegraphics[width=\textwidth]{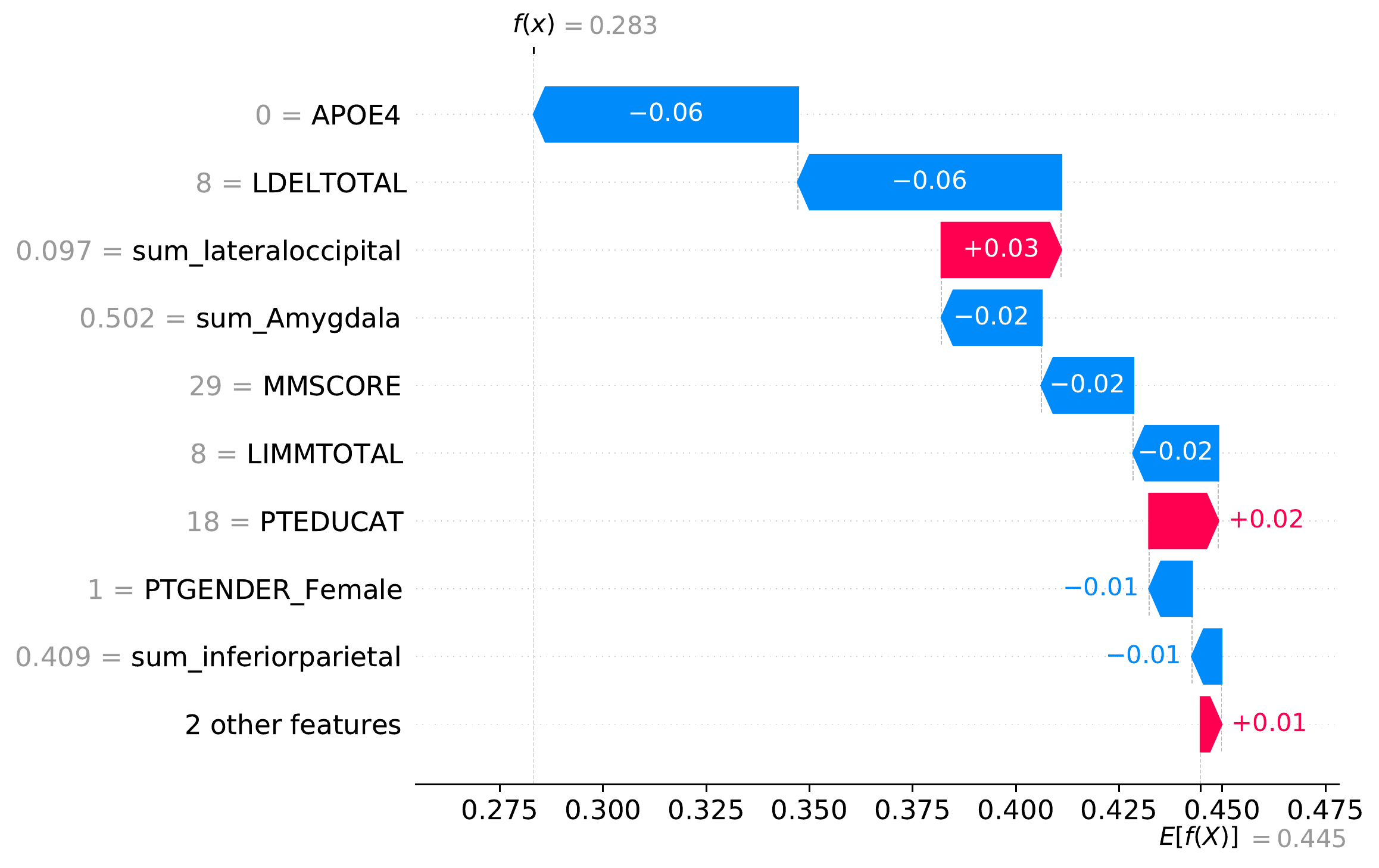}
		\caption{PTID: 128\_S\_0135, diagnosis: pMCI subject which reverted to MCI, model prediction value: 0.283}
		\label{fig:SHAP_Force_plot_128_S_0135}
	\end{subfigure}
	
	\caption{SHAP waterfall plots for interesting ADNI subjects to explain the prediction of the RF trained with feature selection and FS-3 for sMCI vs. pMCI classification. Figure \ref{fig:SHAP_Force_plot_027_S_0256} and Figure \ref{fig:SHAP_Force_plot_941_S_4036} show subjects of the ADNI test set, Figure \ref{fig:SHAP_Force_plot_036_S_4430} and Figure \ref{fig:SHAP_Force_plot_128_S_0135} explain model predictions of two subjects not included in the sMCI vs. pMCI dataset because those pMCI subjects reverted to MCI at a later visit. The arrow length indicates a Shapley value of the feature expression. Pathogenic feature expressions are shown as red and protective expressions as blue arrows. All volumetric features were scaled to a range between zero and one. Mean prediction value (base value): 0.445}
	\label{fig:SHAP_force_plots}
\end{figure}

To investigate explanations for individual model predictions, Figure \ref{fig:SHAP_force_plots} shows SHAP waterfall plots of four ADNI subjects. Those plots visualize the predictions for the RF trained with FS-3, and feature selection for sMCI vs. pMCI classification. SHAP waterfall plots explain the difference between the average model prediction value ($E[f(X)]$) and the subject's model prediction based on Shapley values. In all plots, the individual prediction was the probability of the subject being classified as pMCI. Features with a pathogenic expression are shown as red and protective expressions as blue arrows. The model prediction for the subject with PTID 027\_S\_1387 is explained in Figure \ref{fig:SHAP_Force_plot_027_S_0256}. This is a subject from the ADNI test set and had a diagnosis of pMCI. The model prediction of this subject was 0.735. As this value was higher than 0.50, the subject was correctly classified as a pMCI subject. The most important feature with a pathogenic effect was the volume of the inferior parietal lobules. A relatively small normalized feature value of 0.237 was observed. The Shapley value of this feature was 0.12 and thus, this feature expression increased the model prediction by 0.12. The LDELTOTAL cognitive test score reached a feature value of 3, which was a relatively bad test performance and thus increased the subject's risk to develop AD. Surprisingly, the relatively old age of 85.6 years decreased the patient's risk to develop AD by 0.03.

The model prediction for an sMCI subject (PTID: 037\_S\_4146) is demonstrated in Figure \ref{fig:SHAP_Force_plot_941_S_4036}. This subject was sampled from the ADNI test set and reached a model prediction value of 0.149. The subject had a moderate to a large volume of amygdalae which decreased the subject's risk of prospectively developing AD by -0.15. The subject also has two ApoE$\epsilon$4 alleles and as the presence of ApoE$\epsilon$4 alleles is a risk factor for AD, this increased the patient's risk. Additionally, the relatively high LDELTOTAL cognitive test score of 9 had a protective effect. 

SHAP waterfall plots of subjects not included in the sMCI vs. pMCI dataset because those pMCI subjects reverted to MCI at a later visit (explained in Section \ref{SEC:SubjectSelection}) are visualized in Figure \ref{fig:SHAP_Force_plot_036_S_4430} and Figure \ref{fig:SHAP_Force_plot_128_S_0135}. The prediction of the subject with PTID 036\_S\_4430 is visualized in Figure \ref{fig:SHAP_Force_plot_036_S_4430}. This MCI subject converted to AD 5.54 months after the baseline visit, but reverted to MCI 12.00 months after the baseline, and again converted to AD 23.64 months after the baseline visit. The last diagnosis for this subject was recorded after 83.54 months. The subject reached a model prediction of 0.707 and was thus classified as a pMCI subject. Additionally, the patient had a relatively small LIMMTOTAL cognitive test score of 2. The model learned that this poor test score increased the patient's risk to develop AD by 0.09. Additionally, the subject had relatively small volumes of the amygdalae, which additionally decreased the patient's risk to develop AD in the future. The AD risk of this patient was decreased by 0.03 because the subject has no ApoE$\epsilon$4 alleles.

The SHAP force plot for the subject with PTID 128\_S\_0135 is visualized in Figure \ref{fig:SHAP_Force_plot_128_S_0135}. This MCI subject converted to AD 54.52 months after the baseline visit, reverted to MCI after 71.74 months and again converted to AD 83.90 months after the baseline visit which was also the last diagnosis available. However, in contrast to Figure \ref{fig:SHAP_Force_plot_036_S_4430}, the subject reached a small model prediction value of 0.283 and was therefore classified as an sMCI subject. The most important factor decreasing the patient's risk was the absence of ApoE$\epsilon$4 alleles. This factor decreased the model prediction by 0.06. Additionally, the LDELTOTAL cognitive test score of 8, which was relatively large had a protective effect. The relatively small normalized volume of the lateral occipital sulci decreased the patient's risk by 0.03. One reason for the classification score might be, that the conversion to AD was relatively late for this subject.

\section{Discussion}
\label{Sec:Discussion}

In comparison to previous research \cite{Bloch2021}, which exclusively trained tree-based models, this work trained several RFs, XGBoost models, DTs, SVMs, and LR models to detect different stages of AD. All models were trained using the ADNI dataset and validated using independent test sets of the ADNI, AIBL, and OASIS cohorts. Bayesian Optimization optimized for the best hyperparameters of the models. During this stage, CV was used to estimate the performance for independent test sets. The models were trained using three feature sets. The MRI features included summed volumes, differences, and ratios of predefined brain structures to investigate asymmetry structures associated with different AD stages. Forward feature selection was implemented to focus the models on the most important features and simultaneously avoid correlated features in the datasets. The performances of the different ML models as well as the different feature sets are compared to each other using Friedman tests and pairwise Wilcoxon signed rank tests with Bonferroni adjustment. SHAP summary plots were used to visualize and interpret those models. The resulting Shapley values were compared to permutation importance of all models as well as natural feature importances of the RF and XGBoost models and to log odd's ratios of the LR models. As correlated features reduce the validity of explainability methods like permutation importance and SHAP \cite{10.1007/978-3-030-43823-4_17}, those were also calculated consolidating correlated features to aspects. SHAP force plots investigated individual predictions of interesting subjects.

The experimental results showed the forward feature selection chose brain regions that were previously associated with AD progression \cite{10.1038/nrneurol.2009.215,10.1002/hbm.20934,10.1016/j.neurobiolaging.2003.12.007,10.1016/j.pscychresns.2011.06.014,10.3233/jad-2011-101782,10.1136/jnnp.72.4.491} for all classification tasks and models. The performances achieved for models trained with forward feature selection did not outperform the models trained on the entire feature set.

The pairwise Wilcoxon signed rank tests with Bonferroni adjustment showed, that the results of models trained with FS-3, which included cognitive test results, outperformed those models trained for FS-1 and FS-2 for all classification tasks and the ADNI test set. The improvements for FS-3 models in comparison to FS-1 and FS-2 models were smaller for sMCI vs. pMCI than for the baseline classification tasks. The SHAP summary plots of all feature sets mainly showed biologically plausible associations and the most important features for the CN vs. AD classification using FS-3 and the polynomial SVM were the cognitive test scores LDELTOTAL and MMSCORE. 

The results for the AIBL and OASIS test sets showed less clear advantages of FS-3. Reasons for this were, among others, differences in the subject recruitment process, leading to differences in sociodemographics and differing MRI protocols across studies. However, the CN vs. AD models were successfully transferred to AIBL and OASIS by mostly achieving classification accuracies better than the no information rate. Additionally, the models trained for MCI vs. AD classification and sMCI vs. pMCI classification were successfully transferred to the AIBL dataset. For CN vs. MCI classification, poor results worse than the no information rate were achieved for the AIBL and OASIS datasets. However, the SHAP summary plots of those models mainly showed biologically plausible results. It was observed that age was a highly important feature in some of those models, which might cause problems transferring those models to datasets with differing demographic distributions. Poor results were also achieved for the MCI vs. AD classification and the OASIS dataset.

Some of the black-box models outperformed the simple and interpretable DTs. However, the pairwise Wilcoxon signed rank tests with Bonferroni adjustment ($p-value<0.05$) showed no significant differences. No model stood out among the black-box models. However, different ML models learned different associations which mostly were biologically plausible. The SHAP summary plots were compared to the permutation importance of all models, and natural RF and XGBoost feature importances, as well as absolute log odds of the LR and agreed for many features. The feature rankings of all models were compared to each other using Kendall's rank correlation.  

Because feature dependency structures reduce the validity of SHAP values and permutation importances, those feature importances were also computed by consolidating correlated features using aspects. The results show that the models depended on biologically plausible features. However, the feature rankings of the SHAP values and the permutation importances showed a weaker correlation than those calculated for the models with forward selected features.

Individual predictions, which are important in clinical practice, were interpreted using SHAP waterfall plots.
\subsection{Limitations}
\label{Sec:Limitations}
The approach proposed in this article had several limitations. First, both external datasets had a clear focus on CN subjects and were thus imbalanced which makes the interpretation of model generalizability hard. The external validation of the sMCI vs. pMCI classification, which was medically more interesting than the baseline diagnoses, was based only on 28 AIBL subjects and no OASIS subjects. Future investigations should include more AD datasets knowing those cohorts differ in inclusion criteria. Possible cohorts might be the AD subset~\cite{10.1159/000320988} of the HNR~\cite{10.1067/mhj.2002.123579} or a subset of the National Alzheimer's Coordinating Center~\cite{Beekley2004}. In this context, instead of diagnoses, different biomarkers should be addressed as endpoints. Another idea to increase the number of subjects in the datasets is, to relax the exclusion criteria by also including subjects that reverted to MCI or CN, and use follow-up scans of subjects where the baseline scan failed for the MRI feature extraction pipeline. Due to the availability of data in the cohorts, and minimal invasive recording, only MRI, sociodemographics, the number of ApoE$\epsilon$4 alleles, and cognitive test scores were included in the investigations. However, PET scans and biomarkers have high medical relevance and should thus be considered in future investigations.

Although in comparison to previous research \cite{Bloch2021}, the number of ML models was already increased, prospectively deep learning models like CNNs, which can automatically extract locally textural features from MRI scans should be investigated. However, currently, there is no consensus on whether those methods can improve AD detection. Much previous work in this area suffered from data leakage~\cite{10.1016/j.media.2020.101694} or investigated the less challenging discrimination between AD and CN. The Bayesian optimization used for hyperparameter-tuning is a sequential method. Future work should therefore investigate the use of more effective parallelized methods such as presented in \cite{Snoek2012}.


\section{Conclusion}
\label{Sec:Conclusion}
This work extended a workflow \cite{Bloch2021} to explain ML black-box models trained to distinguish multiple AD stages using Shapley values. The differentiation of sMCI and pMCI subjects is of medical interest to recruit and monitor subjects for therapy studies. The approach was based on non-invasive features including MRI volumes, sociodemographic data, the number of ApoE$\epsilon$4 alleles, and cognitive test results. Volumetric features were extracted from the MRI scans using the FreeSurfer pipeline. The sum, difference, and ratio of the volumes of both hemispheres were calculated to investigate the brain asymmetry in multiple AD stages. Shapley sampling values were calculated to visualize the local feature associations of black-box RFs, XGBoost models, and SVMs. The experiments mainly showed biologically plausible associations and improved results for models including cognitive test scores. Those improvements were smaller for sMCI vs. pMCI classification.

For the investigation of model reproducibility, all models were trained for the ADNI dataset and validated for the external AIBL and OASIS cohorts. The ADNI models achieved reasonable results for AIBL and CN vs. AD, MCI vs. AD, and sMCI vs. pMCI classification. For the OASIS test set, reasonable results were only reached for CN vs. AD classification. 

Some of the performances of the black-box models outperformed the simple and interpretable DTs. None of the black-box models achieved outstanding results. SHAP summary plots were used to visualize the associations, the model learned between the features and the AD diagnosis. The most important features of those plots were previously associated with AD progression. Additionally, those plots showed biologically plausible associations for most of the important features in all classification tasks.

SHAP force plots investigated individual model predictions. The comparison between SHAP values and natural RF and XGBoost feature importances showed high correlations of the SHAP importances, the permutation importance, and the natural feature importances for all models.

The investigation of the feature dependency structure and consolidating correlated features during the computation feature importance computation for sMCI vs. pMCI classification showed, that those models depended on features that were previously associated with AD.

This work outperformed previous work \cite{Bloch2021} for the ADNI and AIBL classification results and CN vs. MCI classification and the AIBL results in MCI vs. AD classification for models trained without cognitive test scores. Additionally, the ADNI and AIBL results achieved for sMCI vs. pMCI classification trained with cognitive test scores outperformed the results of previous work.

\backmatter


\bmhead{Acknowledgments}
	Data collection and sharing for this project was funded by the Alzheimer's Disease Neuroimaging Initiative (ADNI) (National Institutes of Health Grant U01 AG024904) and DOD ADNI (Department of Defense award number W81XWH-12-2-0012). ADNI is funded by the National Institute on Ageing, the National Institute of Biomedical Imaging and Bioengineering, and through generous contributions from the following: AbbVie, Alzheimer's Association; Alzheimer's Drug Discovery Foundation; Araclon Biotech; BioClinica, Inc.; Biogen; Bristol-Myers Squibb Company; CereSpir, Inc.; Cogstate; Eisai Inc.; Elan Pharmaceuticals, Inc.; Eli Lilly and Company; EuroImmun; F. Hoffmann-La Roche Ltd and its affiliated company Genentech, Inc.; Fujirebio; GE Healthcare; IXICO Ltd.; Janssen Alzheimer Immunotherapy Research \& Development, LLC.; Johnson \& Johnson Pharmaceutical Research \& Development LLC.; Lumosity; Lundbeck; Merck \& Co., Inc.; Meso Scale Diagnostics, LLC.; NeuroRx Research; Neurotrack Technologies; Novartis Pharmaceuticals Corporation; Pfizer Inc.; Piramal Imaging; Servier; Takeda Pharmaceutical Company; and Transition Therapeutics. The Canadian Institutes of Health Research is providing funds to support ADNI clinical sites in Canada. Private sector contributions are facilitated by the Foundation for the National Institutes of Health (www.fnih.org). The grantee organization is the Northern California Institute for Research and Education, and the study is coordinated by the Alzheimer's Therapeutic Research Institute at the University of Southern California. ADNI data are disseminated by the Laboratory for Neuro Imaging at the University of Southern California.
Data used in the preparation of this article were obtained from the Alzheimer's Disease Neuroimaging Initiative (ADNI) database (\url{https://adni.loni.usc.edu}, Accessed: 2022-05-01). As such, the investigators within the ADNI contributed to the design and implementation of ADNI and/or provided data but did not participate in analysis or writing of this report. A complete listing of ADNI investigators can be found online (\url{http://adni.loni.usc.edu/wp-content/uploads/how_to_apply/ADNI_Acknowledgement_List.pdf}, Accessed: 2022-05-01).

The authors thank Ahmad Idrissi-Yaghir, Department of Computer Science, University of Applied Sciences and Arts Dortmund, 44227 Dortmund, Germany, for the constructive proofreading of the manuscript.

\section*{Declarations}
\subsection*{Funding}
The work of Louise Bloch was partially funded by a PhD grant from University of Applied Sciences and Arts Dortmund, Dortmund, Germany.
\subsection*{Competing Interests}
The authors declare that they have no competing interests.
\subsection*{Ethics Approval}
Not applicable.
\subsection*{Consent to Participate}
The ADNI study was approved by the institutional review boards of the participating institutions. All participants gave informed written consent. More details can be found online (\url{https://adni.loni.usc.edu}, Accessed: 2022-05-01).

The AIBL study was approved by the institutional ethics committees of Austin Health, StVincent's Health, Hollywood Private Hospital and Edith Cowan University. All participants gave written informed consent before participating in the study.

All OASIS participants consented to Knight Aging and Disability Resource Center (ADRC)-related projects following procedures approved by the Institutional Review Board of Washington University School of Medicine. Participants consented to the use of their data by the scientific community and data sharing terms have been approved by the Washington University Human Research Protection Office. 
\subsection*{Consent for Publication}
Consent for publication has been granted by ADNI administrators.
\subsection*{Availability of Data and Materials}
Data used in preparation of this article were
obtained from the Alzheimer's Disease
Neuroimaging Initiative (ADNI) database (\url{https://
adni.loni.usc.edu}, Accessed: 2022-05-01) and Open Access Series of Imaging Studies (OASIS) (\url{https://www.oasis-brains.org/}, Accessed: 2022-05-01). Details about data access are detailed there. The authors had no special access privileges others would not have to the data obtained from the Alzheimer's Disease Neuroimaging Initiative (ADNI) or Open Access Series of Imaging Studies (OASIS) databases.
\subsection*{Code Availability}
The workflow implementation will be available online after acceptance: \url{https://github.com/LouiseBloch/AlzheimerExplainableMLCorrelations}.
\subsection*{Authors' Contributions}
The conceptualization of the study was carried out by CMF and LB, CMF and LB planned the experiments. LB implemented the software, executed the experiments, analysed the data and has written the original draft under the supervision of CMF. All authors read and approved the final manuscript.

\bibliography{submission}


\end{document}